\documentclass[a4paper,12pt]{article}
\usepackage[margin=1in]{geometry}
\usepackage[utf8]{inputenc}
\usepackage[numbers]{natbib}
\usepackage{graphicx}
\usepackage{amsmath}
\usepackage[version=4]{mhchem}
\usepackage{siunitx}
\usepackage{longtable,tabularx}
\usepackage{xcolor}
\usepackage{booktabs}
\usepackage{algorithm}
\usepackage{algpseudocode}

\usepackage{bm}
\usepackage{amsfonts}
\usepackage{bbm}
\usepackage{nicefrac}       
\usepackage{microtype}      
\usepackage{authblk}
\usepackage{hyperref}
\usepackage[capitalize,noabbrev]{cleveref}
\usepackage{wrapfig}
\usepackage{subcaption}
\usepackage{comment}
\usepackage{empheq}
\usepackage[most]{tcolorbox}
\usepackage{float}

\newcommand{\mbb}{\mathbb}
\newcommand{\mbf}{\mathbf}
\newcommand{\mcl}{\mathcal}
\newcommand{\bs}{\boldsymbol}

\newcommand{\T}{\textnormal}
\newcommand{\x}{\mathbf{x}}
\newcommand{\X}{\mathcal{X}}

\title{A Bayesian latent Gaussian process framework for aerodynamic uncertainty quantification}

\author{Geoffrey Davis \footnote{Undergraduate research assistant, Aerospace Engineering, The Pennsylvania State University, University Park, PA 16802} and Ashwin Renganathan \footnote{Assistant professor, Aerospace Engineering and the Institute of Computational and Data Science (ICDS), The Pennsylvania State University, University Park, PA 16802}}

\affil{The Pennsylvania State University, University Park, PA 16802}
\date{}

\begin{document}

\maketitle

\begin{abstract}
Predicting the aerodynamic performance (e.g. lift, drag, and moment coefficients) of an aircraft is challenging -- computational models are biased and direct simulations are prohibitive. A pragmatic way to overcome this limitation is by calibrating low-fidelity computational predictions with experimental measurements. This, however, requires calibrating against \emph{sparse} measurements contaminated with \emph{uncertainty} in both the control inputs and the measured aerodynamic response. We develop a methodology to address this problem based on Gaussian process surrogates and the classical Kennedy-O'Hagan calibration. A surrogate model learned on abundant-but-cheap low-fidelity data is calibrated with a sparse set of measurement data. Crucialy, we develop a Bayesian latent Gaussian process based approach that marginalizes the calibrated surrogate model over the input uncertainty, while also matching the marginal mean and variance of the measured output uncertainty. Once calibrated, our surrogate model predicts the uncertainty in aerodynamic coefficients with very high accuracy, including at extrapolative input settings. We validate our calibrated surrogate model predictions against measurement data with \emph{true} uncertainty intervals to demonstrate that the model places $94.2-95.8\%$ of its predictive samples inside the released $95\%$ truth intervals, with endpoint cumulative probabilities very close to the nominal 0.025 and 0.975 levels.
\end{abstract}

\section{Introduction}

Computational aerodynamic tools are now used routinely in aircraft design, analysis, and certification studies, which makes uncertainty quantification (UQ) central to credible simulation-based prediction. Although the need for UQ is widely recognized, rigorous uncertainty propagation in computational fluid dynamics (CFD) remains difficult because aerodynamic predictions can depend on uncertain operating conditions, numerical approximations, surrogate-model error, and model-form discrepancy~\cite{Oberkampf2004}. These challenges are especially pronounced when only sparse validation data are available. Verification, validation, and UQ (VVUQ) frameworks therefore emphasize the need to characterize parametric input uncertainty, numerical uncertainty, and model-form uncertainty together rather than treating them as separate afterthoughts~\cite{Oberkampf2004}.

Community benchmark problems provide a useful setting for developing and comparing such methods. The AIAA Fluid Dynamics Technical Committee's Uncertainty Quantification Discussion Group (UQDG) has organized a series of UQ challenge problems for aerodynamics~\cite{Cary2022}. This paper addresses the \emph{Second Uncertainty Quantification Challenge Problem for Aerodynamics}~\cite{cary2026summary}, which builds on the first challenge problem~\cite{cary2024overview}. The first challenge asked whether a NACA~2412 airfoil with a 30\% chord flap would satisfy specified lift and pitching-moment requirements under uncertain inputs. The second challenge shifts the focus from requirement satisfaction to predictive uncertainty: participants are given sparse truth data with epistemic uncertainty bounds and must predict uncertainty in aerodynamic coefficients at unseen operating conditions. This is the focus of the article and we present more detail as follows.

%

\subsection{The UQ challenge problem}
\label{sec:uqchallenge}
The second AIAA UQ challenge problem is anchored on the NACA~2412 airfoil at a nominal Reynolds number of $700{,}000$, a 30\% chord flap, and sea-level conditions; the resulting Mach number lies in the incompressible regime. The operating envelope is defined by angles of attack in $[-5,10]$ degrees and flap angles in $[-5,15]$ degrees, with epistemic uncertainty also assigned to the Reynolds number. Within this Reynolds number, angle-of-attack, and flap-angle parameter space, truth data are provided at seven operating conditions. At each of these locations, the aerodynamic coefficients -- drag $C_d$, lift $C_l$, and quarter-chord pitching moment $C_m$ -- are supplied together with epistemic uncertainty intervals on both the input parameters and the coefficients; these data are summarized in \Cref{tab:given_data}.

\begin{figure}
    \centering
    \includegraphics[width=0.75\linewidth]{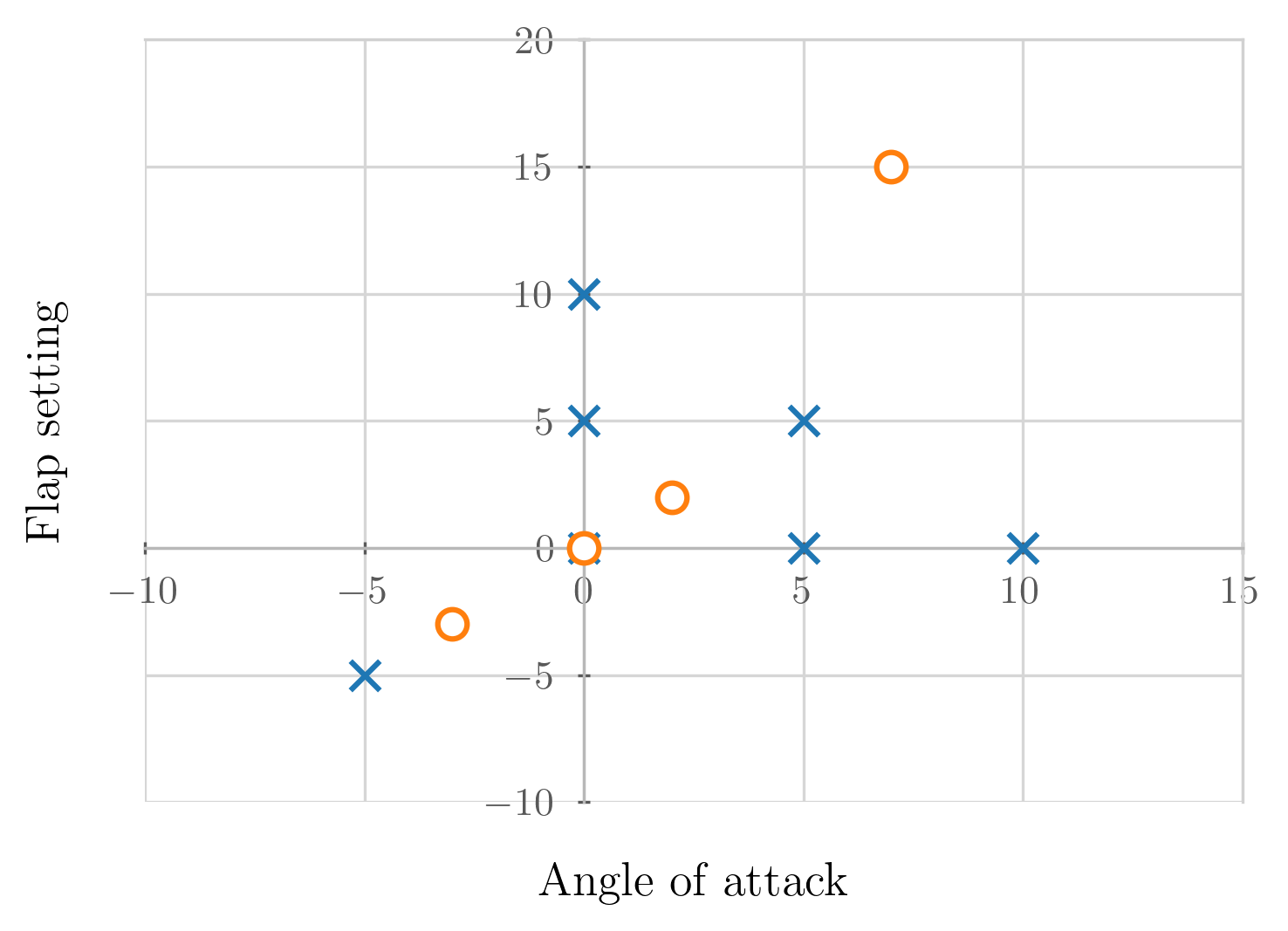}
    \caption{Operating points for the UQ challenge problem. The blue markers denote the seven calibration points where truth data are provided, and the orange markers denote the four prediction points where uncertainty estimates are required.}
    \label{fig:aiaa_graph}
\end{figure}

The computational model used to generate the initial low-fidelity data is XFOIL~\cite{drela1989xfoil}, a lightweight aerodynamic analysis tool appropriate for the operating conditions considered here. We use the XFOIL data to train surrogate models for the aerodynamic coefficients and then calibrate those surrogates with the truth data in \Cref{tab:given_data}. The resulting calibrated model is used to predict uncertainty in the aerodynamic coefficients at four unseen operating conditions, shown by the orange symbols in \Cref{fig:aiaa_graph}. The released truth intervals for these four prediction locations~\cite{cary2026summary} are used only for a posteriori assessment of the surrogate predictions.
\begin{table}[h]
\centering
\caption{Provided calibration data with epistemic uncertainty intervals}
\begin{tabular}{ccc|ccc}
\toprule
$\alpha \pm 0.02^\circ$ & $\beta_{\mathrm{flap}} \pm0.1^\circ$ & Re $\pm 3500$ & $C_l \pm 0.009$ & $C_d \pm0.0008$ & $C_m\pm0.008$\\
\midrule
0 & 0 & $\num{7e5}$ & $0.214$ & $0.0121$ & $-0.047$ \\
5 & 0 & $\num{7e5}$ & $0.737$ & $0.0143$ & $-0.043$ \\
10 & 0 & $\num{7e5}$ & $1.209$ & $0.0205$ & $-0.033$ \\
0 & 5 & $\num{7e5}$ & $0.536$ & $0.0131$ & $-0.094$ \\
0 & 10 & $\num{7e5}$ & $0.836$ & $0.0157$ & $-0.137$ \\
5 & 5 & $\num{7e5}$ & $1.041$ & $0.0176$ & $-0.087$ \\
-5 & -5 & $\num{7e5}$ &$-0.644$ & $0.0148$ & $-0.004$ \\
\bottomrule
\end{tabular}
\label{tab:given_data}
\end{table}

\subsection{Background and literature review}

\subsubsection{XFOIL aerodynamics model for low-fidelity data generation}
\label{sec:xfoil}
We use XFOIL v6.99 with a 100-panel baseline setup that accompanied the challenge problem. For cases at higher flap angles or angles of attack, where XFOIL may fail to converge, we incorporated a driver that first accumulates converged solutions at lower angles of attack for the same flap setting and then continues toward the desired condition. To train the XFOIL surrogate, we sample flap angle, angle of attack, and Reynolds number over the challenge envelope: angle of attack from $-5^\circ$ to $10^\circ$, flap angle from $-5^\circ$ to $15^\circ$, and Reynolds number from $665000$ to $735000$. A set of $100$ training points is generated via Latin hypercube sampling~\cite{loh1996latin}, and XFOIL is used to compute the corresponding aerodynamic coefficients. These data define the simulator response that is emulated by the surrogate models reviewed next.

\subsubsection{Gaussian process models}
Gaussian process (GP) models are nonparametric probabilistic models that are particularly suited for small data setting in the presence of noise. GP models have been widely used for surrogate modeling in aerodynamics~\cite{renganathan2021enhanced,ashwin2022data}, reliability analysis~\cite{renganathan2022multifidelity, renganathan2023camera,booth2025contour, renganathan2024efficient}, multidisciplinary analysis~\cite{yuan2026remal}, fluid-structure interactions~\cite{adhikary2026adaptive}, and optimization~\cite{carlson2025multiobjective, renganathan2025q,renganathan2021lookahead}. Recently, \citet{davis2026uncertainty} applied them to aerodynamic UQ. Our approach builds on GP based regression and we now briefly review some of their background.

Let $f: \mathbb{R}^d \rightarrow \mathbb{R}$ be a black-box function; we place a Gaussian process (GP) prior on $f$, $f(\x) \sim \mcl{GP}(0, k(\x, \cdot))$, where $k(\cdot, \cdot): \X \times \X \rightarrow \mbb{R}_{+}$ denotes the covariance function (also known as the \emph{kernel}). Observations from $f$ are modeled as $y_i = f(\x_i) + \epsilon_i$ for $i=1,\ldots,n$, where each $\epsilon_i$ is assumed to be zero-mean Gaussian noise with unknown variance $\sigma^2$. Using the observed dataset $\mcl{D}_n = \{\x_i, y_i, \sigma^2\}_{i=1}^n$, we fit a posterior GP---parameterized by hyperparameters $\bs{\Omega}$---to obtain the following conditional posterior distribution~\cite{Rasmussen2006}:
\begin{equation*}
\begin{split}
Y(\x) \mid \mcl{D}_n, \bs{\Omega} &\sim \mcl{GP}(\mu_n(\x), \sigma^2_n(\x)), \\
\mu_n(\x) &= \mbf{k}_n^\top [\mbf{K}_n + \tau^2 \mbf{I}]^{-1} \mbf{y}_n, \\
\sigma^2_n(\x) &= k(\x, \x) - \mbf{k}_n^\top [\mbf{K}_n + \tau^2 \mbf{I}]^{-1} \mbf{k}_n,
\end{split}
\end{equation*}
where $\mbf{k}_n \equiv k(\x, X_n)$ is the vector of covariances between the point $\x$ and all observation sites in $\mcl{D}_n$, $\mbf{K}_n \equiv k(X_n, X_n)$ is the covariance matrix over the observed points, $\mbf{I}$ is the identity matrix, and $\mbf{y}_n$ is the vector of corresponding outputs. The observation sites are denoted by $X_n = [\x_1,\ldots,\x_n]^\top \in \mbb{R}^{n \times d}$. Such ``standard'' GPs do not have a natural way of accommodating input uncertainty; they must be fit to some chosen value, typically the centroid, within the given epistemic range. To overcome this limitation, we introduce latent GPs (later in \Cref{sec:latent_gps}), which treat the input as a random variable to rigorously account for uncertainty.

%
%

\subsubsection{Literature review}

GP surrogates provide a probabilistic representation of an expensive deterministic or noisy model response from a finite set of simulator evaluations \cite{sacks1989design,santner2003design,Rasmussen2006,gramacy2020surrogates}. Their value for uncertainty quantification is twofold: the posterior mean provides a smooth response approximation, while the posterior covariance provides a local measure of surrogate uncertainty. These properties make GP surrogates a natural tool for aerodynamic UQ, where repeated evaluations of computational models may be expensive, convergence may be nonuniform across the input space, and uncertainty must be propagated through quantities of interest such as $C_l$, $C_d$, and $C_m$. 

\paragraph{GP surrogates in model calibration.}
Model calibration refers to learning how much a model of a physical system falls short of predicting reality, so that the model can be corrected for improved predictive accuracy. 
The classical Bayesian calibration framework of Kennedy and O'Hagan models the physical response $\zeta(\x)$ as the sum of a computer-model response and an additive model-discrepancy term \cite{kennedy2001bayesian}. GP models are a natural fit for model calibration under the Kennedy-O'Hagan (KOH) framework~\cite{kennedy2001bayesian,higdon2004combining,higdon2008computer}.
In a common notation,
\begin{equation}
    y_i = \zeta(\x_i) + \epsilon_i, 
    \qquad
    \zeta(\x) = \eta(\x,\theta) + \delta(\x),
    \label{eq:koh_review}
\end{equation}
where $\eta(\x,\theta)$ is the simulator evaluated at physical input $\x$ and calibration parameter $\theta$, $\delta(\x)$ is the model-form discrepancy, and $\epsilon_i$ denotes observational error. In the original formulation, GP priors are used to represent the unknown response of the simulator and the discrepancy. When the simulator is inexpensive, $\eta$ may be evaluated directly during inference; when the simulator is expensive, a GP surrogate is first trained on a designed set of computer runs and then embedded within the calibration model. This surrogate-based interpretation is now standard in Bayesian calibration because it permits posterior inference over calibration parameters, discrepancy fields, and predictions without requiring repeated high-cost simulator calls.

Subsequent work extended the KOH construction to more complex data settings. \citet{higdon2004combining} demonstrated Bayesian calibration procedures that combine field observations with computer simulations for prediction and uncertainty quantification. The same line of work was extended to high-dimensional simulator output, where basis representations are combined with GP models to make calibration tractable for functional or spatially distributed quantities \cite{higdon2008computer}. \citet{bayarri2007framework} developed a broader validation framework for computer models that separates calibration, discrepancy assessment, and prediction, emphasizing the need to account for both experimental uncertainty and simulator inadequacy. Modular Bayesian strategies, in which emulator training, calibration, and prediction are separated to improve stability or reduce feedback between model components, have also been proposed for computer-model analysis \cite{liu2009modularization}.

For computational fluid dynamics and aerodynamics, the KOH structure is particularly relevant because even carefully verified solvers may remain biased relative to physical truth due to turbulence modeling, transition modeling, geometry idealization, numerical discretization, or reduced-order physical assumptions. In such cases, a GP discrepancy model can be interpreted as a spatially correlated correction field over the aerodynamic input space. Multifidelity and model-form uncertainty studies in CFD have used related GP discrepancy constructions to combine low-fidelity simulations, higher-fidelity simulations, and available experimental data while propagating input uncertainty through the corrected predictive model \cite{wang2017propagation}. This view is consistent with the present work: we train a surrogate model on XFOIL data and learn an additive correction model that represents the residual difference between XFOIL predictions and the supplied truth data.

A well-known limitation of the KOH framework is the potential lack of identifiability between calibration parameters and the discrepancy function. When $\delta(\x)$ is sufficiently flexible, it can absorb errors that might otherwise be attributed to $\theta$, leading to weakly identified or physically misleading calibration-parameter posteriors. This issue has been examined from several perspectives, including the role of discrepancy modeling in physical-parameter learning \cite{brynjarsdottir2014learning}, model discrepancy and identifiability analysis \cite{arendt2012quantification}, and asymptotic properties of calibration estimators for imperfect computer models \cite{tuo2015efficient,tuo2016theoretical}. Alternative formulations, such as orthogonal or constrained discrepancy priors, have been proposed to reduce confounding between calibration parameters and discrepancy fields \cite{plumlee2017bayesian}.

The present work uses the KOH idea primarily for predictive correction rather than for inference on tunable physical calibration parameters. The internal XFOIL settings are fixed, and the unknown term of interest is the discrepancy between the XFOIL-based aerodynamic response and the truth data. This distinction reduces the emphasis on physical-parameter identifiability and shifts the modeling objective toward calibrated prediction and uncertainty propagation. In this setting, the additive construction
\begin{equation}
    g(\x) = f_X(\x) + \delta(\x),
    \label{eq:composite_gp_review}
\end{equation}
where $f_X$ is the GP surrogate of XFOIL and $\delta$ is a GP discrepancy model, provides a direct surrogate-based analog of \Cref{eq:koh_review}. Unlike common implementations that train the discrepancy only on centroid residuals, the proposed ``distributional calibration'' (more detail in \Cref{sec:distributional_calibration}) treats the truth-data intervals as finite-dimensional marginal constraints on the corrected GP. Thus, the truth data are used not only to correct the posterior mean, but also to calibrate the posterior variance of the corrected aerodynamic surrogate.

\paragraph{GP surrogates with uncertain inputs.}
Standard GP regression assumes that each input location is observed exactly. This assumption is restrictive in UQ settings where operating conditions, experimental settings, or geometric parameters are reported with measurement error, epistemic intervals, or incomplete knowledge. A direct response is to treat uncertain inputs probabilistically and integrate the GP prediction over the input distribution. Early work on GP prediction with uncertain inputs developed moment-matching and Gaussian approximation strategies for propagating input uncertainty through Bayesian kernel models, particularly for multi-step-ahead forecasting \cite{girard2003gpuncertain,quinonero2003propagation}. Related methods derived analytic or approximate expressions for GP prediction under Gaussian input uncertainty, especially when squared-exponential kernels make the required integrals tractable \cite{dallaire2009learning}.  \citet{mchutchon2011gaussian} further showed that input noise can be interpreted, under local linearization, as a form of heteroscedastic output noise, leading to practical training procedures for GPs with noisy inputs.

A related but distinct class of models treats the input coordinates themselves as latent variables. In Gaussian process latent variable models (GP-LVMs), unknown latent coordinates are inferred jointly with the GP mapping, providing a probabilistic nonlinear dimension-reduction framework \cite{lawrence2005probabilistic}. Bayesian extensions place priors on the latent coordinates and integrate over their uncertainty using variational inference or related approximations \cite{titsias2010bayesian,damianou2016variational}. Although GP-LVMs were originally developed for latent representations rather than physical input uncertainty, the same statistical idea is useful when the true physical input is unobserved but constrained by prior information. In the present aerodynamic setting, the latent variables are not abstract coordinates; they are physical quantities such as angle of attack, Reynolds number, and flap setting, constrained to lie within specified epistemic intervals.

Input uncertainty has also been considered directly in Bayesian calibration of time-consuming simulators. \citet{perrin2019taking} formulated calibration procedures that account for uncertain simulator inputs rather than assuming that field inputs are known exactly. Such approaches are important when discrepancies between simulation and observation can be caused either by model-form error or by uncertainty in the conditions at which the physical data were obtained. Collapsing an uncertain input interval to a midpoint can then bias the inferred discrepancy, underestimate predictive uncertainty, or produce an overconfident correction model.

The latent-input GP construction used in this work follows this general philosophy while targeting the epistemic interval structure of the AIAA UQ challenge data. For each observation, the recorded input is an interval or box $U_i$, and the true but unknown input $\tilde{\x}_i$ is modeled as a latent variable supported on $U_i$. A prior density $\pi_i(\tilde{\x}_i)$ encodes the epistemic information within the box, and the GP likelihood is evaluated at the latent locations rather than at fixed centroids. This formulation reduces to standard GP regression when each $U_i$ collapses to a point, but it retains the ability to infer plausible input locations when the intervals have nonzero width. Computationally, the latent inputs can be inferred by full Bayesian sampling, variational inference, or maximum a posteriori optimization. The latter provides a practical compromise for the small truth-data regime considered here.

This latent-input treatment is complementary to the Kennedy--O'Hagan discrepancy model. The discrepancy GP accounts for model-form error between XFOIL and truth, while the latent-input GP accounts for epistemic uncertainty in the locations at which the truth data are observed. When the test inputs are also uncertain, prediction requires marginalization over the test input distribution in addition to conditioning on the learned latent training inputs. The resulting predictive uncertainty combines the GP posterior variance, the dispersion induced by uncertain inputs, and the calibrated discrepancy uncertainty. This is the main distinction between the present framework and a conventional surrogate-bias correction trained only at nominal input centroids.

\subsubsection{Contributions}
Together, these ideas lead to a Bayesian surrogate framework for aerodynamic UQ with epistemic uncertainty in both inputs and outputs. The main contributions of this paper are as follows.
\begin{itemize}
    \item We construct a Kennedy--O'Hagan-style additive correction model in which one GP surrogate represents the XFOIL response and a second GP represents the discrepancy between XFOIL and the supplied truth data.
    \item We incorporate epistemic input intervals through a latent-input GP formulation, allowing the unknown true calibration inputs to be inferred within their prescribed uncertainty bounds rather than fixed at nominal centroids.
    \item We introduce a distributional calibration step for the corrected GP so that the supplied truth intervals constrain both the posterior mean and the marginal predictive variance of the aerodynamic coefficients.
    \item We demonstrate the predictive UQ capabilities of our method by compariing against the true uncertainty estimates at a set of prediction points.
    \item Our software implementation is openly available at \url{https://github.com/csdlpsu/uqchallenge2} for full reproducibility of our results.
\end{itemize}
The remainder of the manuscript develops these elements in sequence. \Cref{sec:method} presents the latent-input and distributionally calibrated GP methodology used in this work. \Cref{sec:experiments} describes the experimental setup and UQ results, and \Cref{sec:conclusions} summarizes the main conclusions.

\section{Methodology}
\label{sec:method}
\begin{figure}[htb!]
\centering
	\includegraphics[width=1\linewidth]{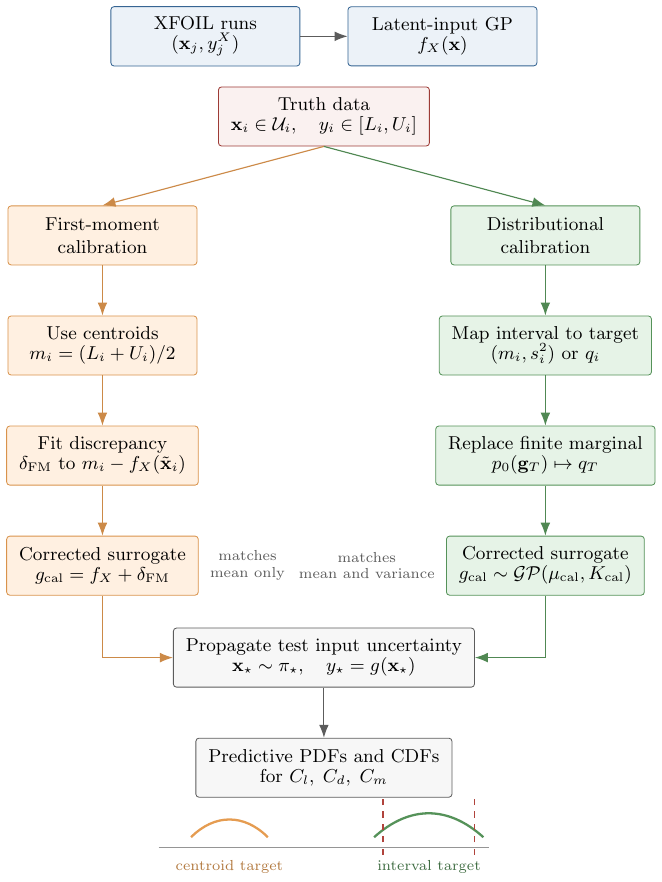}
\caption{Overview of the distinction between the first-moment and distributional calibration approaches.}
\label{fig:overview}
\end{figure}

Our overall approach involves three steps. First, using data generated by XFOIL, we learn a GP surrogate model for the aerodynamic coefficients $C_d$, $C_l$, and $C_m$ as functions of angle of attack $\alpha$, flap angle $\beta_{\mathrm{flap}}$, and Reynolds number; the input domain is fixed as described above. Second, a Bayesian latent GP fitting step learns maximum a posteriori (MAP) estimates of the unknown input locations at the given $7$ truth-data points. Third, a correction GP ($\delta$) is fit at the MAP estimates of the inputs and calibrated to match the given epistemic uncertainty intervals of the aerodynamic coefficients at the $7$ truth points. We now present details of the Bayesian latent GP model fitting and the distributional model calibration approach, with their distinctions summarized in \Cref{fig:overview}.

\subsection{Bayesian latent Gaussian process models}
\label{sec:latent_gps}
We now present a Gaussian process regression model for the common situation in which each recorded input is not known exactly but is instead provided as an epistemic interval (more generally, an axis–aligned box); in this work, we refer to this model class as ``latent GPs''. Let $d\in\mathbb{N}$ denote the input dimension and suppose we observe pairs $\{(\mathcal{U}_i,y_i)\}_{i=1}^n$, where $y_i\in\mathbb{R}$ is a scalar response and $\mathcal{U}_i \in \prod_{j=1}^d[\ell_i^{\,j},u_i^{\,j}]\subset\mathbb{R}^d$ is the set within which the unknown true input lies. We write $\tilde{\x}_i\in\mathbb{R}^d$ for this unknown true input, with $\tilde{\x}_i\in\mathcal{U}_i$. To express the epistemic uncertainty in $\tilde{\x}_i$, we place a prior density $\pi_i$ on $\tilde{\x}_i$ supported on $\mathcal{U}_i$; then, we estimate $\tilde{\x}$ from data via Bayesian inference.

Recall that we place a GP prior on $f$, and observe its noisy response according to
\begin{equation*}
\label{eq:likelihood}
y_i \;=\; f(\tilde{\x}_i) \;+\; \varepsilon_i,\qquad \varepsilon_i \stackrel{\text{iid}}{\sim} \mathcal{N}(0,\sigma^2).
\end{equation*}
Let $\tilde{X}=[\tilde{\x}_1,\ldots,\tilde{\x}_n]^\top\in\mathbb{R}^{n\times d}$ and $\mathbf{y}=(y_1,\ldots,y_n)^\top\in\mathbb{R}^n$. Then, the marginal likelihood of $\mathbf{y}$ given $\tilde{X}$, $\theta$, and $\sigma^2$ is multivariate normal,
\begin{equation}
\label{eq:marginal-lik}
\mathbf{y}\mid \tilde{X},\theta,\sigma^2 \;\sim\; \mathcal{N}\!\Big(\mu(\tilde{X}),\,K_\theta(\tilde{X},\tilde{X})+\sigma^2 \mbf I_n\Big),
\end{equation}
where $\mu(\tilde{X})=\big(\mu(\tilde{\x}_1),\ldots,\mu(\tilde{\x}_n)\big)^\top$, $K_\theta(\tilde{X},\tilde{X})\in\mathbb{R}^{n\times n}$ has $(i,j)$ entry $k_\theta(\tilde{\x}_i,\tilde{\x}_j)$, and $\mbf I_n$ is the $n\times n$ identity matrix.

In order to respect the box constraints $\tilde{\x}_i\in\mathcal{U}_i$ during gradient–based inference, it is convenient to reparameterize each latent input with an unconstrained vector $\mathbf{u}_i\in\mathbb{R}^d$ and a smooth, invertible map into the box. A simple choice is the elementwise logistic (sigmoid) map. Writing $\boldsymbol{\ell}_i=(\ell_i^{\,1},\ldots,\ell_i^{\,d})^\top$ and $\mathbf{u\!p}_i=(u_i^{\,1},\ldots,u_i^{\,d})^\top$ for the lower and upper corners of $\mathcal{U}_i$, and $\sigma(t)=1/(1+e^{-t})$ for the logistic function applied coordinate-wise defines
\begin{equation}
\label{eq:box-transform}
\tilde{\x}_i(\mathbf{u}_i) \;=\; \boldsymbol{\ell}_i \;+\; \sigma(\mathbf{u}_i)\,\odot\,\big(\mathbf{u\!p}_i-\boldsymbol{\ell}_i\big),
\end{equation}
where $\odot$ denotes the Hadamard (elementwise) product. The transformation \eqref{eq:box-transform} guarantees $\tilde{\x}_i(\mathbf{u}_i)\in\mathcal{U}_i$ for all $\mathbf{u}_i\in\mathbb{R}^d$, and the chain rule carries gradients $\partial/\partial \mathbf{u}_i$ through to the marginal likelihood in \eqref{eq:marginal-lik}. 

The full Bayesian target is the joint posterior density of the unknowns given the data:
\begin{equation}
\label{eq:posterior}
p\big(\tilde{X},\theta,\sigma^2 \,\big|\, \mathbf{y}\big) \;\propto\; 
\underbrace{p\big(\mathbf{y}\,\big|\,\tilde{X},\theta,\sigma^2\big)}_{\text{Gaussian likelihood from \eqref{eq:marginal-lik}}}\;
\underbrace{p(\tilde{X})}_{\text{product of input priors on the boxes}}\;
\underbrace{p(\theta)\,p(\sigma^2)}_{\text{hyperparameter priors}},
\end{equation}
where $p(\tilde{X})=\prod_{i=1}^n \pi_i(\tilde{\x}_i)$ encodes the epistemic beliefs on the inputs. One may explore \eqref{eq:posterior} by Markov chain Monte Carlo (MCMC), such as Hamiltonian Monte Carlo (HMC)~\cite{betancourt2017conceptual}, or by variational inference (VI)~\cite{blei2017variational}. These methods propagate uncertainty in both the inputs and the hyperparameters, but they are computationally intensive for large $n$ because each evaluation of the Gaussian likelihood involves a linear solve with $K_\theta(\tilde{X},\tilde{X})+\sigma^2 I_n$.

A pragmatic alternative is maximum a posteriori (MAP) estimation of $\tilde{X}$ together with the hyperparameters. Let $\mathbf{u}=(\mathbf{u}_1,\ldots,\mathbf{u}_n)$ collect the unconstrained variables and write $\tilde{X}(\mathbf{u})$ for the corresponding latent inputs via \eqref{eq:box-transform}. The negative log–posterior objective is
\begin{equation}
\label{eq:map-objective}
\begin{split}
\mathcal{J}(\mathbf{u},\theta,\sigma^2)
\,=& \frac{1}{2}\big(\mathbf{y}- \mu(\tilde{X}(\mathbf{u}))\big)^\top\!\!\Big[K_\theta\big(\tilde{X}(\mathbf{u}),\tilde{X}(\mathbf{u})\big)+\sigma^2 I_n\Big]^{-1}\!\big(\mathbf{y}-\mu(\tilde{X}(\mathbf{u}))\big)
\;+\; \\
&\frac{1}{2}\log\!\det\!\Big[K_\theta\big(\tilde{X}(\mathbf{u}),\tilde{X}(\mathbf{u})\big)+\sigma^2 I_n\Big]
\;+\; \frac{n}{2}\log(2\pi)
\;-\; \log p(\tilde{X}(\mathbf{u})) \\
& \;-\; \log p(\theta)
\;-\; \log p(\sigma^2).
\end{split}
\end{equation}
The first quadratic term in \eqref{eq:map-objective} measures data misfit under the GP prior, the log–determinant penalizes model complexity by the marginal predictive covariance volume at the training locations, and the final three terms add the contributions of the input and hyperparameter priors. A stable and effective optimization strategy alternates between two differentiable steps: with the latent inputs fixed, one maximizes the marginal likelihood over $(\theta,\sigma^2)$ using standard GP training; with $(\theta,\sigma^2)$ fixed, one takes a few gradient steps in $\mathbf{u}$ to reduce $\mathcal{J}$. The logistic transformation \eqref{eq:box-transform} ensures the box constraints are maintained throughout and allows gradients to flow to $\mathbf{u}$ by the chain rule. In practice it is numerically advantageous, once the alternating steps have converged, to rebuild the GP at the final $\tilde{X}$ and refit $(\theta,\sigma^2)$ before computing predictions.

\begin{figure}[htb!]
    \centering
    \includegraphics[width=1\linewidth]{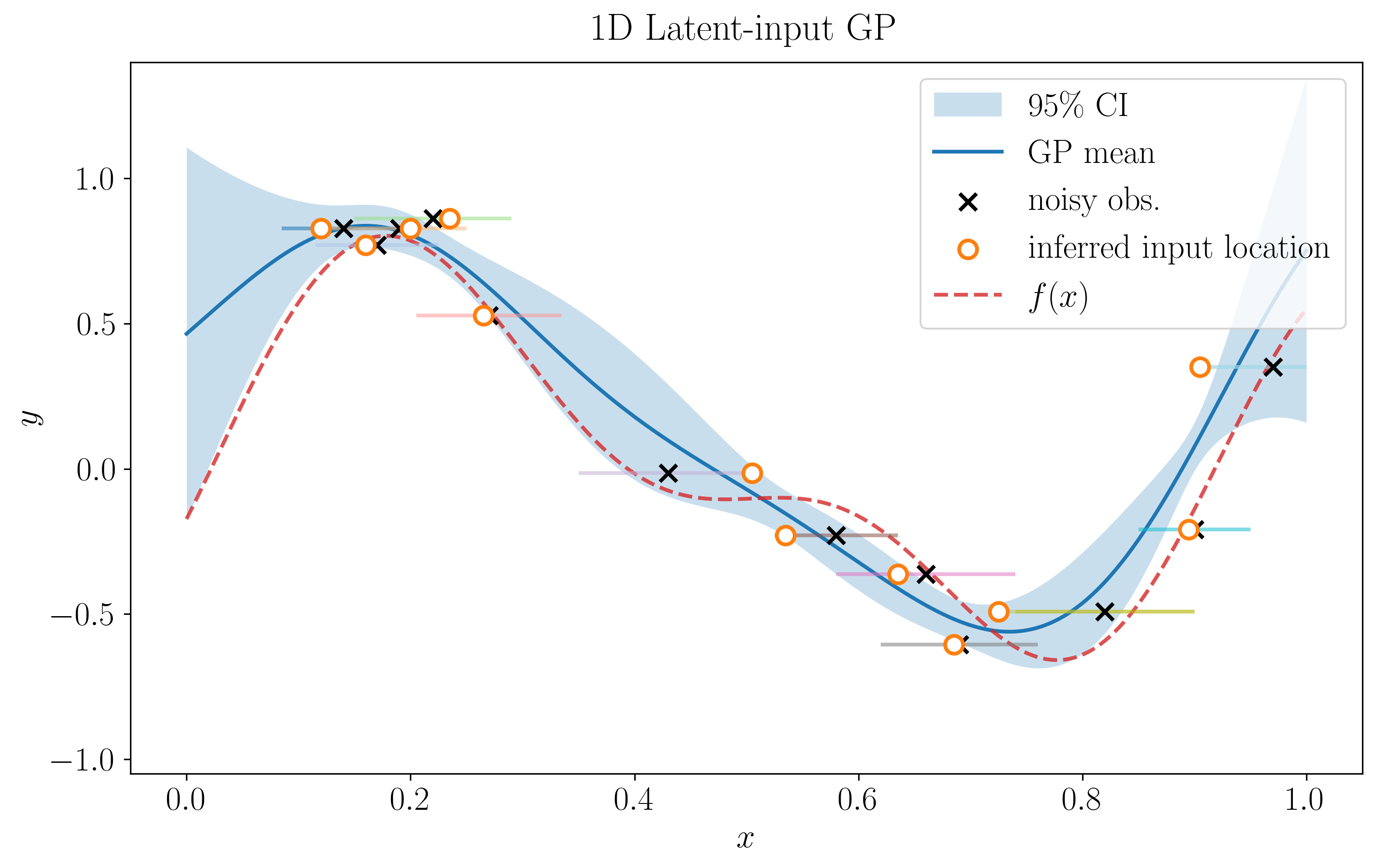}
    \caption{Illustration of the Bayesian latent GP approach for fitting data with epistemic uncertainty in inputs and outputs. The orange circles represent MAP estimates of the unknown input location $\tilde{\x}$; the horizontal lines represent the epistemic intervals that express prior uncertainty over $\x$.}
    \label{fig:latentgp}
\end{figure}

For a fixed test input $\x_\star\in\mathbb{R}^d$, the GP posterior is Gaussian. Writing $k_\star=K_\theta(\tilde{X},\x_\star)\in\mathbb{R}^{n}$ for the vector of covariances between the training inputs and the test input, and $k_{\star\star}=k_\theta(\x_\star,\x_\star)$ for the prior variance at the test input, the posterior mean and variance of $f(\x_\star)$ are
\begin{equation}
\label{eq:pred-point}
\begin{split}
\tilde \mu(\x_\star) \;=\; \mu(\x_\star) \;+\; k_\star^\top\!\Big[K_\theta(\tilde{X},\tilde{X})+\sigma^2 I_n\Big]^{-1}\!\big(\mathbf{y}-\mu(\tilde{X})\big),
\qquad \\
s^2(\x_\star) \;=\; k_{\star\star} \;-\; k_\star^\top\!\Big[K_\theta(\tilde{X},\tilde{X})+\sigma^2 I_n\Big]^{-1}\!k_\star.
\end{split}
\end{equation}
Each symbol in \eqref{eq:pred-point} is defined in terms of the learned latent inputs $\tilde{X}$ and the fitted hyperparameters. When the test input is itself epistemically uncertain, specified by a target box $\mathcal{U}_\star$ and an associated belief density $\pi_\star$ on $\mathcal{U}_\star$, one reports a prediction that is marginalized over this uncertainty. The law of total expectation and the law of total variance yield
\begin{equation}
\label{eq:pred-interval}
\bar{\mu}_\star \;=\; \mathbb{E}_{\x_\star\sim \pi_\star}\!\big[\tilde \mu(\x_\star)\big],
\qquad
\bar{s}^{\,2}_\star \;=\; \mathbb{E}_{\x_\star\sim \pi_\star}\!\big[s^2(\x_\star)\big]
\;+\; \mathrm{Var}_{\x_\star\sim \pi_\star}\!\big(\tilde \mu(\x_\star)\big),
\end{equation}
where the first term in $\bar{s}^{\,2}_\star$ represents the average conditional uncertainty of the GP at a fixed input and the second term quantifies the additional dispersion due to the epistemic spread of the input itself. When $\pi_\star$ collapses to a point mass at a known $\x_\star$, the second term vanishes, and \eqref{eq:pred-interval} reduces to \eqref{eq:pred-point}. The expectations in \eqref{eq:pred-interval} are efficiently approximated by low–order tensor product Gaussian quadrature in low dimensions, by sparse grids in moderate dimensions, or by stratified Monte Carlo sampling when a non–uniform $\pi_\star$ is required. In our work, we draw iid samples from $\pi_\star$ and take a Monte Carlo average to compute the expectation and variance in \eqref{eq:pred-interval}. We show an illustration of the latent GP on a simple $1$D function in \Cref{fig:latentgp} -- the horizontal lines indicate the input epistemic intervals and the orange circle is the MAP estimate of the input.

The latent-input construction seamlessly reduces to standard GP regression in the limit that each box $\mathcal{U}_i$ shrinks to a singleton. For wider boxes there is an intuitive confounding between large kernel lengthscales and diffuse input locations; this is mitigated by weakly informative priors on the lengthscales (for example, log-normal priors centered on the physical scales of the inputs) and by modest shrinkage of $\tilde{\x}_i$ toward the box midpoints through the prior $p(\tilde{X})$. 

The latent GP provides a MAP estimate of $\tilde{\x}$, which is then used to fit standard GPs at the MAP estimates of the inputs for downstream UQ. Unlike existing approaches that calibrate with the moments of the given truth data, we calibrate with the full distribution of the given data, as explained next.

\subsection{Distributional calibration of the correction Gaussian process}
\label{sec:distributional_calibration}
The purpose of the distributional calibration method is
to use the available truth-data bounds not only to correct the posterior mean,
but also to calibrate the posterior uncertainty of the corrected model.  In
particular, at each truth-data input location (that is, the MAP estimates from the latent GP), the calibrated corrected GP is
constructed so that $\mu_{\mathrm{cal}}(\x_i)=c_i$ and
$\mu_{\mathrm{cal}}(\x_i)\pm z_{0.975}\sigma_{\mathrm{cal}}(\x_i)=[L_i,U_i]$,
where $[L_i,U_i]$ is the given epistemic truth interval, $c_i$ is its centroid, and
$z_{0.975}=1.959963984\ldots$ is the standard normal quantile associated with a
two-sided pointwise 95\% Gaussian interval.  The same construction is applied
separately to each scalar aerodynamic response: $C_d$, $C_l$, and $C_m$.

For convenience, we denote the MAP estimate of $\tilde{\x}$ as $\x^T$ and introduce the following notation. Let the XFOIL training data be
$\mathcal{D}_X=\{(\x^{X}_j,y^{X}_j)\}_{j=1}^{N_X}$, and let the truth-data
bounds be $\mathcal{D}_T=\{(\x^{T}_i,\,[L_i,U_i])\}_{i=1}^{N_T}$.  Define the
matrix of truth-data input locations as $X_T=[\x^{T}_1,\ldots,\x^{T}_{N_T}]^\top$.
For each truth-data interval, define the centroid and half-width as
$c_i=(L_i+U_i)/2$ and $h_i=(U_i-L_i)/2$.
As previously mentioned, we treat the given epistemic intervals as 95\% Gaussian
credible intervals; then, the target marginal variance for the composite GP at
$\x_i^T$ is
\begin{equation*}
    v_i^\star
    =
    \left(
    \frac{h_i}{z_{0.975}}
    \right)^2,
    \label{eq:target_variance_95}
\end{equation*}
which follows from the fact that the half-width $h_i = z_{0.975} \sigma_{\T{cal}}$. More generally, for a two-sided pointwise probability level $p$,
\begin{equation*}
    z_p
    =
    \Phi^{-1}
    \left(
    \frac{1+p}{2}
    \right),
    \qquad
    v_i^\star
    =
    \left(
    \frac{h_i}{z_p}
    \right)^2 ,
\end{equation*}
where $\Phi^{-1}$ is the inverse standard normal cumulative distribution
function. Stack the truth centroids into $\mathbf{c}
    =
    \left[
    c_1,\ldots,c_{N_T}
    \right]^\top,$ and denote the target covariance matrix at the truth-data locations by
$\mathbf{V}_T$.  The simplest choice is a diagonal target covariance,
\begin{equation}
    \mathbf{V}_T
    =
    \mathrm{diag}
    \left(
    v_1^\star,\ldots,v_{N_T}^\star
    \right),
    \label{eq:diagonal_target_covariance}
\end{equation}
which assumes independence of the individual truth measurements. 
Because the composite GP is an additive combination of two GPs, the resulting composite GP
$g_0(\x) = f_X(\x) + \delta_0(\x)$ is also a GP under the mild assumption that
$f_X$ and $\delta_0$ are independent. Then,
\begin{equation*}
    g_0(\x)
    \sim
    \mathcal{GP}
    \left(
    m_0(\x),
    k_0(\x,\x')
    \right),
\end{equation*}
with
\begin{equation*}
    m_0(\x)
    =
    \mu_X(\x)
    +
    m_\delta(\x),
    \qquad
    k_0(\x,\x')
    =
    k_X(\x,\x')
    +
    k_\delta(\x,\x').
\end{equation*}
%

A ``first-moment'' calibration uses the centroid mismatch
\begin{equation*}
    d_i
    =
    c_i
    -
    \mu_X(\x_i^T)
\end{equation*}
as training data for the correction GP.  This enforces a mean correction but
does not use the interval width $U_i-L_i$.  The distributional calibration
method instead treats the truth bounds as a target marginal distribution for the
corrected GP at the truth-data input locations.

Let
\begin{equation*}
    \mathbf{g}_T
    =
    g_0(X_T)
    =
    \left[
    g_0(\x_1^T),\ldots,g_0(\x_{N_T}^T)
    \right]^\top .
\end{equation*}
Under the uncalibrated corrected GP,
\begin{equation*}
    \mathbf{g}_T
    \sim
    \mathcal{N}
    \left(
    \mathbf{m}_{0T},
    \mathbf{K}_{TT}
    \right),~\T{where} ~ \mathbf{m}_{0T}
    =
    m_0(X_T),
    \qquad
    \mathbf{K}_{TT}
    =
    k_0(X_T,X_T).
\end{equation*}
The idea is to replace $\mathbf{g}_T \sim \mathcal{N}
    \left(
    \mathbf{m}_{0T},
    \mathbf{K}_{TT}
    \right)$ with the distributional
    calibration target:
\begin{equation}
    \mathbf{g}_T
    \sim
    \mathcal{N}
    \left(
    \mathbf{c},
    \mathbf{V}_T
    \right).
    \label{eqn:target_truth_marginal}
\end{equation}
The overall calibrated distribution is defined by replacing the prior finite-dimensional
marginal distribution of $\mathbf{g}_T$ with the target distribution in
\Cref{eqn:target_truth_marginal}, while preserving the original conditional
distribution of the process away from $X_T$.  Equivalently,
\begin{equation*}
    p_{\mathrm{cal}}
    \left(
    g \mid \mathcal{D}_X,\mathcal{D}_T
    \right)
    =
    \int
    p_0
    \left(
    g \mid \mathbf{g}_T=\mathbf{z},\mathcal{D}_X
    \right)
    q_T(\mathbf{z})
    \,
    d\mathbf{z},
\end{equation*}
where
\begin{equation*}
    q_T(\mathbf{z})
    =
    \mathcal{N}
    \left(
    \mathbf{z};
    \mathbf{c},
    \mathbf{V}_T
    \right).
\end{equation*}
That is, we first draw a sample $\mbf{z}$ from the target truth distribution $\mcl{N}(\mbf{c}, \mbf{V}_T)$. Then, conditional on $\mathbf{g}_T=\mbf{z}$, we draw the rest of the function using the original GP conditional distribution. This changes the marginal distribution at $\mbf{X}_T$, but leaves the original spatial covariance structure intact away from those points. Therefore, crucially, we calibrate our surrogate model with both the centroid and epistemic intervals of the given $7$ truth points.


\subsubsection{Closed-form calibrated mean and covariance}
\label{subsec:closed_form_calibration}

The calibrated composite GP has closed form. For any prediction input $\x$, define
$\mathbf{k}_{\x T}=k_0(\x,X_T)=\left[k_0(\x,\x_1^T),\ldots,k_0(\x,\x_{N_T}^T)\right]$.
Similarly, $\mathbf{k}_{T\x'}=k_0(X_T,\x')$.
Since the covariance matrix $\mathbf{K}_{TT}$ is positive definite, the inverse
below is evaluated through a Cholesky solve. The calibrated corrected GP is
\begin{equation*}
    g_{\mathrm{cal}}(\x)
    \sim
    \mathcal{GP}
    \left(
    m_{\mathrm{cal}}(\x),
    k_{\mathrm{cal}}(\x,\x')
    \right),
\end{equation*}
with mean
\begin{equation}
    m_{\mathrm{cal}}(\x)
    =
    m_0(\x)
    +
    \mathbf{k}_{\x T}
    \mathbf{K}_{TT}^{-1}
    \left(
    \mathbf{c}
    -
    \mathbf{m}_{0T}
    \right),
    \label{eq:calibrated_mean}
\end{equation}
and covariance
\[
    k_{\mathrm{cal}}(\x,\x')
    =
    k_0(\x,\x')
    -
    \mathbf{k}_{\x T}
    \mathbf{K}_{TT}^{-1}
    \mathbf{k}_{T\x'}
    +
    \mathbf{k}_{\x T}
    \mathbf{K}_{TT}^{-1}
    \mathbf{V}_T
    \mathbf{K}_{TT}^{-1}
    \mathbf{k}_{T\x'} .
\]
For a finite prediction set $X_\star=[\x_1^\star,\ldots,\x_{N_\star}^\star]^\top$,
the corresponding matrix form is
\begin{equation*}
    \mathbf{m}_{\mathrm{cal},\star}
    =
    \mathbf{m}_{0,\star}
    +
    \mathbf{K}_{\star T}
    \mathbf{K}_{TT}^{-1}
    \left(
    \mathbf{c}
    -
    \mathbf{m}_{0T}
    \right),
\end{equation*}
and
\[
    \mathbf{K}_{\mathrm{cal},\star\star}
    =
    \mathbf{K}_{\star\star}
    -
    \mathbf{K}_{\star T}
    \mathbf{K}_{TT}^{-1}
    \mathbf{K}_{T\star}
    +
    \mathbf{K}_{\star T}
    \mathbf{K}_{TT}^{-1}
    \mathbf{V}_T
    \mathbf{K}_{TT}^{-1}
    \mathbf{K}_{T\star}.
    \label{eq:calibrated_covariance_matrix}
\]
We provide details of the derivation in \Cref{sec:derivation}.

\subsubsection{Verification of the calibration conditions}
\label{subsec:calibration_verification}

Evaluating Eq.~\eqref{eq:calibrated_mean} at the truth-data locations gives
\begin{align*}
    m_{\mathrm{cal}}(X_T)
    &=
    \mathbf{m}_{0T}
    +
    \mathbf{K}_{TT}
    \mathbf{K}_{TT}^{-1}
    \left(
    \mathbf{c}
    -
    \mathbf{m}_{0T}
    \right)
    \nonumber = \mathbf{c}.
\end{align*}
Thus the calibrated corrected GP posterior mean exactly matches the centroid of
the truth-data epistemic interval at every truth-data input location.
Similarly, evaluating calibrated variance at
$X_\star=X_T$ gives
\begin{align*}
    \mathbf{K}_{\mathrm{cal}}(X_T,X_T)
    &=
    \mathbf{K}_{TT}
    -
    \mathbf{K}_{TT}
    \mathbf{K}_{TT}^{-1}
    \mathbf{K}_{TT}
    +
    \mathbf{K}_{TT}
    \mathbf{K}_{TT}^{-1}
    \mathbf{V}_T
    \mathbf{K}_{TT}^{-1}
    \mathbf{K}_{TT}
    \nonumber = \mathbf{V}_T.
\end{align*}
Therefore, for the diagonal target covariance in
Eq.~\eqref{eq:diagonal_target_covariance},
\begin{equation*}
    \sigma_{\mathrm{cal}}^2(\x_i^T)
    =
    v_i^\star
    =
    \left(
    \frac{U_i-L_i}{2z_{0.975}}
    \right)^2.
\end{equation*}
It follows that
\begin{align*}
    m_{\mathrm{cal}}(\x_i^T)
    -
    z_{0.975}
    \sigma_{\mathrm{cal}}(\x_i^T)
    &=
    c_i
    -
    z_{0.975}
    \frac{h_i}{z_{0.975}}
    =
    c_i-h_i
    =
    L_i,
    \\
    m_{\mathrm{cal}}(\x_i^T)
    +
    z_{0.975}
    \sigma_{\mathrm{cal}}(\x_i^T)
    &=
    c_i
    +
    z_{0.975}
    \frac{h_i}{z_{0.975}}
    =
    c_i+h_i
    =
    U_i.
\end{align*}
Hence the pointwise 95\% Gaussian bounds of the calibrated corrected GP coincide
exactly with the epistemic truth bounds at the truth-data locations. The full methodology is summarized in \Cref{alg:latent_dc_mc}.

\subsection{Monte Carlo simulation to propagate uncertainty}
After distributional calibration, each scalar aerodynamic coefficient
$q \in \{C_d,C_l,C_m\}$ is represented by a calibrated corrected Gaussian
process
$g_{q,\mathrm{cal}}(\mathbf{x}) \sim
\mathcal{GP}\!\left(m_{q,\mathrm{cal}}(\mathbf{x}),
k_{q,\mathrm{cal}}(\mathbf{x},\mathbf{x}')\right)$. For notational compactness, write
$m_q(\mathbf{x})=m_{q,\mathrm{cal}}(\mathbf{x})$ and
$v_q(\mathbf{x})=k_{q,\mathrm{cal}}(\mathbf{x},\mathbf{x})$. At a fixed input
condition, the calibrated predictive distribution is therefore
$q_\star \mid \mathbf{x}_\star,\mathcal{D}_X,\mathcal{D}_T
\sim \mathcal{N}\!\left(m_q(\mathbf{x}_\star),v_q(\mathbf{x}_\star)\right)$.

For the $4$ given prediction points of interest, recall that we have an epistemic uncertainty interval, to which we assign a belief density $\pi_\star$ supported on
that interval. Propagating this input uncertainty through the calibrated GP gives
the marginal predictive distribution
\[
p(q_\star \mid \mathcal{D}_X,\mathcal{D}_T,U_\star)
=
\int_{\x \in \mcl{U}_\star}
\mathcal{N}\!\left(q_\star;\,m_q(\mathbf{x}),v_q(\mathbf{x})\right)
\pi_\star(\mathbf{x})\,\mathrm{d}\mathbf{x}.
\]
Although the conditional prediction at any fixed input is Gaussian, this
marginal distribution is generally non-Gaussian because both the
mean and variance of the calibrated GP vary over the uncertain input region.

To estimate the full predictive distribution of the quantities of interest, we draw
independent samples $\mathbf{x}_\star^{(r)}\sim\pi_\star$, $r=1,\ldots,N_{\rm MC}$,
and evaluate the calibrated GP at each sample. Then,  we additionally draw $\xi^{(r)}\sim\mathcal{N}(0,1)$ and form
posterior predictive samples as
\[
q_\star^{(r)}
=
m_q^{(r)}+\sqrt{v_q^{(r)}}\,\xi^{(r)}.
\]
The empirical distribution of $\{q_\star^{(r)}\}_{r=1}^{N_{\rm MC}}$ is then
used to construct predictive densities, cumulative distribution functions, and
credible intervals. For example, the reported $95\%$ uncertainty interval is
taken as the empirical $2.5\%$ and $97.5\%$ quantiles of these samples. The same
procedure is applied independently to the calibrated GP models for $C_d$,
$C_l$, and $C_m$.

%

\begin{algorithm}[t]
\footnotesize
\caption{Latent input distributional calibration and Monte Carlo propagation}
\label{alg:latent_dc_mc}
\begin{algorithmic}[1]
\Require XFOIL data $\mathcal{D}_X^q= \{ ( \mathbf{x}_{j}^{X},y_{j}^{X,q} ) \}_{j=1}^{N_X}$ and truth data
$\mathcal{D}_T^q=\{(\mathcal{U}_i^T,[L_i^q,U_i^q])\}_{i=1}^{N_T}$ for
$q\in\mathcal{Q}=\{C_d,C_l,C_m\}$; target input box $\mathcal{U}_\star$ with belief density
$\pi_\star$; level $p=0.95$; sample size $N_{\rm MC}$; jitter $\eta>0$.
\Ensure Predicted empirical PDFs, CDFs, and interval mass for each $q$.
\For{$q\in\mathcal{Q}$}
    \State Fit the XFOIL surrogate
    $f_X^q(\mathbf{x})\mid\mathcal{D}_X^q\sim
    \mathcal{GP}(\mu_X^q(\mathbf{x}),k_X^q(\mathbf{x},\mathbf{x}'))$.
    \State Infer latent truth inputs by MAP: $\mathbf{X}_{T,q}=[\widetilde{\mathbf{x}}_{1,q},\ldots,
    \widetilde{\mathbf{x}}_{N_T,q}]^\top$.
    \State Choose an uncalibrated discrepancy GP
    $\delta_0^q(\mathbf{x})\sim
    \mathcal{GP}(m_\delta^q(\mathbf{x}),k_\delta^q(\mathbf{x},\mathbf{x}'))$.
    \State Convert output intervals to target moments:
    $c_i^q=(L_i^q+U_i^q)/2$,
    $h_i^q=(U_i^q-L_i^q)/2$,
    $v_i^{\star,q}=(h_i^q/z_p)^2$, with
    $z_p=\Phi^{-1}((1+p)/2)$. Set
    $\mathbf{c}_q=(c_1^q,\ldots,c_{N_T}^q)^\top$ and
    $\mathbf{V}_{T,q}=\operatorname{diag}(v_1^{\star,q},\ldots,v_{N_T}^{\star,q})$.
    \State Compute $\mathbf{m}_{0T}^q=m_0^q(\mathbf{X}_{T,q})$ and
    $\widetilde{\mathbf{K}}_{TT}^q
    =k_0^q(\mathbf{X}_{T,q},\mathbf{X}_{T,q})+\eta\mathbf{I}$.
    Solve
    $\widetilde{\mathbf{K}}_{TT}^q\boldsymbol{\alpha}_q
    =\mathbf{c}_q-\mathbf{m}_{0T}^q$.
    \For{$r=1,\ldots,N_{\rm MC}$}
        \State Draw $\mathbf{x}_\star^{(r)}\sim\pi_\star$ on $\mathcal{U}_\star$ and
        $\xi^{(r)}\sim\mathcal{N}(0,1)$.
        \State Set $\mathbf{k}_{rT}^q
        =k_0^q(\mathbf{x}_\star^{(r)},\mathbf{X}_{T,q})$ and solve
        $\widetilde{\mathbf{K}}_{TT}^q\mathbf{b}_r^q=(\mathbf{k}_{rT}^q)^\top$.
        \State Evaluate the calibrated predictive moments
        $m_r^q=m_0^q(\mathbf{x}_\star^{(r)})+\mathbf{k}_{rT}^q\boldsymbol{\alpha}_q$ and
        $v_r^q=k_0^q(\mathbf{x}_\star^{(r)},\mathbf{x}_\star^{(r)})
        -\mathbf{k}_{rT}^q\mathbf{b}_r^q
        +(\mathbf{b}_r^q)^\top\mathbf{V}_{T,q}\mathbf{b}_r^q$.
        \State Draw the posterior predictive sample
        $q_\star^{(r)}=m_r^q+\sqrt{\max(v_r^q,0)}\,\xi^{(r)}$.
    \EndFor
    \State Report empirical PDF, CDF, and interval mass.
\EndFor
\end{algorithmic}
\end{algorithm}

\section{Experiments}
\label{sec:experiments}
Our primary goal with the experiments is to estimate the uncertainty at the $4$ prediction points (orange circles in \Cref{fig:aiaa_graph}) in the form of probability density functions and cumulative distribution functions. The true uncertainties at these points were reported in \citep[Figs. 2--5]{cary2026summary} as epistemic bounds. We treat these bounds as $95\%$ confidence bounds, consistent with how we treated the bounds on the given truth data. The XFOIL GP is trained on $100$ Latin hypercube designs and then calibrated using the given $7$ truth points. Once calibrated, the uncertainty estimates are obtained from $10,000$ samples drawn uniformly at random from the input uncertainty bounds and propagated through the calibrated model as described in the previous section.

\subsection{Model validation}
\label{sec:model_validation}
\begin{figure}[htb!]
    \centering
    \begin{subfigure}{0.33\textwidth}
        \includegraphics[width=1\linewidth]{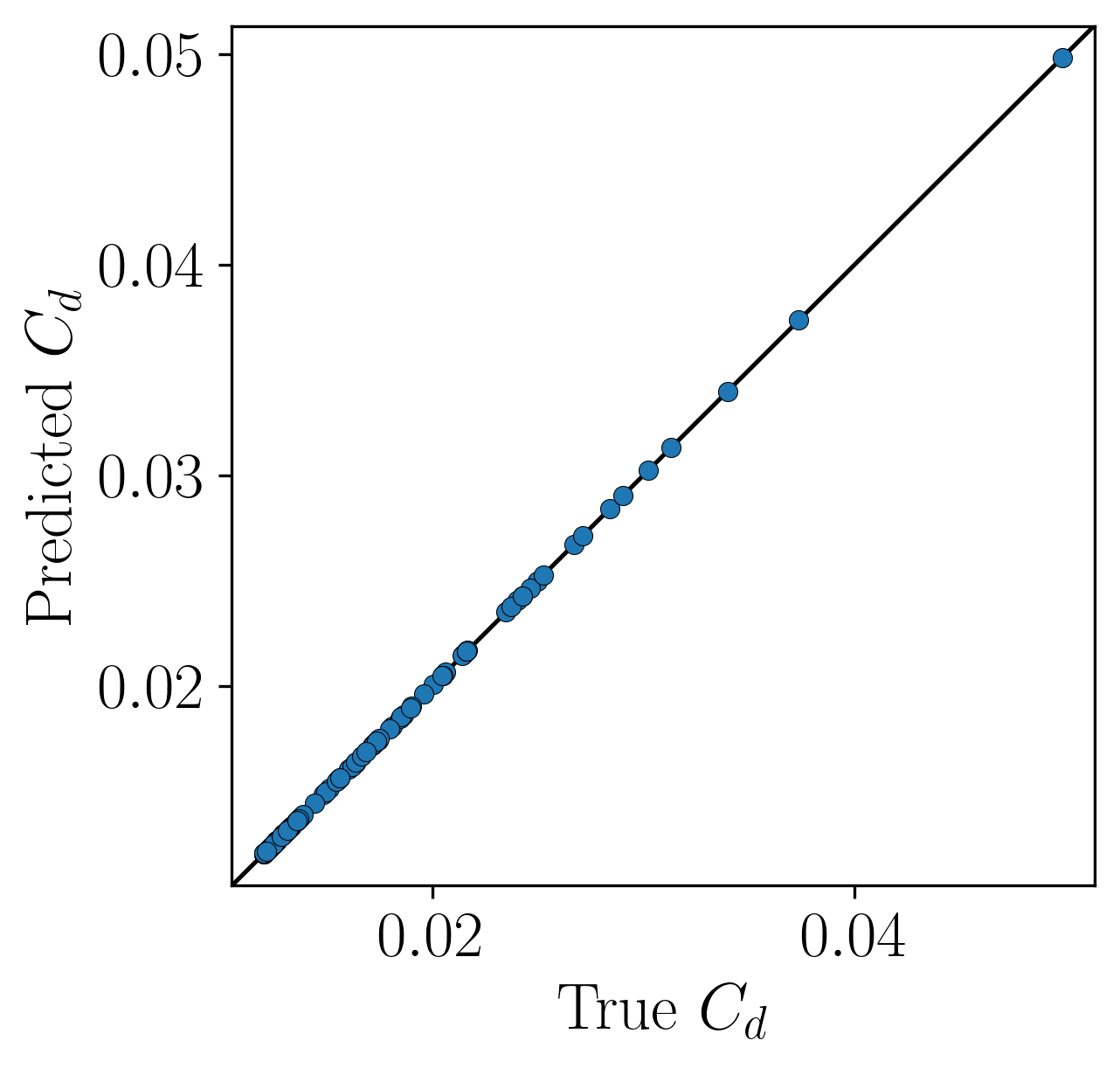}    
    \end{subfigure}%
    \begin{subfigure}{0.33\textwidth}
        \includegraphics[width=1\linewidth]{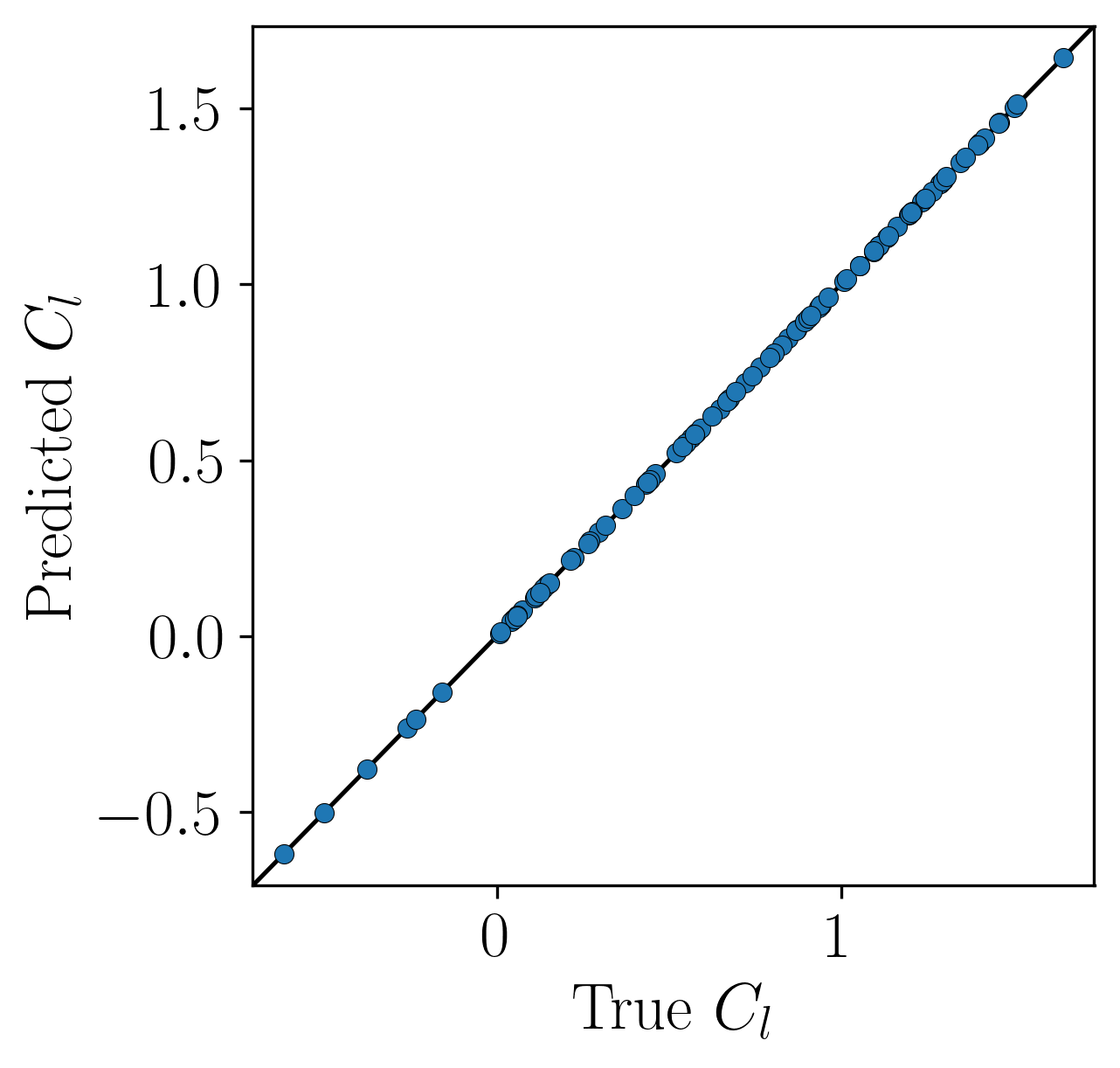} 
    \end{subfigure}%
    \begin{subfigure}{0.33\textwidth}
        \includegraphics[width=1\linewidth]{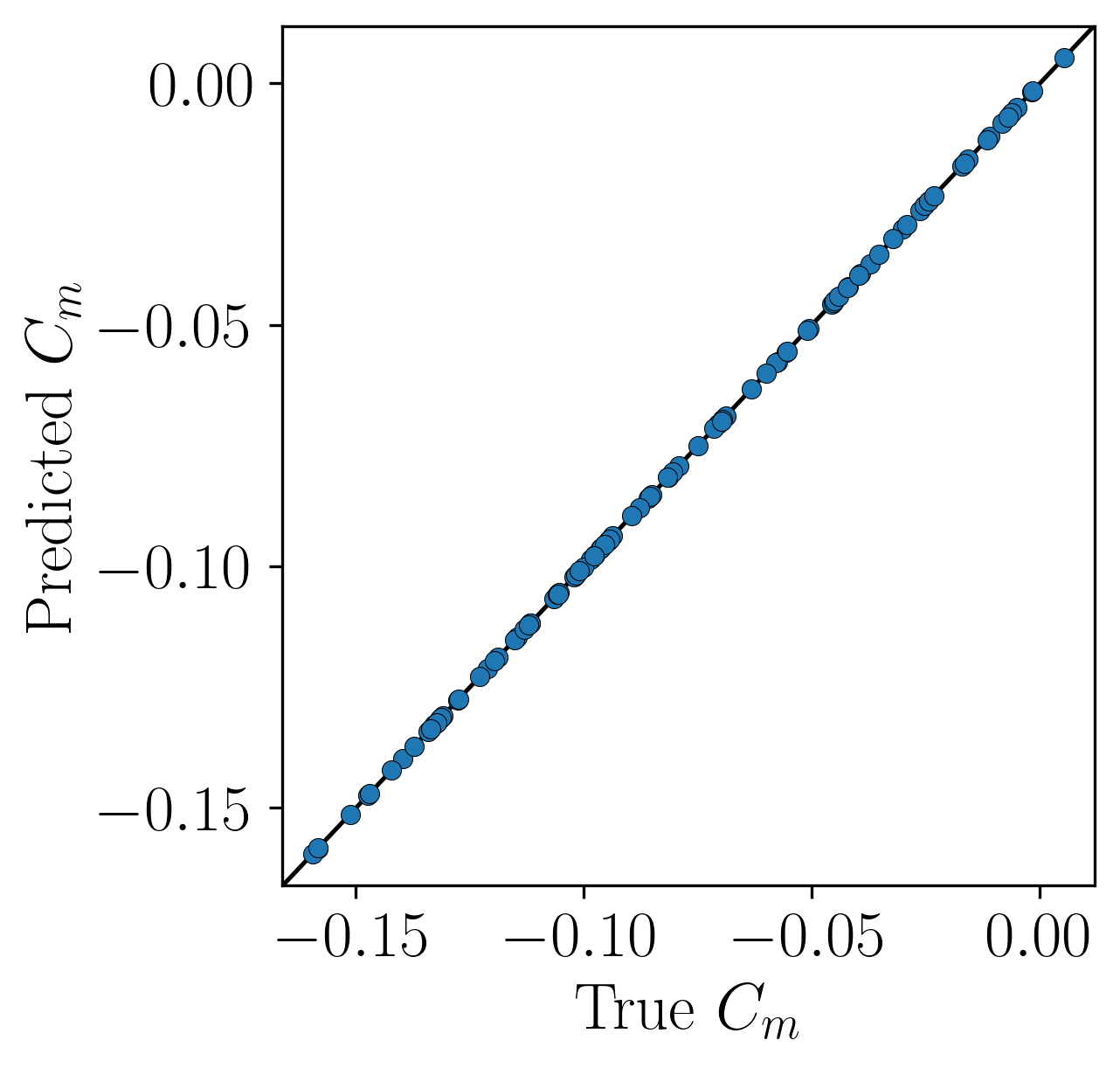}    
    \end{subfigure}\\
\begin{subfigure}{0.33\textwidth}
        \includegraphics[width=1\linewidth]{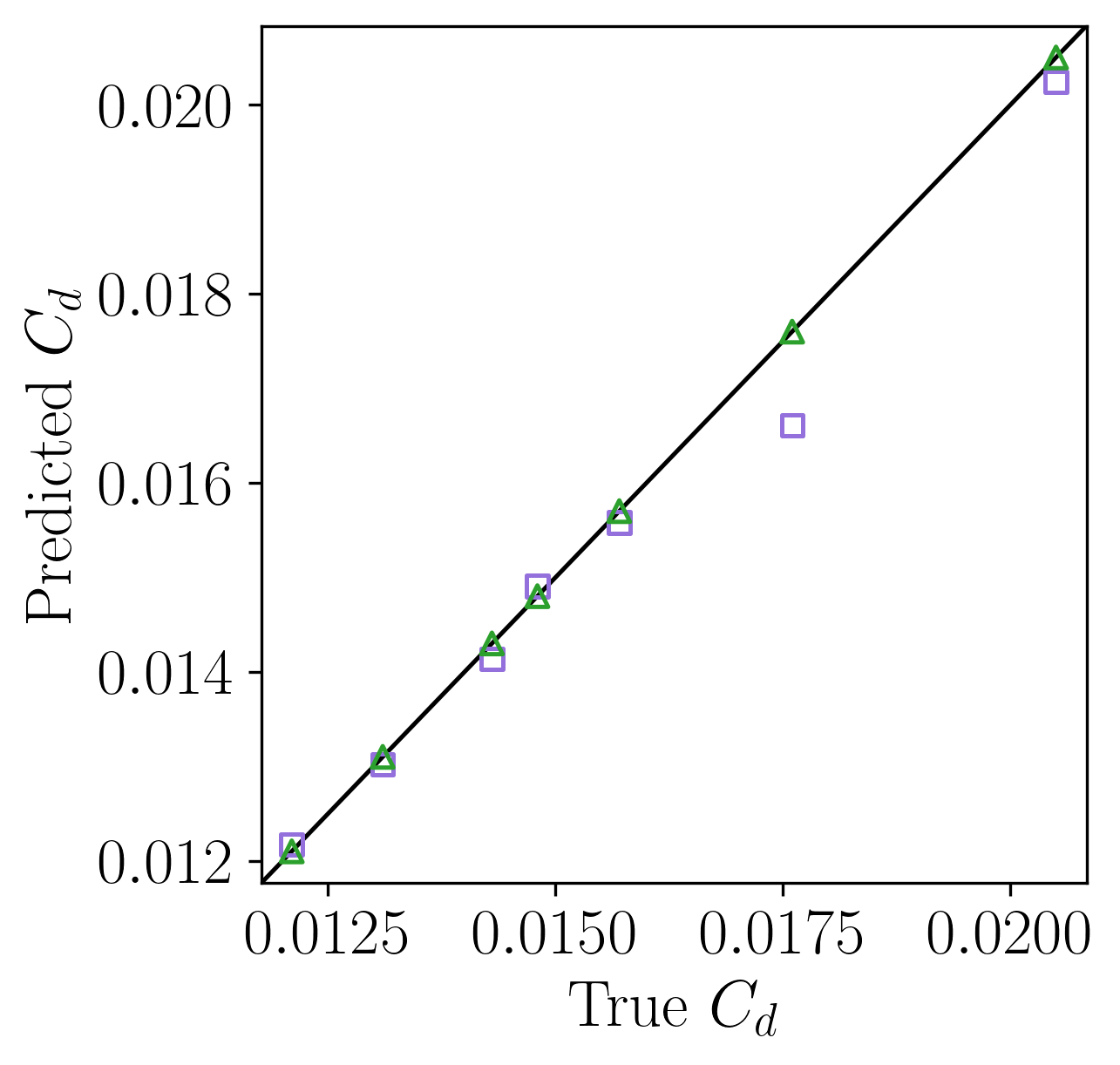}    
    \end{subfigure}%
    \begin{subfigure}{0.33\textwidth}
        \includegraphics[width=1\linewidth]{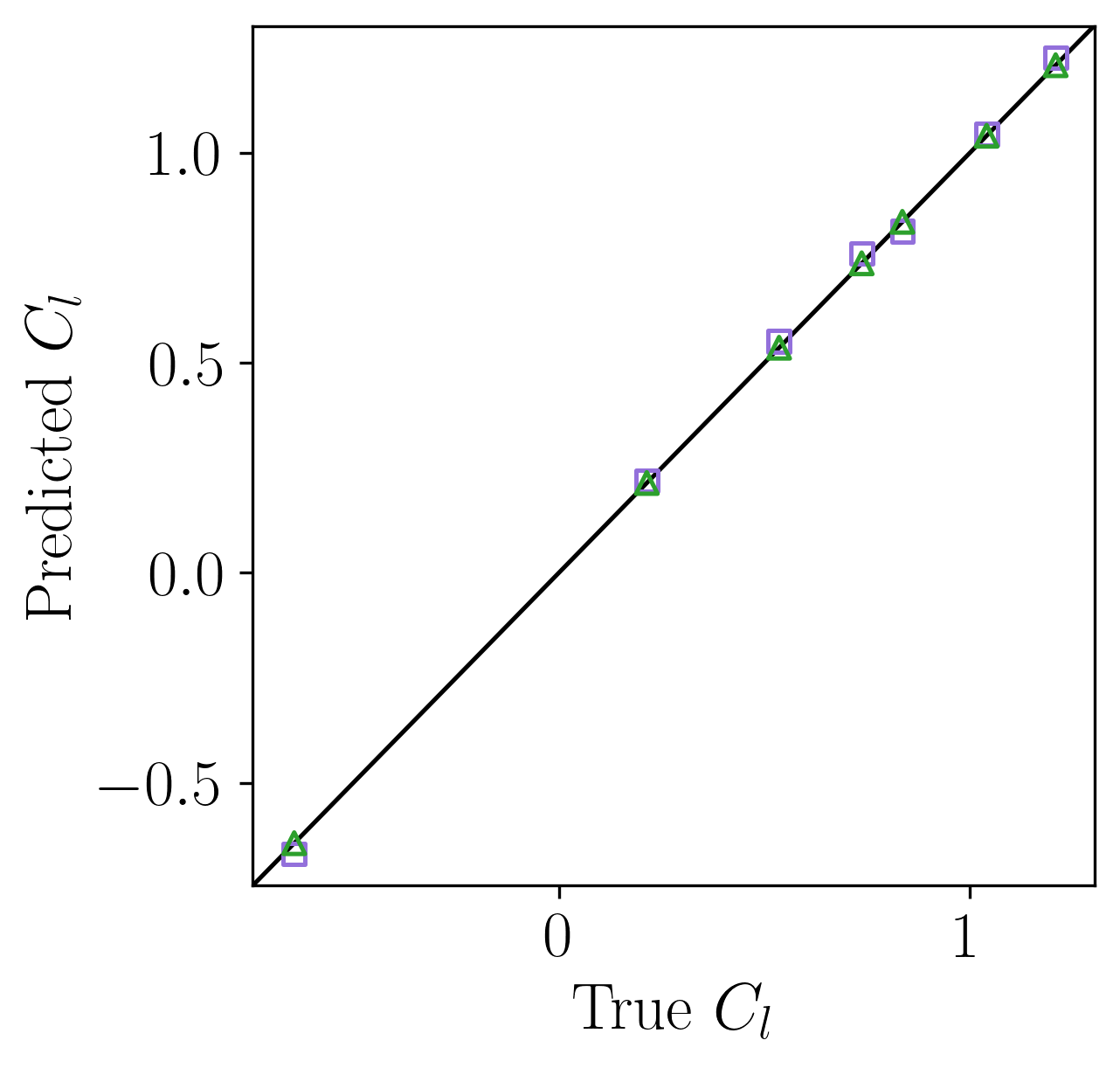}    
    \end{subfigure}%
    \begin{subfigure}{0.33\textwidth}
        \includegraphics[width=1\linewidth]{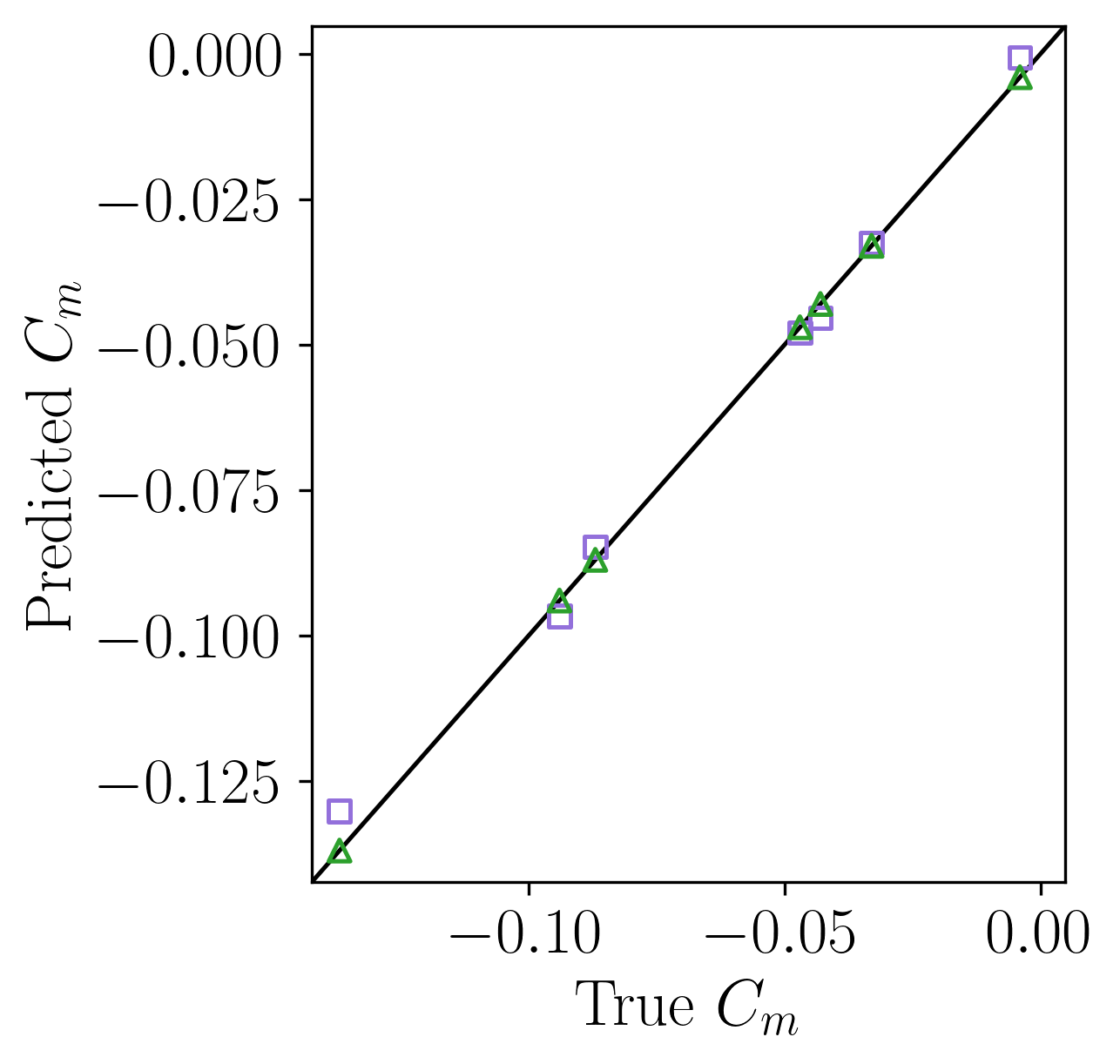}    
    \end{subfigure}
    
    \caption{Top row: validation of our XFOIL-data-fit GP on a held-out test set. Bottom row: validation of our composite GP on the $7$ data points in \Cref{tab:given_data}; the squares and triangles denote the vanilla and latent GP results, respectively.}
    \label{fig:xfoil_validation}
\end{figure}

We validate our GP surrogates in two stages. In the first stage, we validate the XFOIL-data-fit GP posterior mean predictions against a held-out set of $100$ XFOIL data points; this is shown in the top row of \Cref{fig:xfoil_validation}. In the second stage, the composite GP posterior mean predictions are validated against the $7$ validation points; this is included as the bottom row of \Cref{fig:xfoil_validation}. This figure establishes accuracy in both the XFOIL GP and the overall calibrated GP we fit.

\subsection{UQ results}
\label{sec:uq_results}

\begin{figure}[p]
    \centering
    \begin{subfigure}{.49\textwidth}
                \caption*{First-moment calibrated.}
        \centering
        \includegraphics[width=\linewidth]{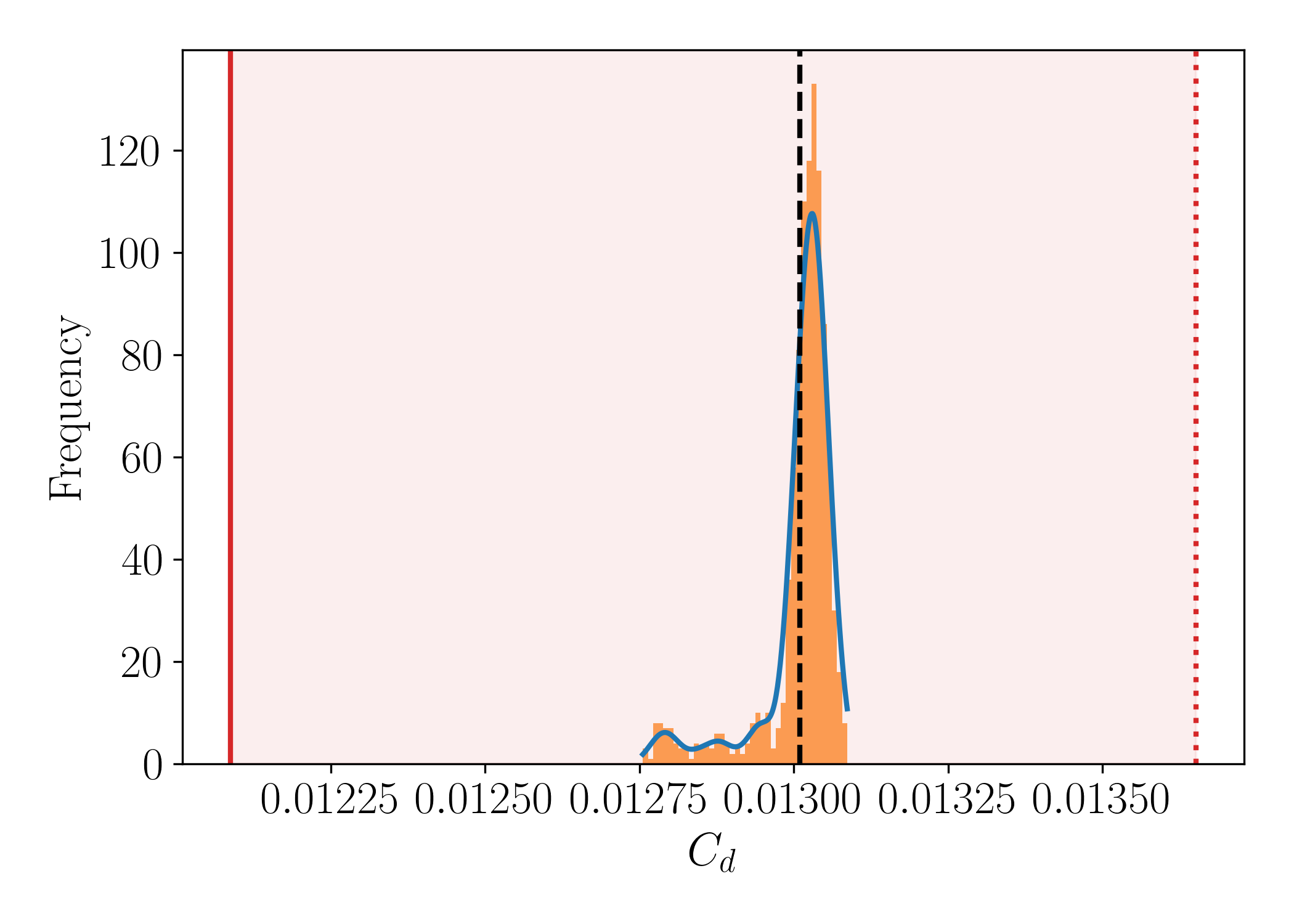}
    \end{subfigure}%
    \begin{subfigure}{.49\textwidth}
                \caption*{Distribution-calibrated.}
        \centering
        \includegraphics[width=\linewidth]{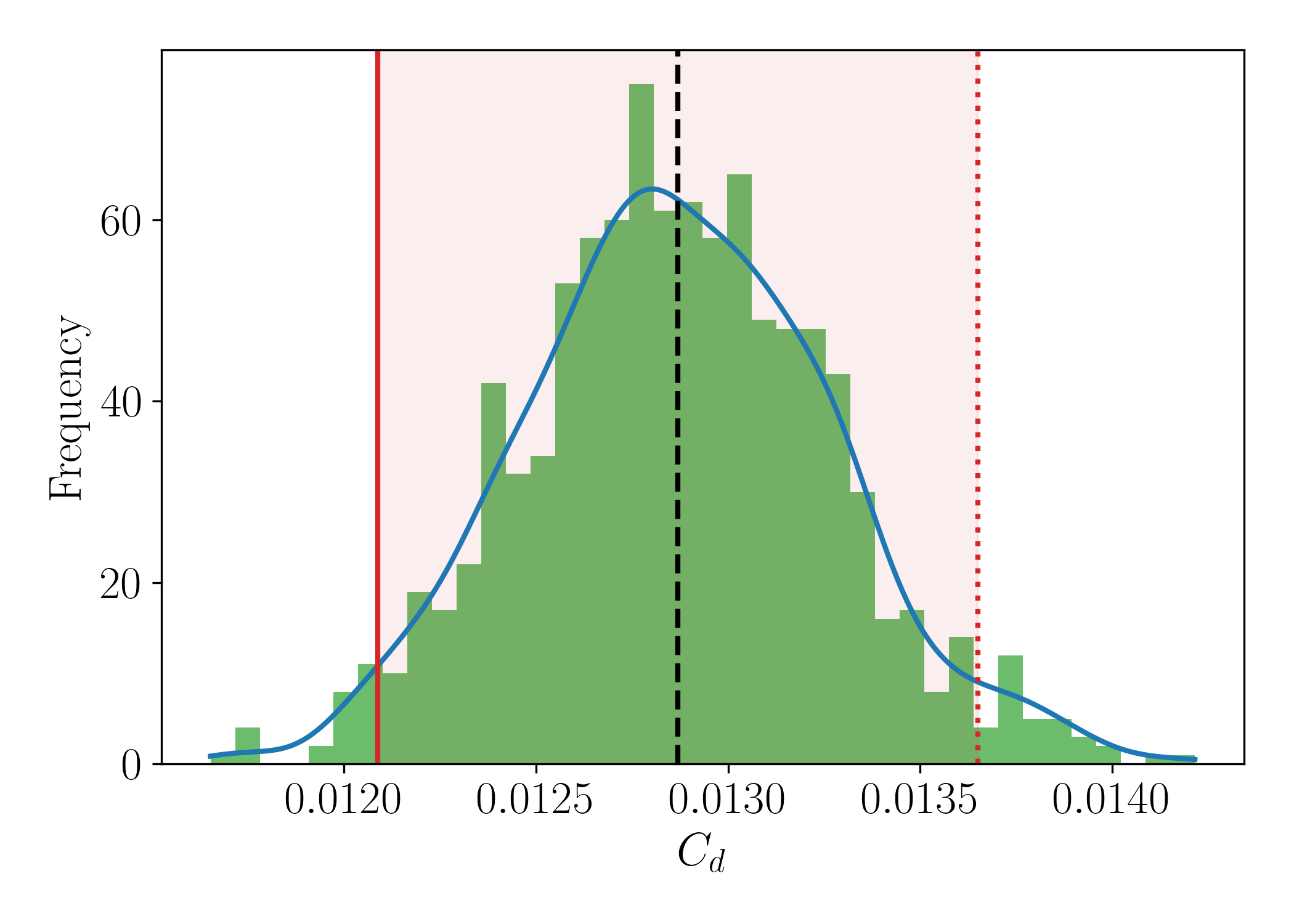}
    \end{subfigure}\\
    \caption*{$C_d$}
    \begin{subfigure}{.49\textwidth}
        \centering
        \includegraphics[width=\linewidth]{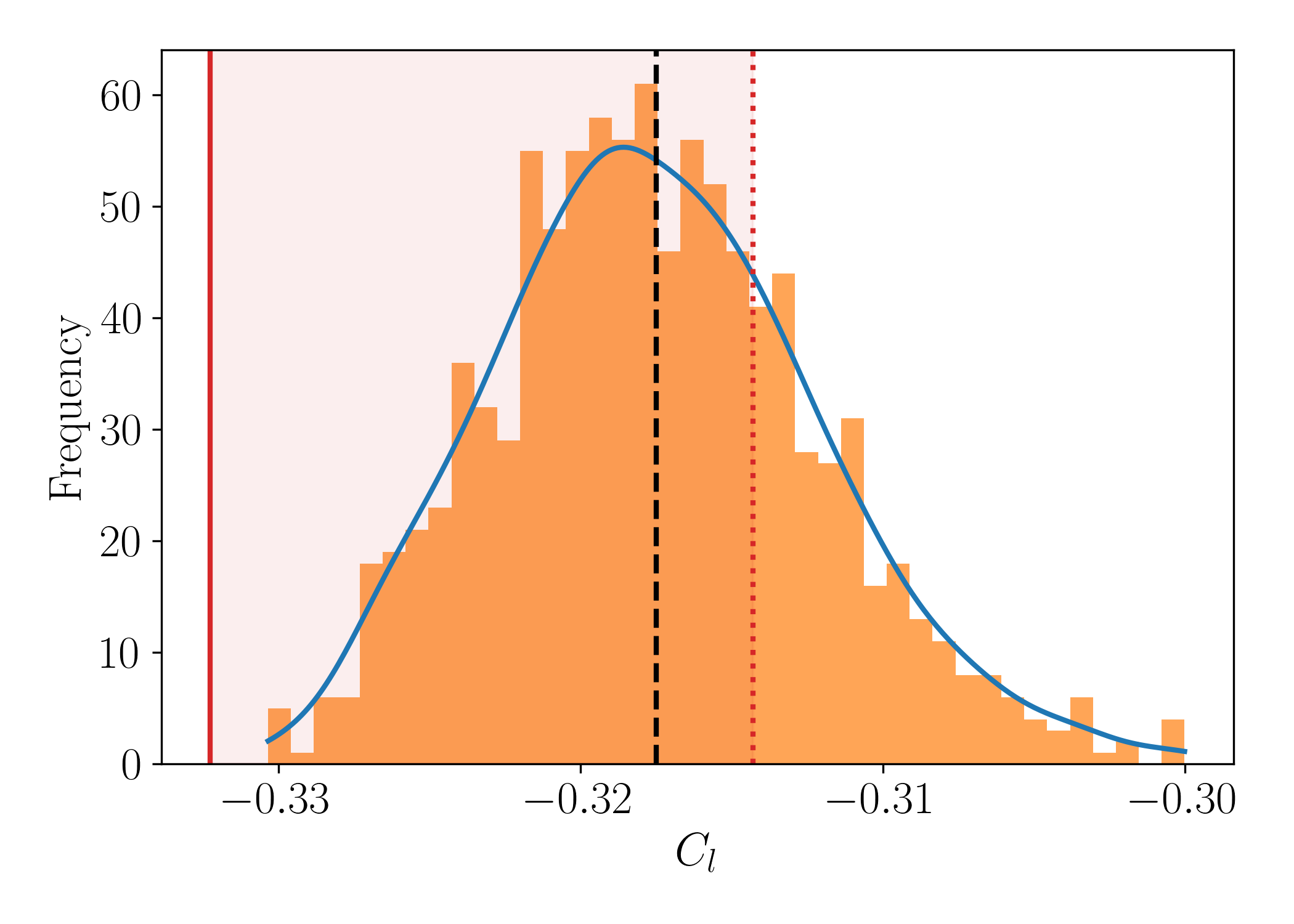}
    \end{subfigure}%
    \begin{subfigure}{.49\textwidth}
        \centering
        \includegraphics[width=\linewidth]{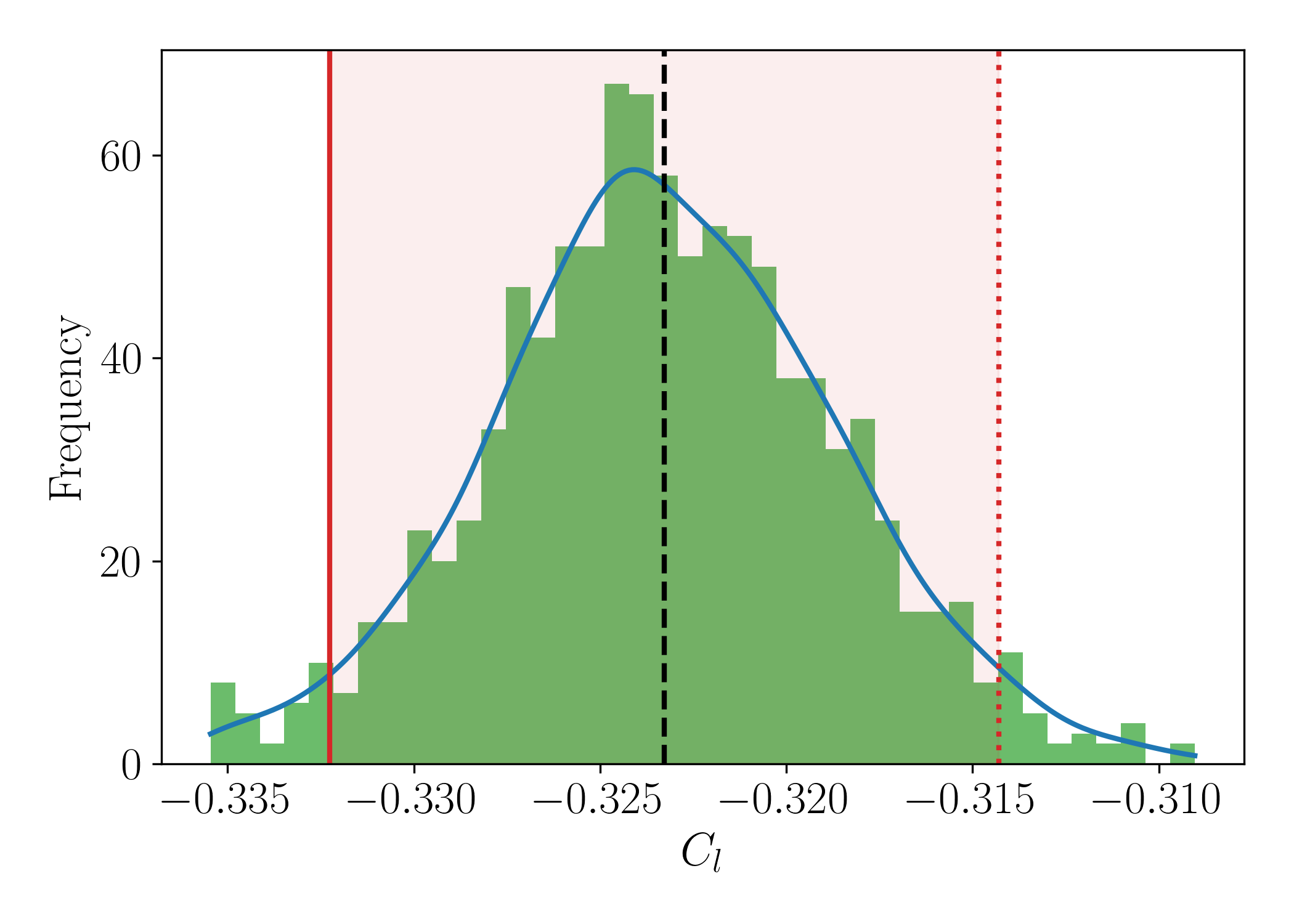}
    \end{subfigure}\\
        \caption*{$C_l$}
    \begin{subfigure}{.49\textwidth}
        \centering
        \includegraphics[width=\linewidth]{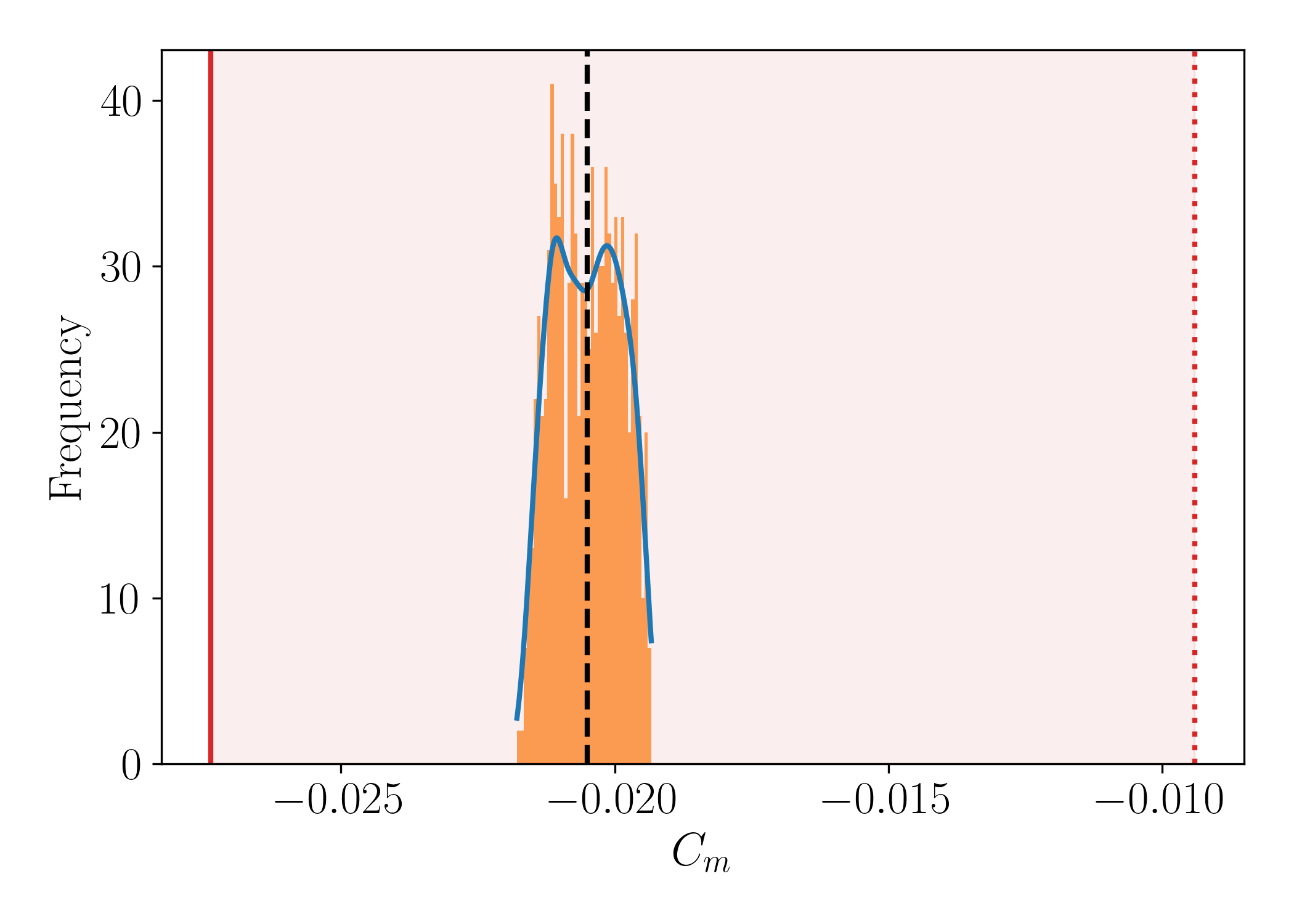}
    \end{subfigure}%
    \begin{subfigure}{.49\textwidth}
        \centering
        \includegraphics[width=\linewidth]{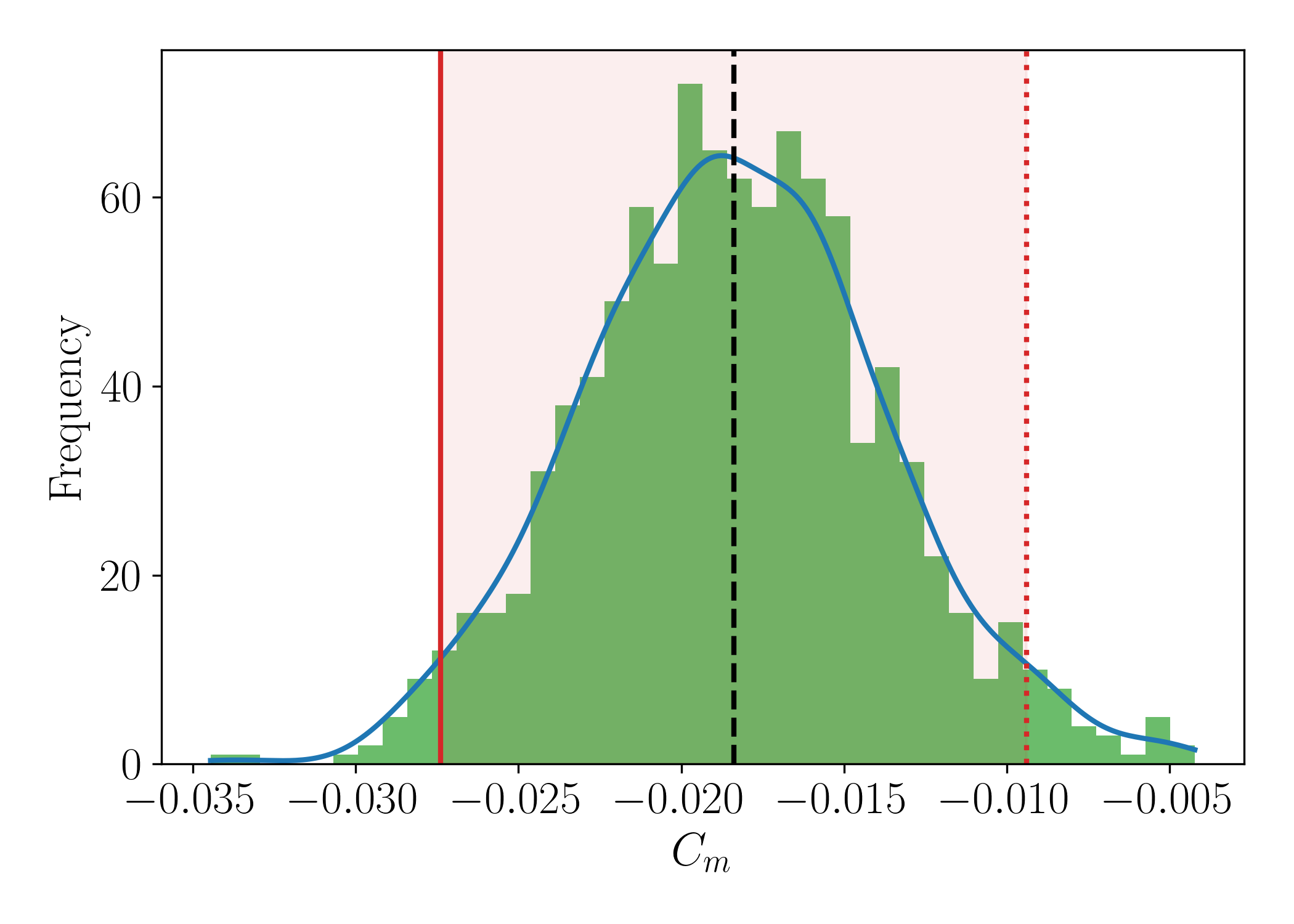}
    \end{subfigure}\\
            \caption*{$C_m$}
        \vspace{3mm}
    \begin{subfigure}{1\textwidth}
        \centering
        \includegraphics[width=\linewidth]{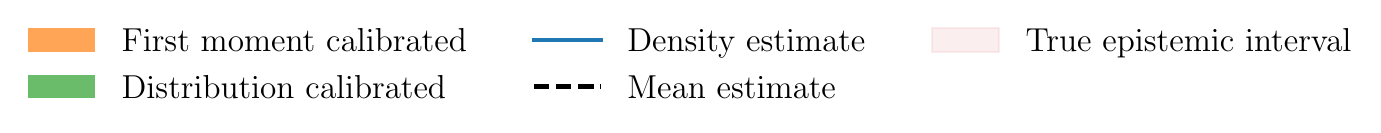}
    \end{subfigure}
    \caption{Prediction bounds at $\alpha=-3^\circ$ and $\beta_{\mathrm{flap}}=-3^\circ$. Each row shows one aerodynamic coefficient, with the first-moment-calibrated result on the left and the distribution-calibrated result on the right.}
    \label{fig:prediction_bounds_aoa_m3_flap_m3}
\end{figure}

\begin{figure}[p]
    \centering
    \begin{subfigure}{.49\textwidth}
                \caption*{First-moment calibrated.}
        \centering
        \includegraphics[width=\linewidth]{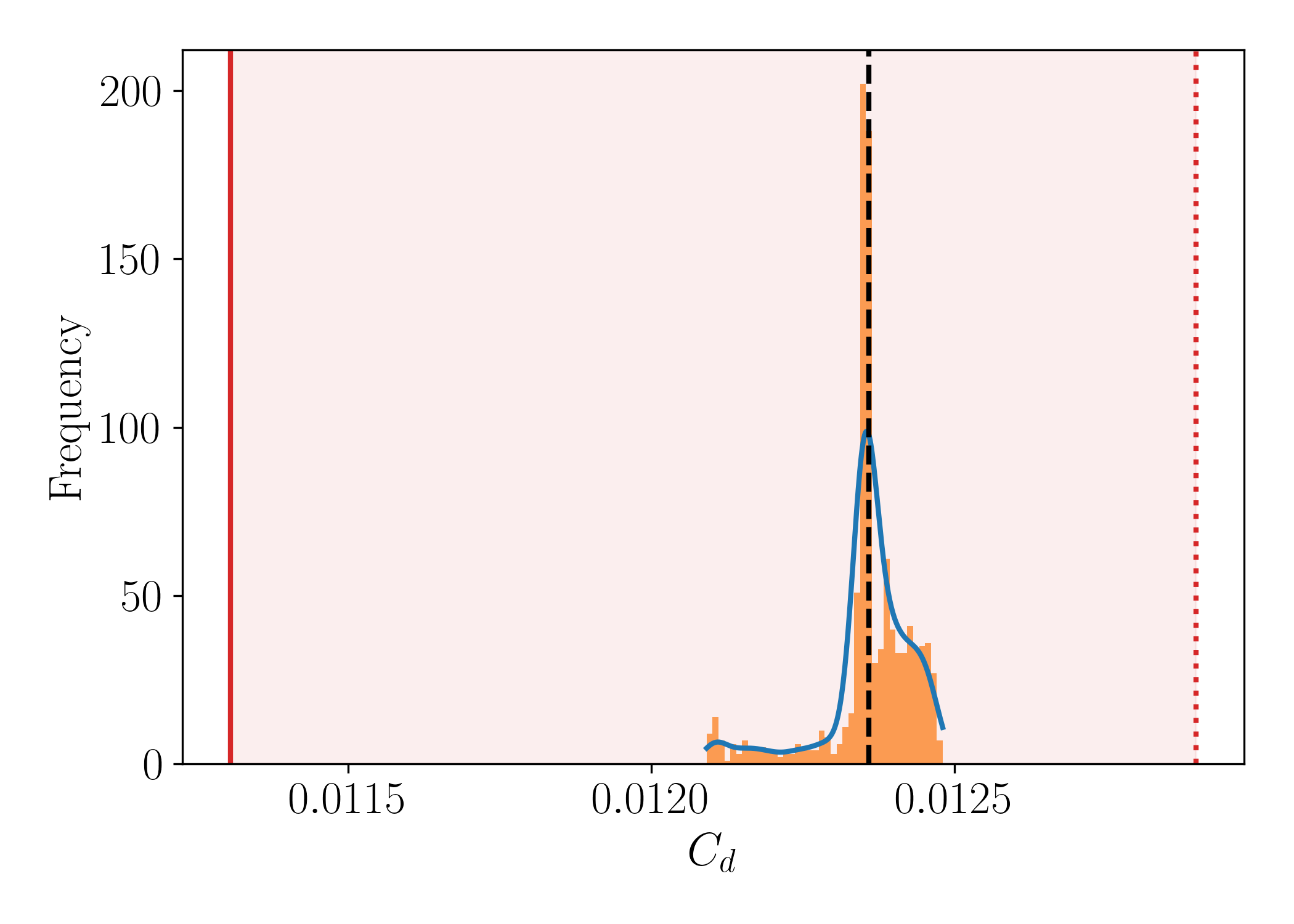}
    \end{subfigure}%
    \begin{subfigure}{.49\textwidth}
                \caption*{Distribution-calibrated.}
        \centering
        \includegraphics[width=\linewidth]{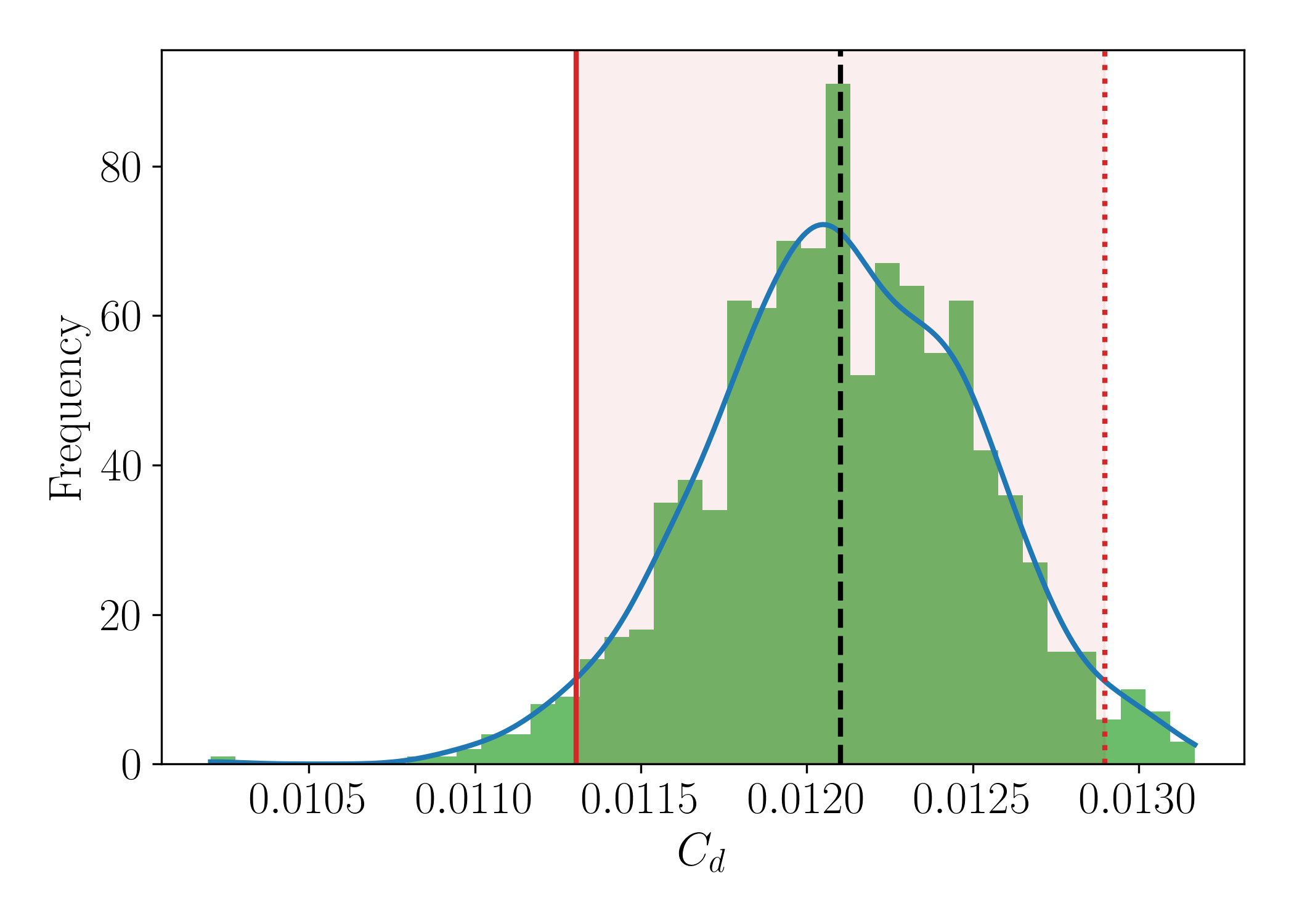}
    \end{subfigure}\\
           \caption*{$C_d$}
    \begin{subfigure}{.49\textwidth}
        \centering
        \includegraphics[width=\linewidth]{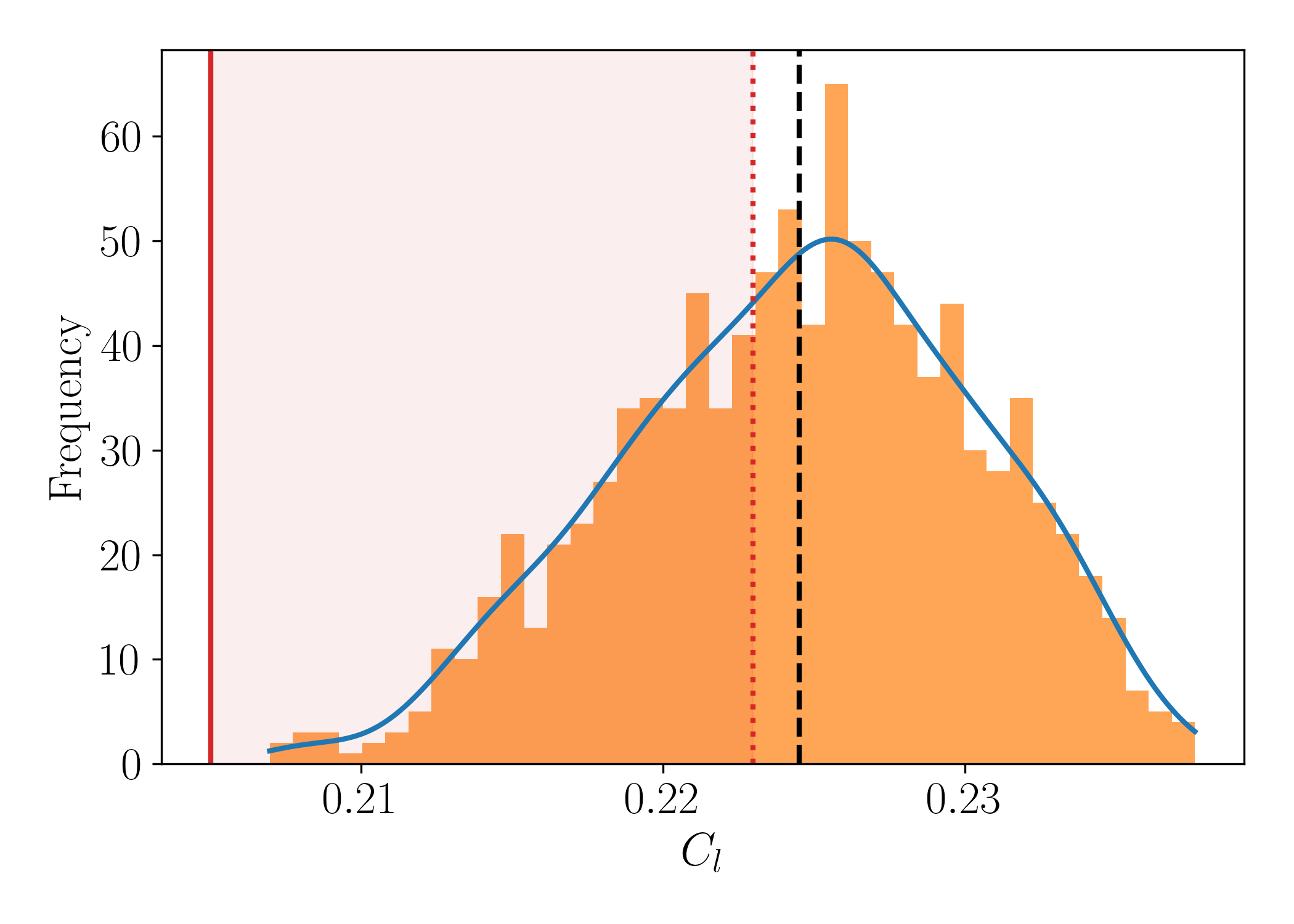}
    \end{subfigure}%
    \begin{subfigure}{.49\textwidth}
        \centering
        \includegraphics[width=\linewidth]{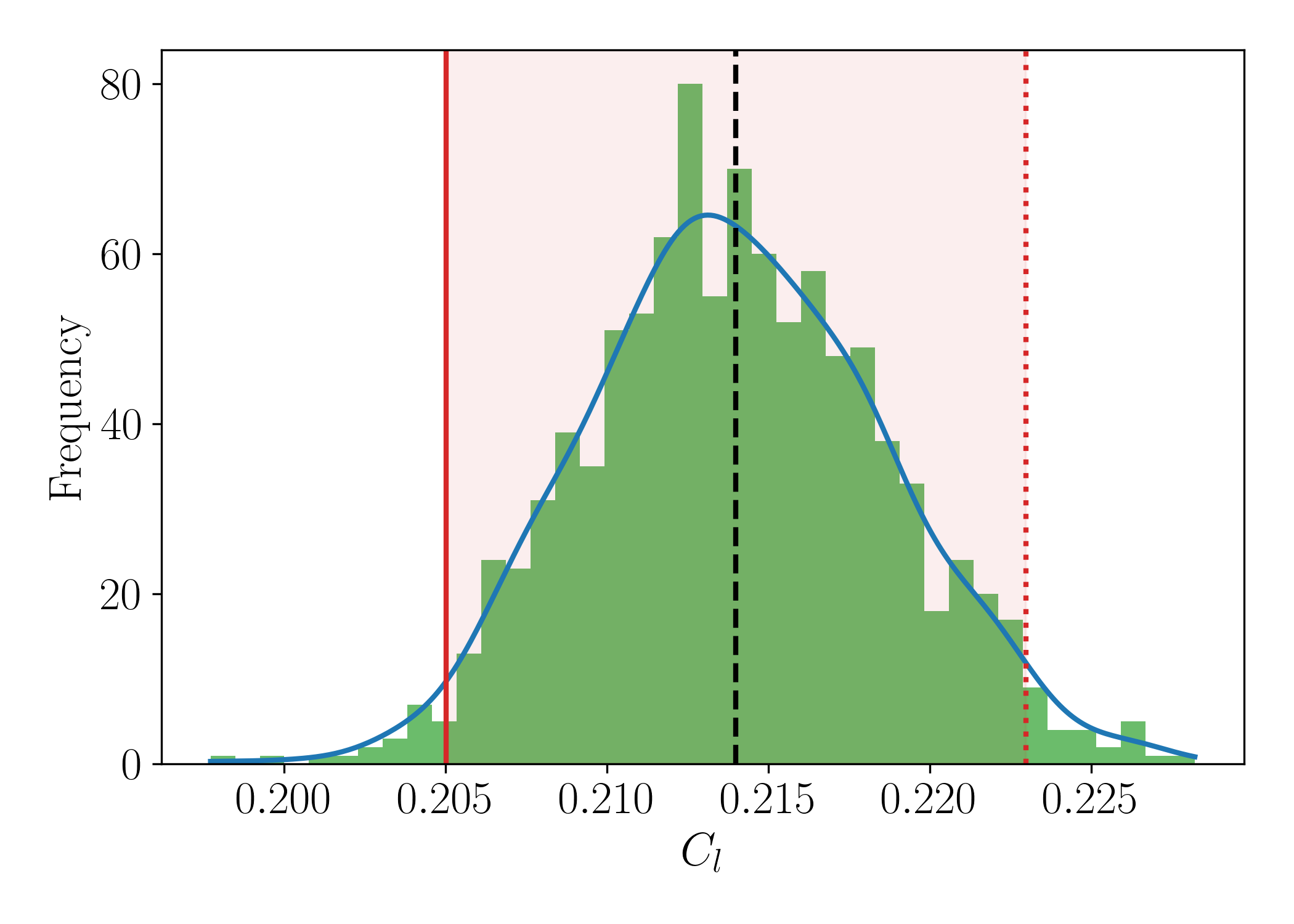}
    \end{subfigure}\\
               \caption*{$C_l$}
    \begin{subfigure}{.49\textwidth}
        \centering
        \includegraphics[width=\linewidth]{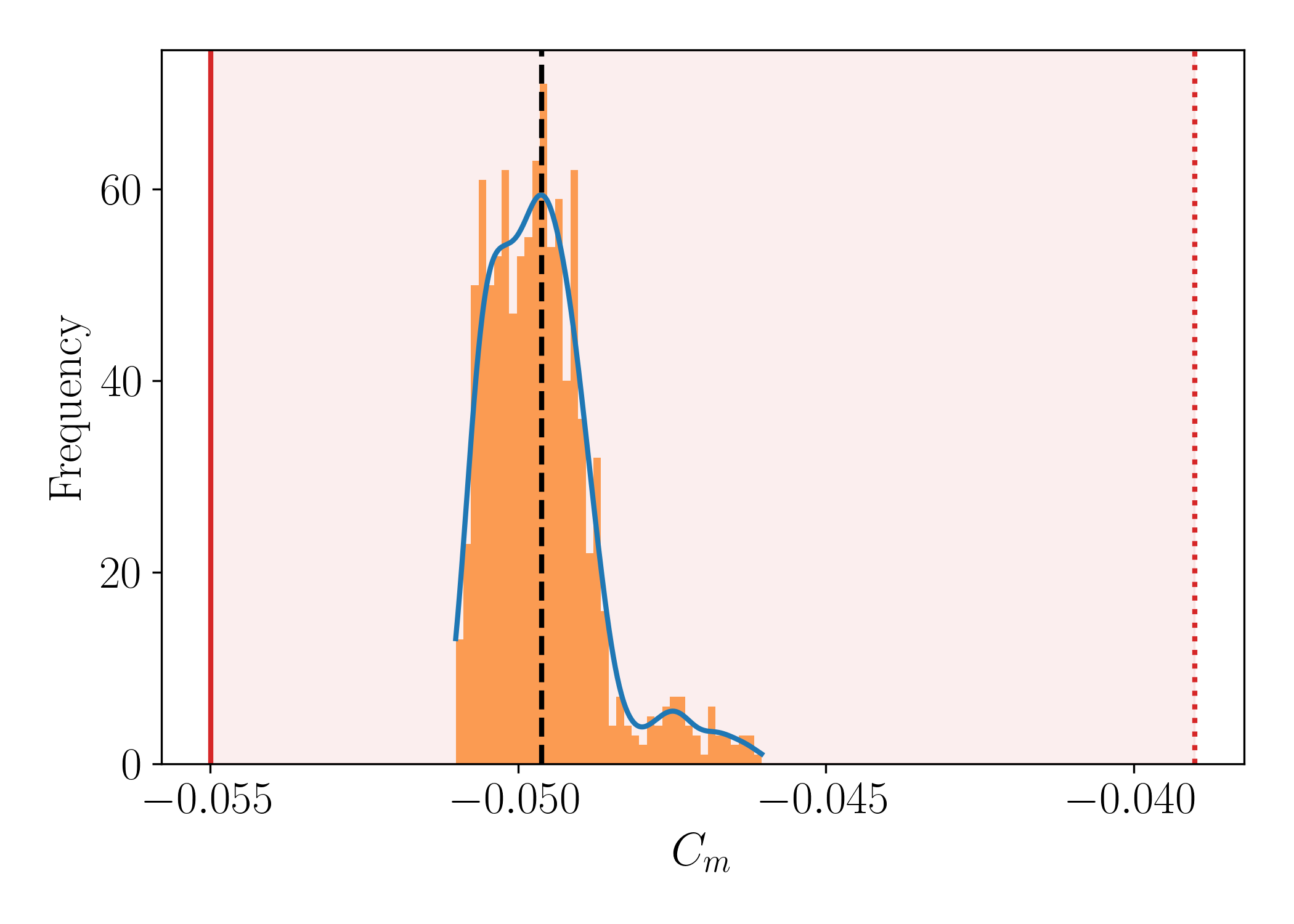}
    \end{subfigure}%
    \begin{subfigure}{.49\textwidth}
        \centering
        \includegraphics[width=\linewidth]{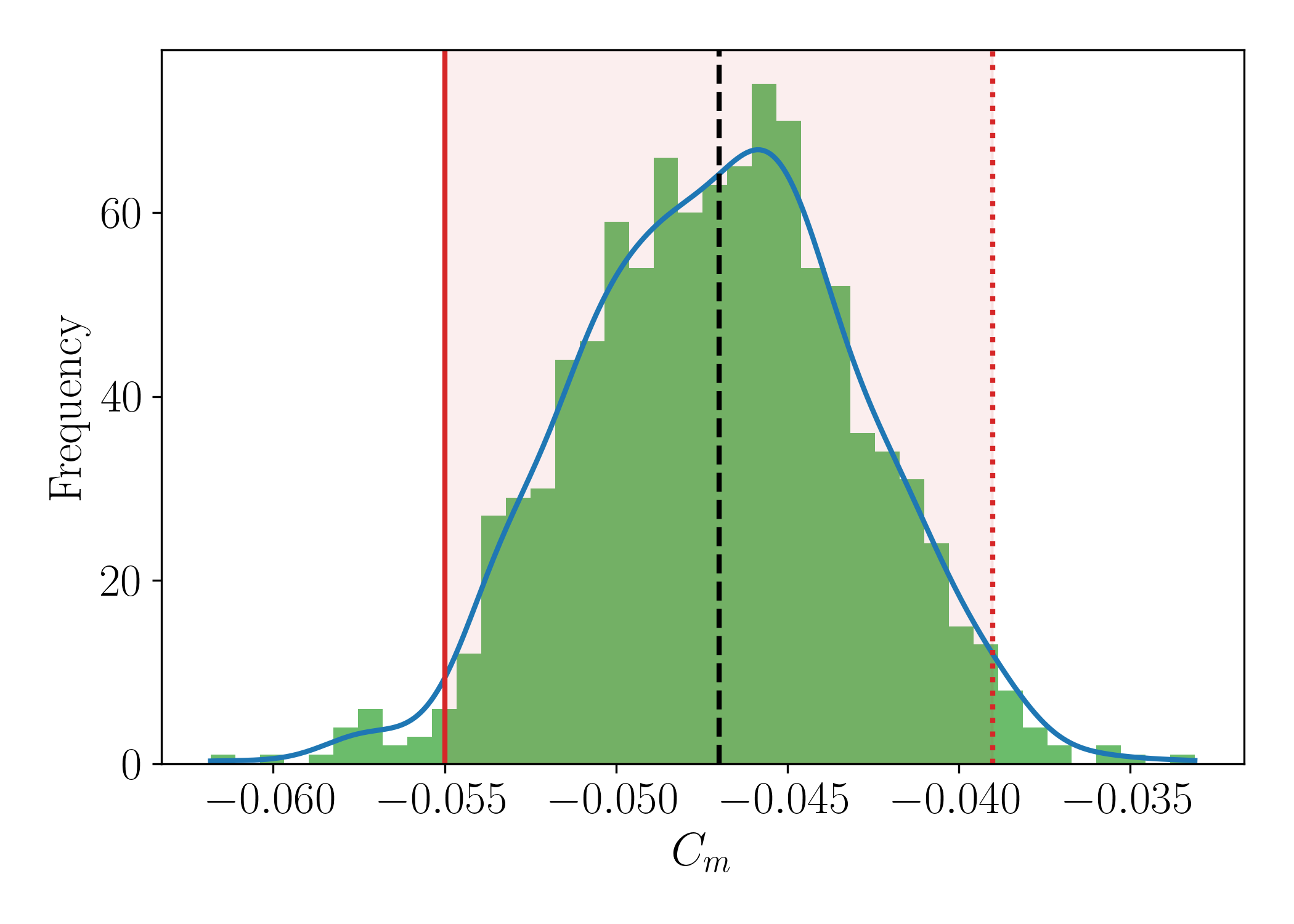}
    \end{subfigure}\\
               \caption*{$C_m$}
        \vspace{3mm}
    \begin{subfigure}{1\textwidth}
        \centering
        \includegraphics[width=\linewidth]{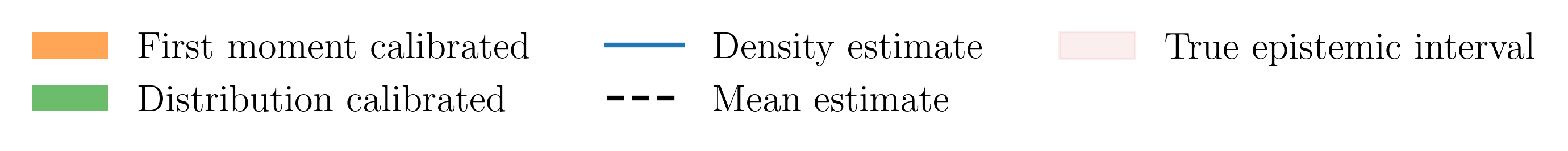}
    \end{subfigure}
    \caption{Prediction bounds at $\alpha=0^\circ$ and $\beta_{\mathrm{flap}}=0^\circ$. Each row shows one aerodynamic coefficient, with the first-moment-calibrated result on the left and the distribution-calibrated result on the right.}
    \label{fig:prediction_bounds_aoa_0_flap_0}
\end{figure}

\begin{figure}[p]
    \centering
    \begin{subfigure}{.49\textwidth}
                \caption*{First-moment calibrated.}
        \centering
        \includegraphics[width=\linewidth]{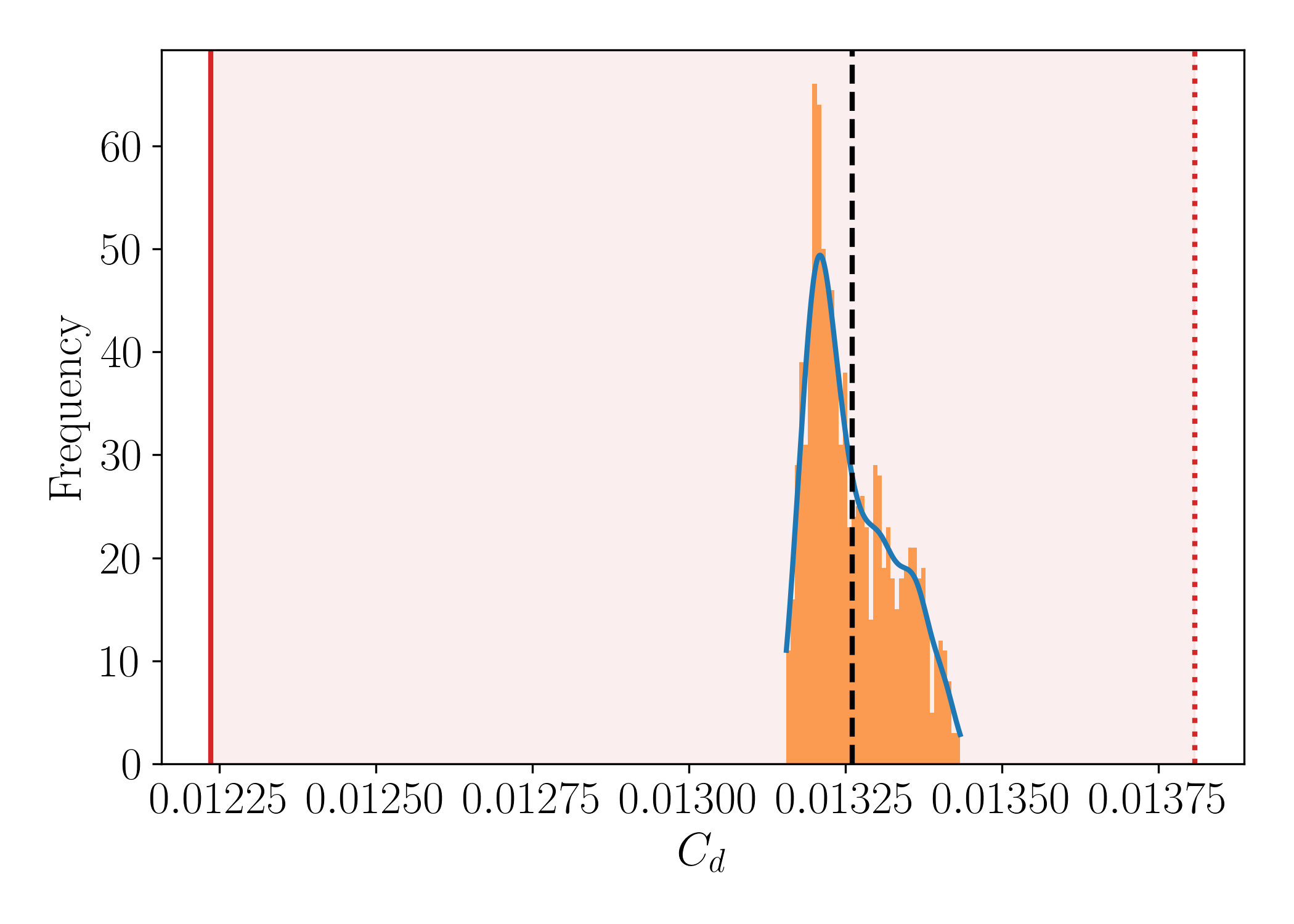}
    \end{subfigure}%
    \begin{subfigure}{.49\textwidth}
                \caption*{Distribution-calibrated.}
        \centering
        \includegraphics[width=\linewidth]{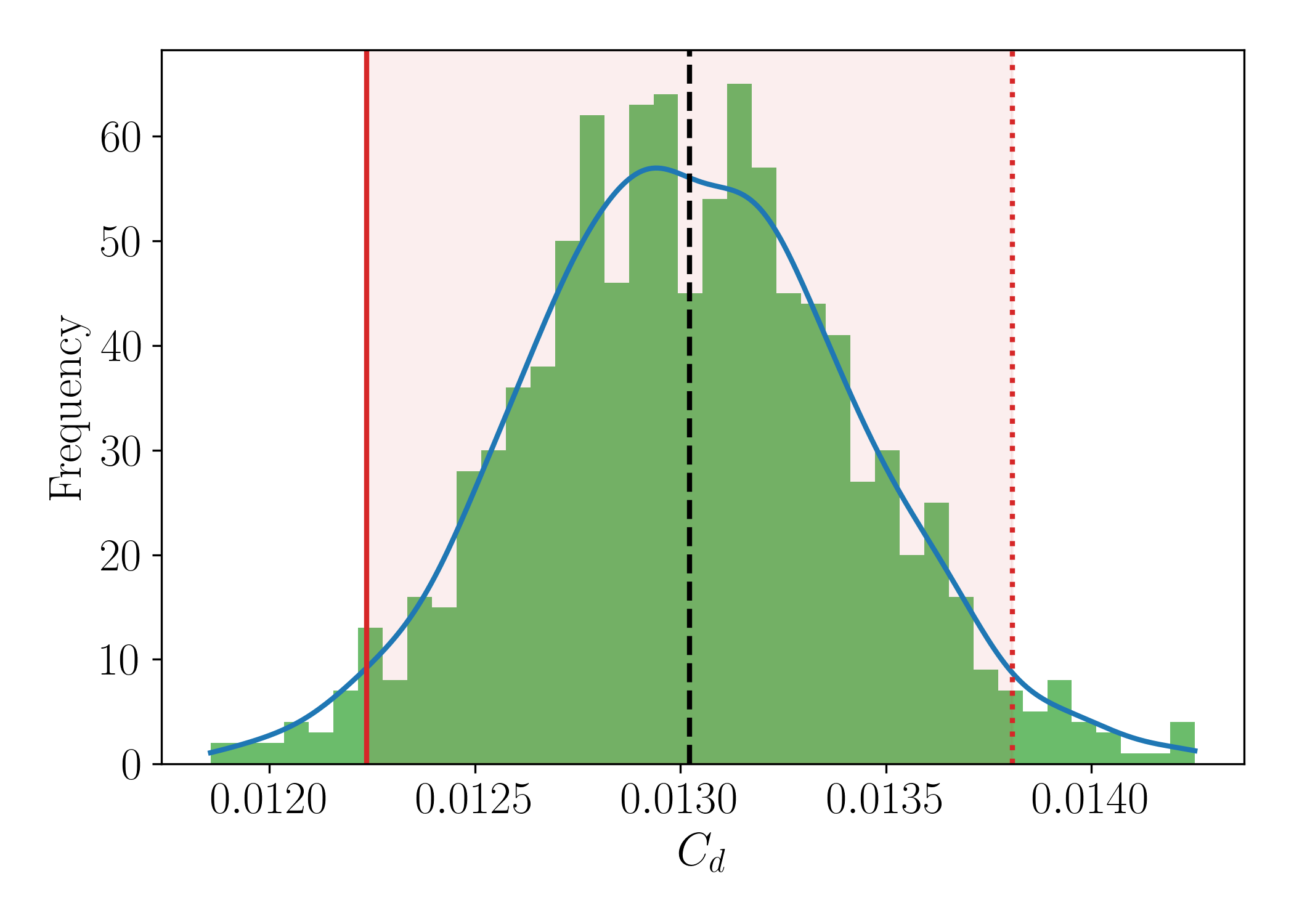}
    \end{subfigure}\\
               \caption*{$C_d$}
    \begin{subfigure}{.49\textwidth}
        \centering
        \includegraphics[width=\linewidth]{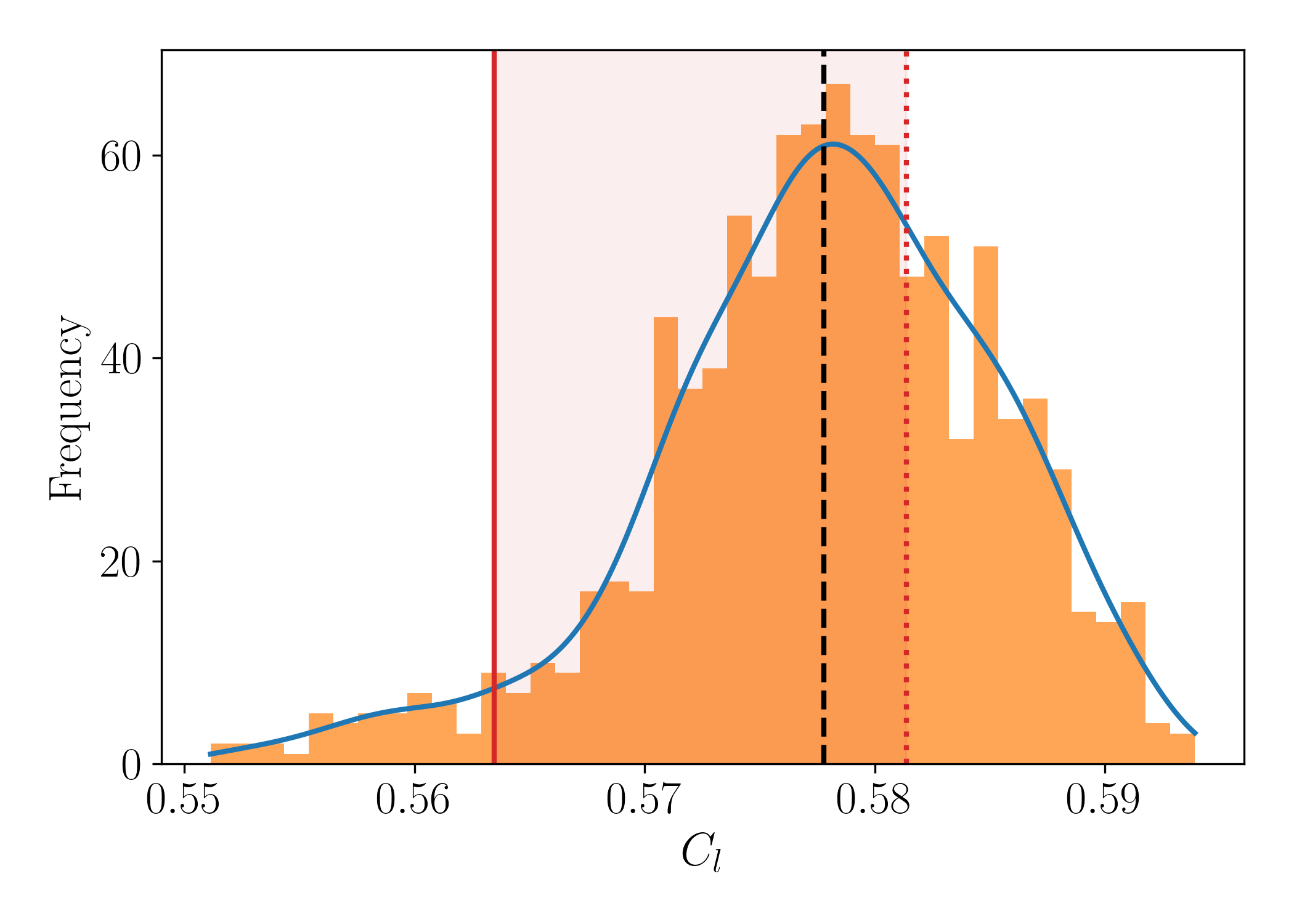}
    \end{subfigure}%
    \begin{subfigure}{.49\textwidth}
        \centering
        \includegraphics[width=\linewidth]{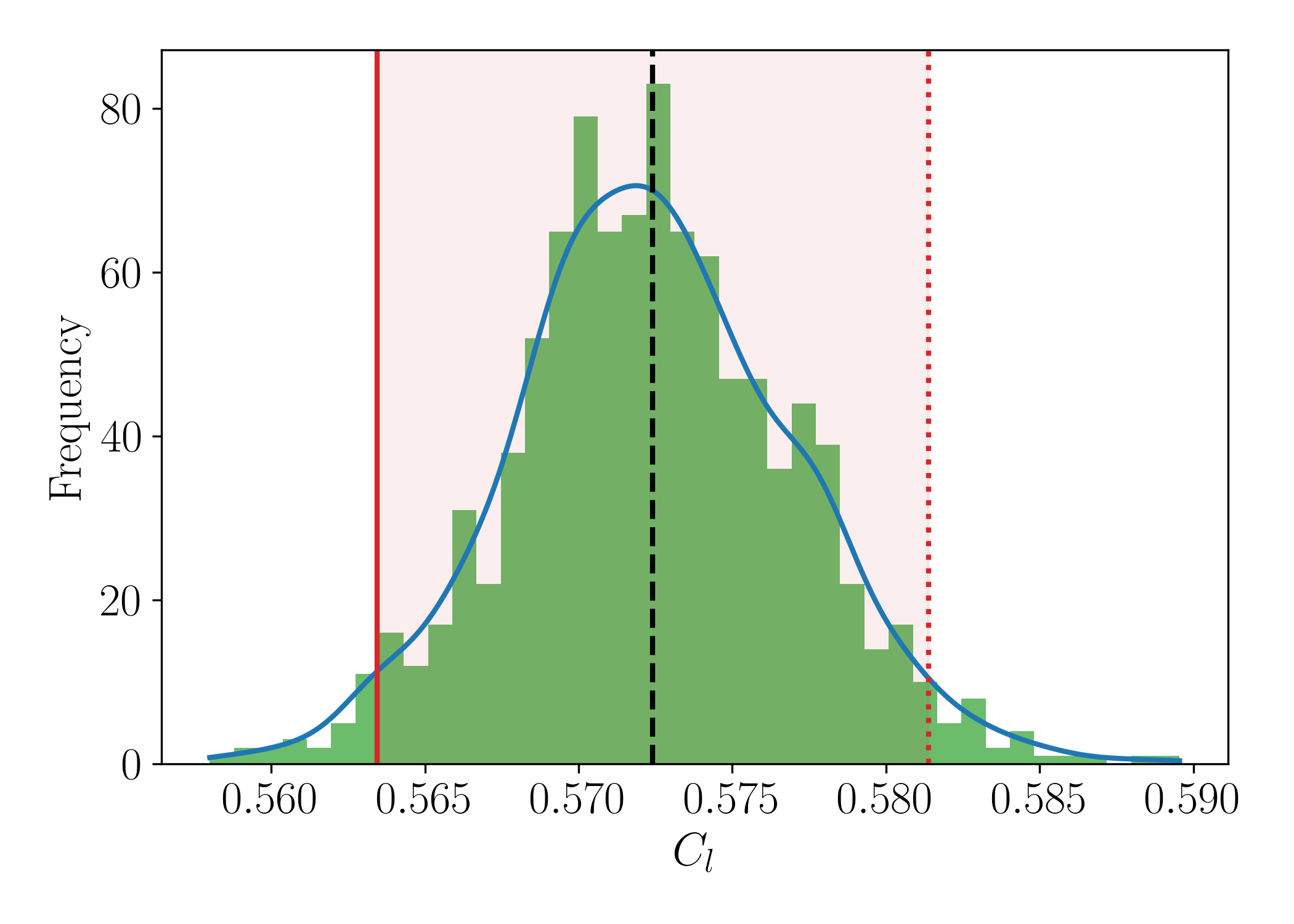}
    \end{subfigure}\\
               \caption*{$C_l$}
    \begin{subfigure}{.49\textwidth}
        \centering
        \includegraphics[width=\linewidth]{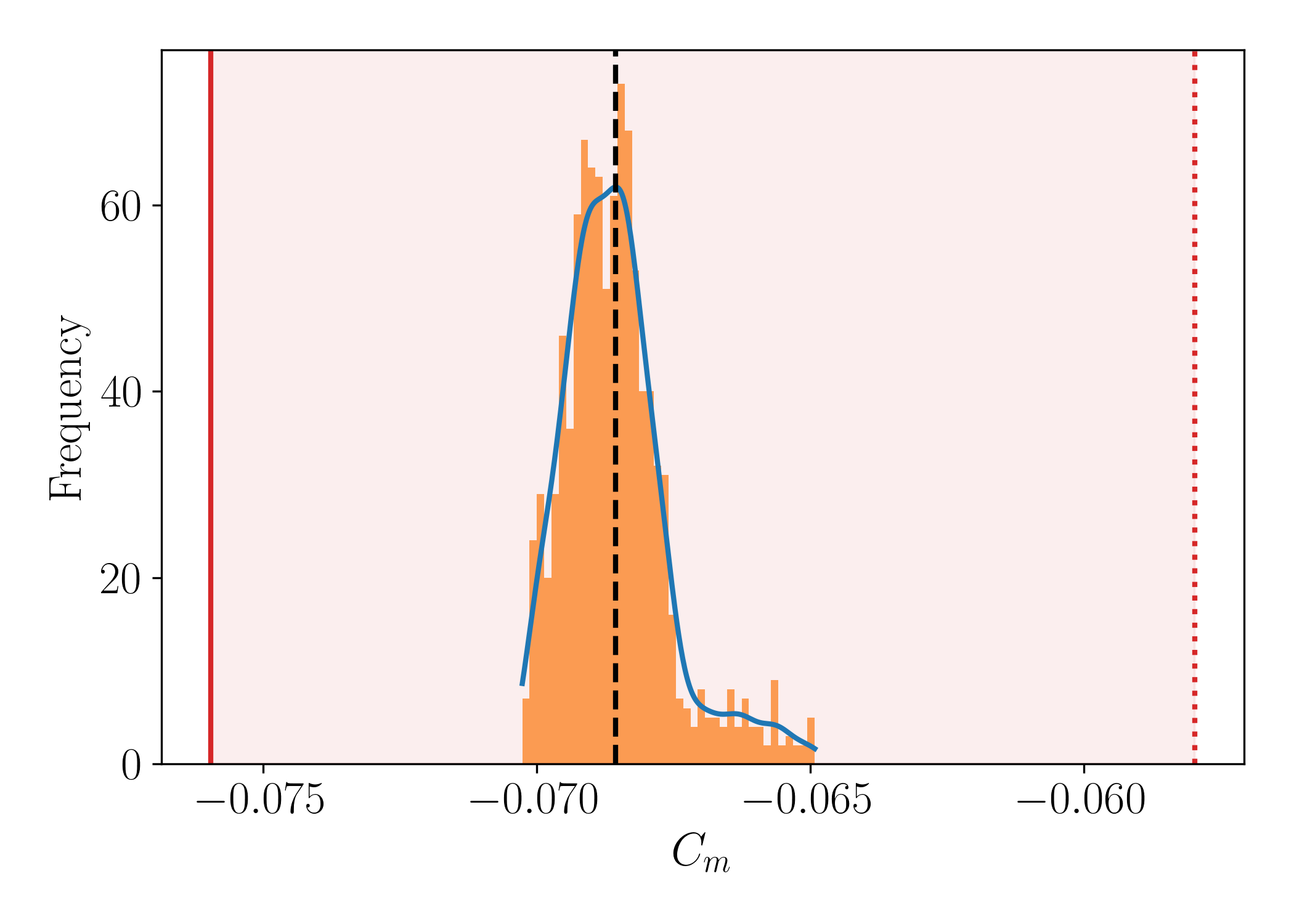}
    \end{subfigure}%
    \begin{subfigure}{.49\textwidth}
        \centering
        \includegraphics[width=\linewidth]{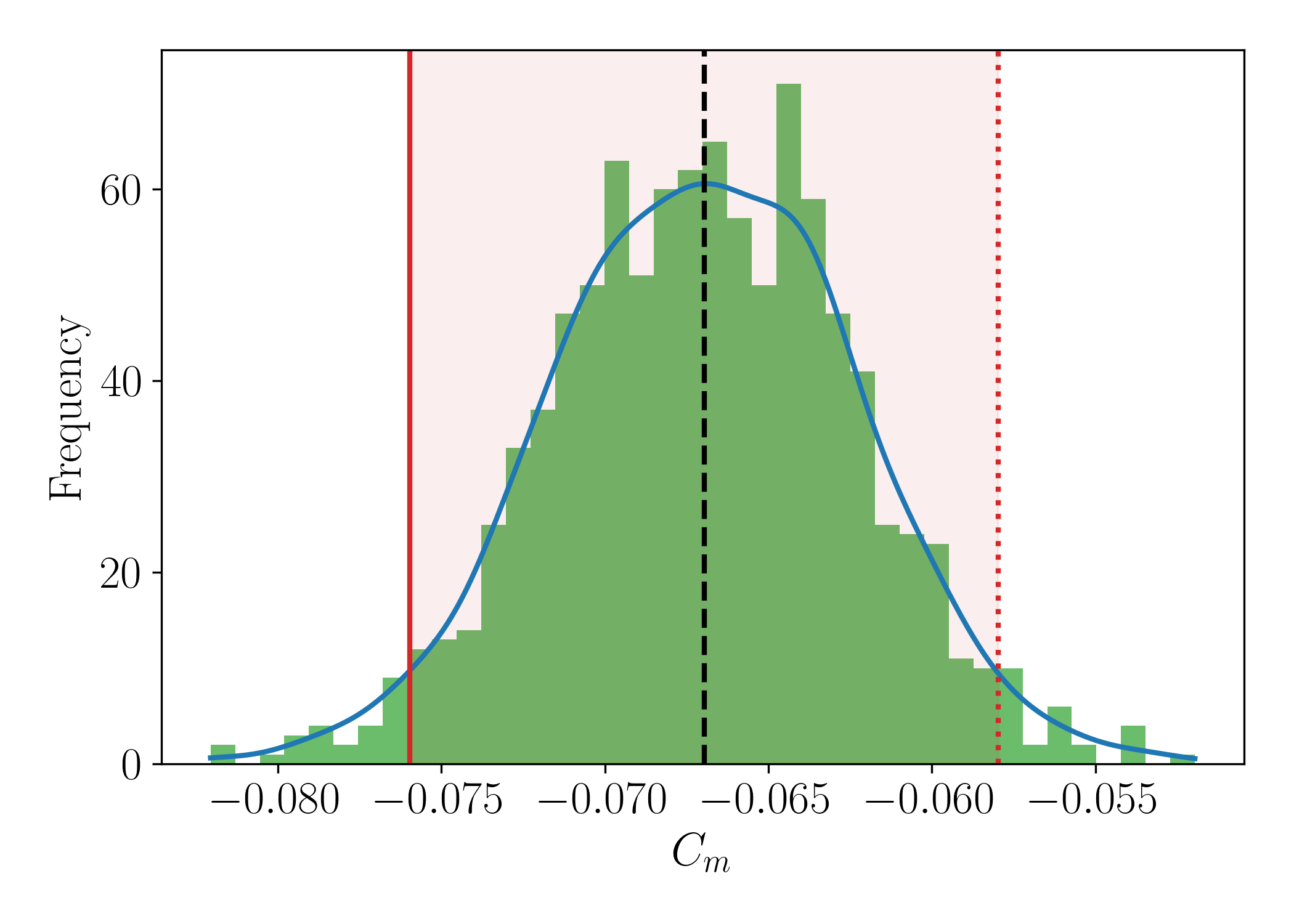}
    \end{subfigure}\\
               \caption*{$C_m$}
    \vspace{3mm}
    \begin{subfigure}{1\textwidth}
        \centering
        \includegraphics[width=1\linewidth]{figures/legend.png}

    \end{subfigure}
    \caption{Prediction bounds at $\alpha=2^\circ$ and $\beta_{\mathrm{flap}}=2^\circ$. Each row shows one aerodynamic coefficient, with the first-moment-calibrated result on the left and the distribution-calibrated result on the right.}
    \label{fig:prediction_bounds_aoa_2_flap_2}
\end{figure}

\begin{figure}[p]
    \centering
    \begin{subfigure}{.49\textwidth}
                \caption*{First-moment calibrated.}
        \centering
        \includegraphics[width=\linewidth]{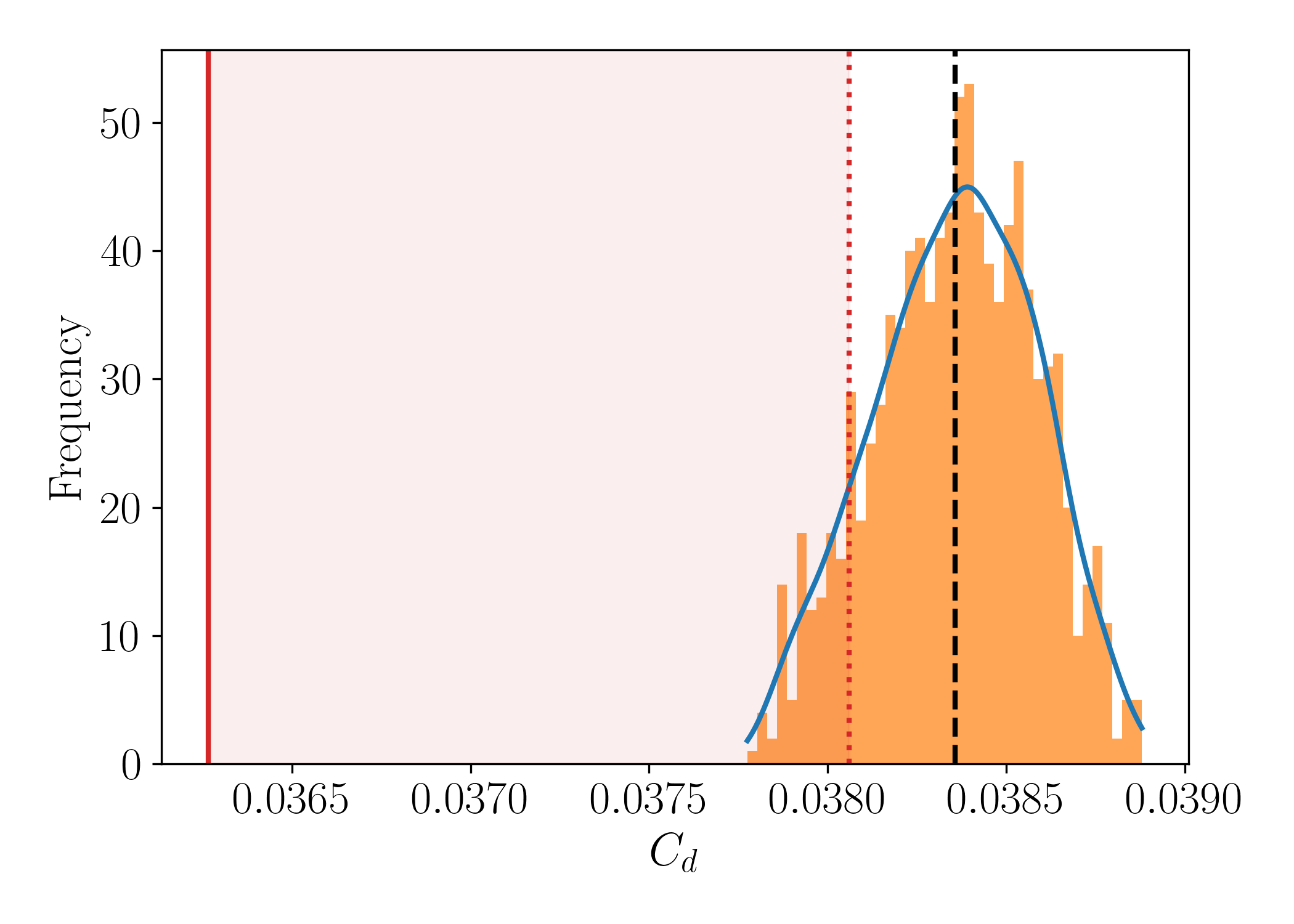}
    \end{subfigure}%
    \begin{subfigure}{.49\textwidth}
                \caption*{Distribution-calibrated.}
        \centering
        \includegraphics[width=\linewidth]{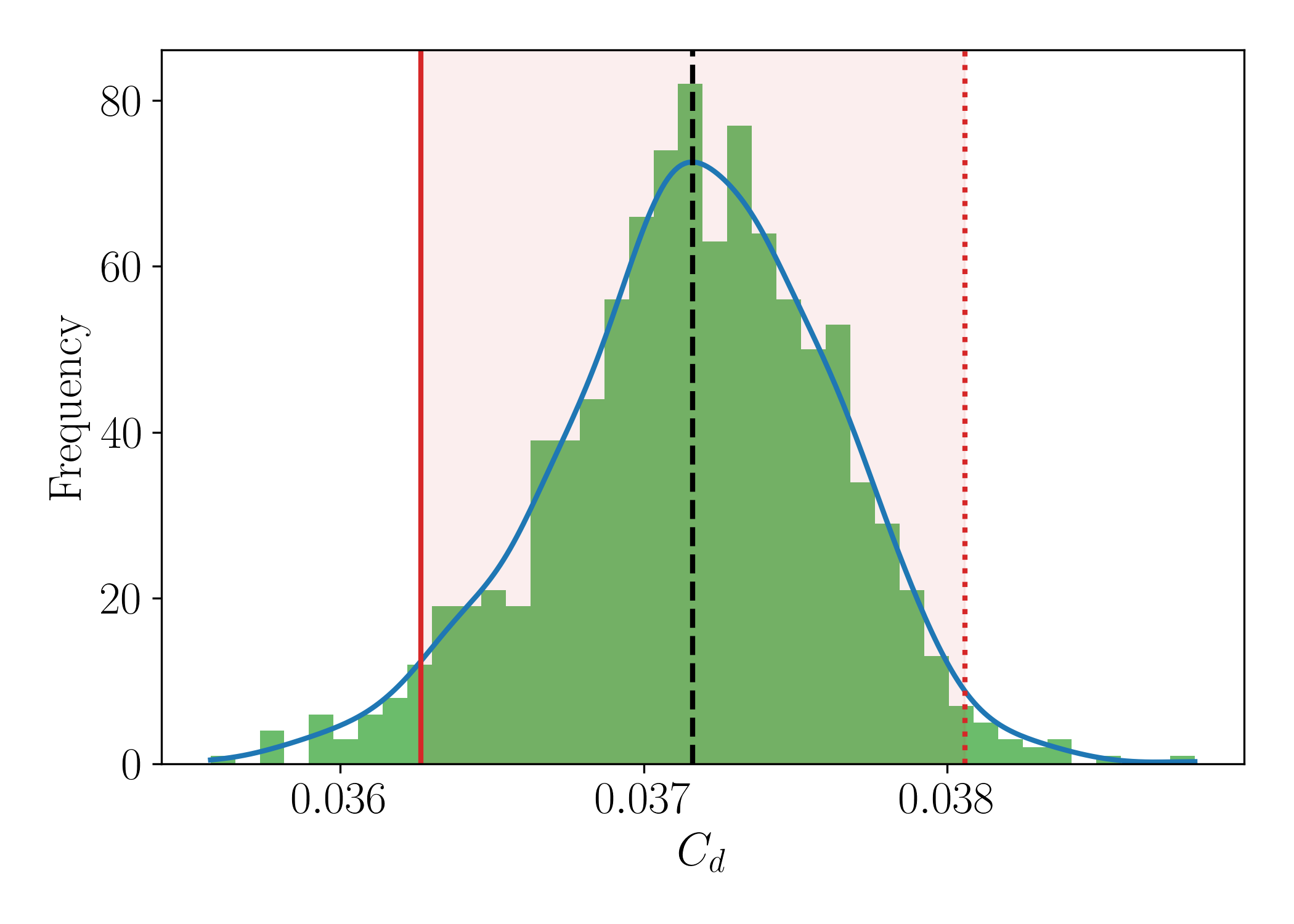}
    \end{subfigure}\\
               \caption*{$C_d$}
    \begin{subfigure}{.49\textwidth}
        \centering
        \includegraphics[width=\linewidth]{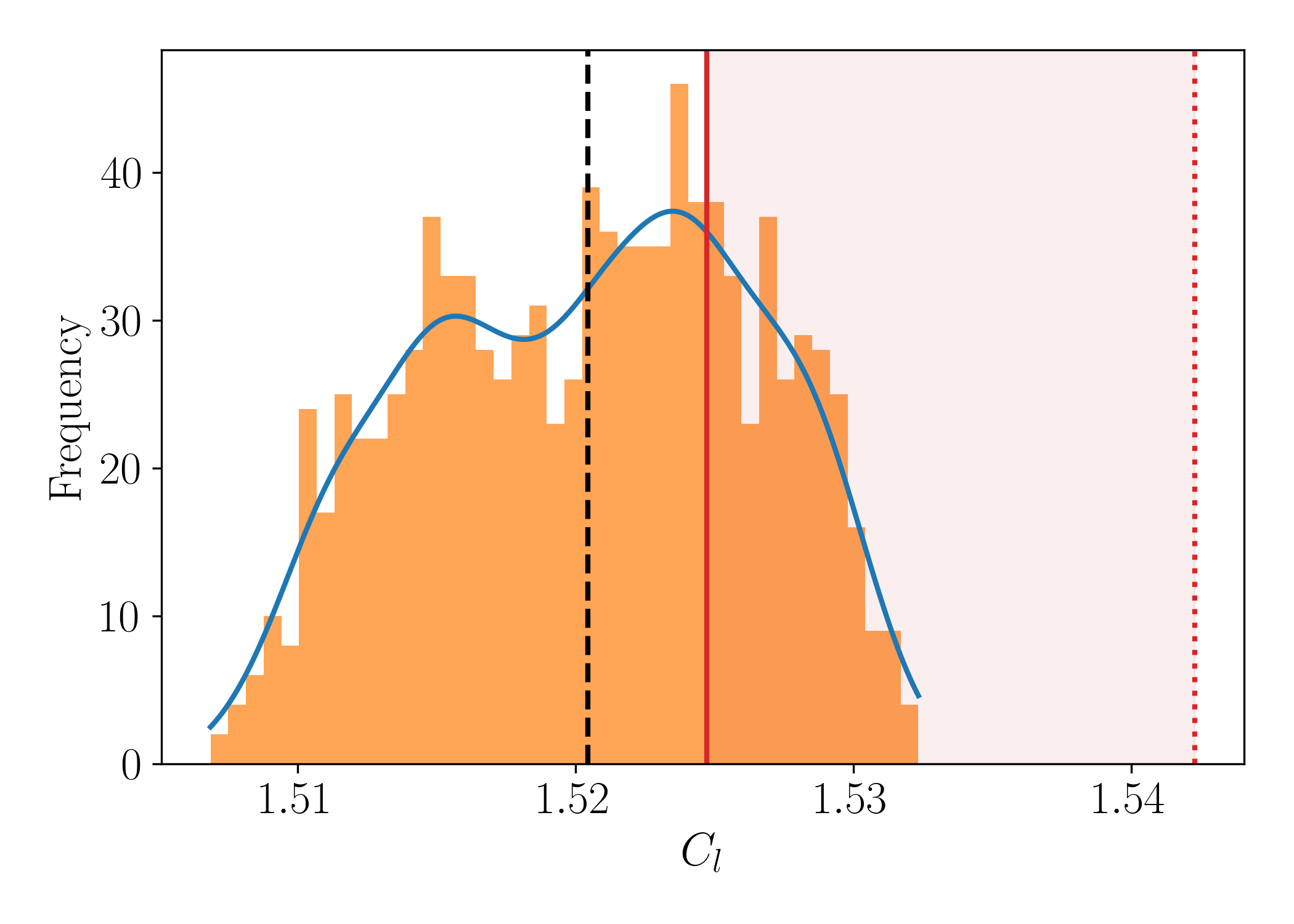}
    \end{subfigure}%
    \begin{subfigure}{.49\textwidth}
        \centering
        \includegraphics[width=\linewidth]{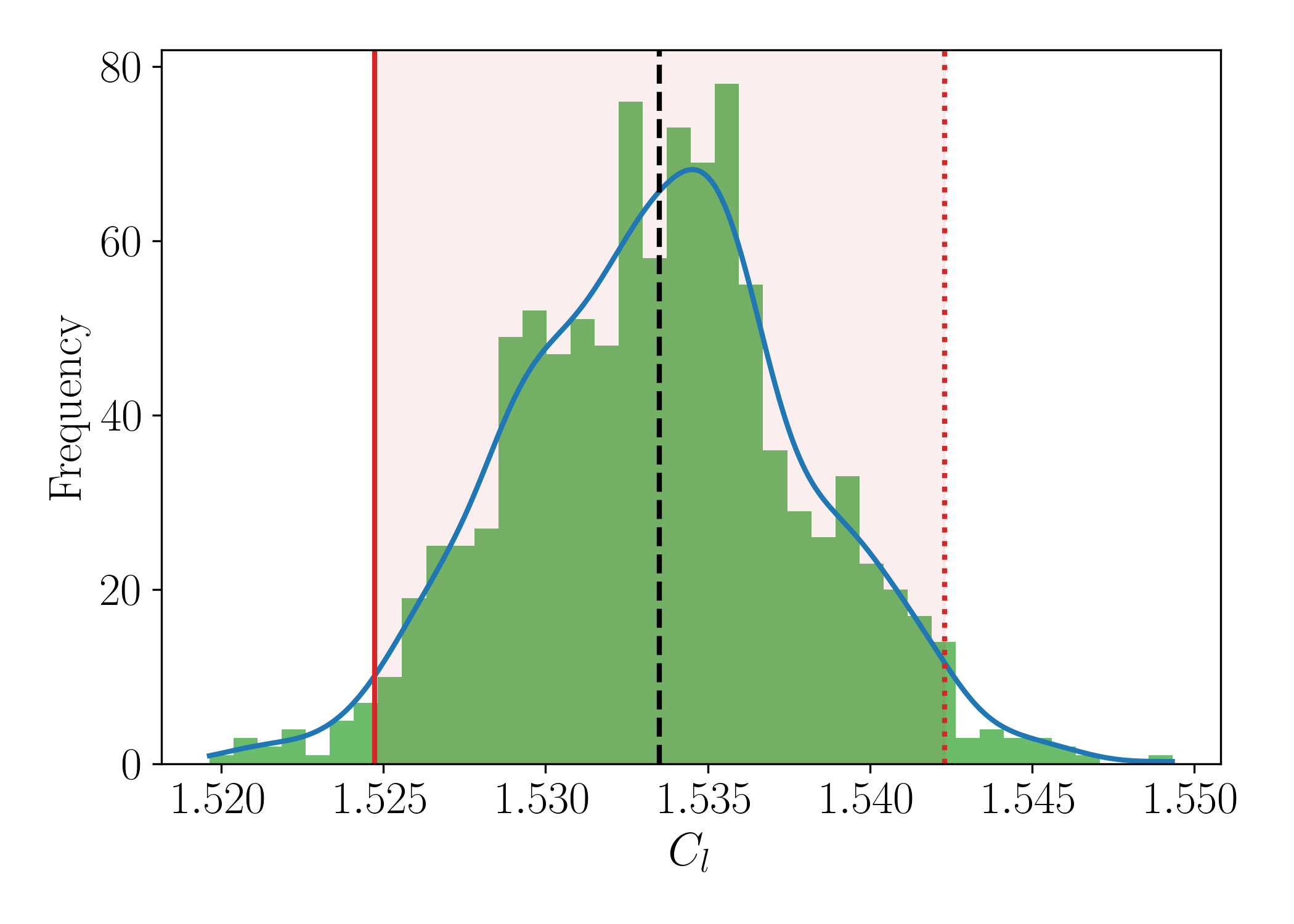}
    \end{subfigure}\\
               \caption*{$C_l$}
    \begin{subfigure}{.49\textwidth}
        \centering
        \includegraphics[width=\linewidth]{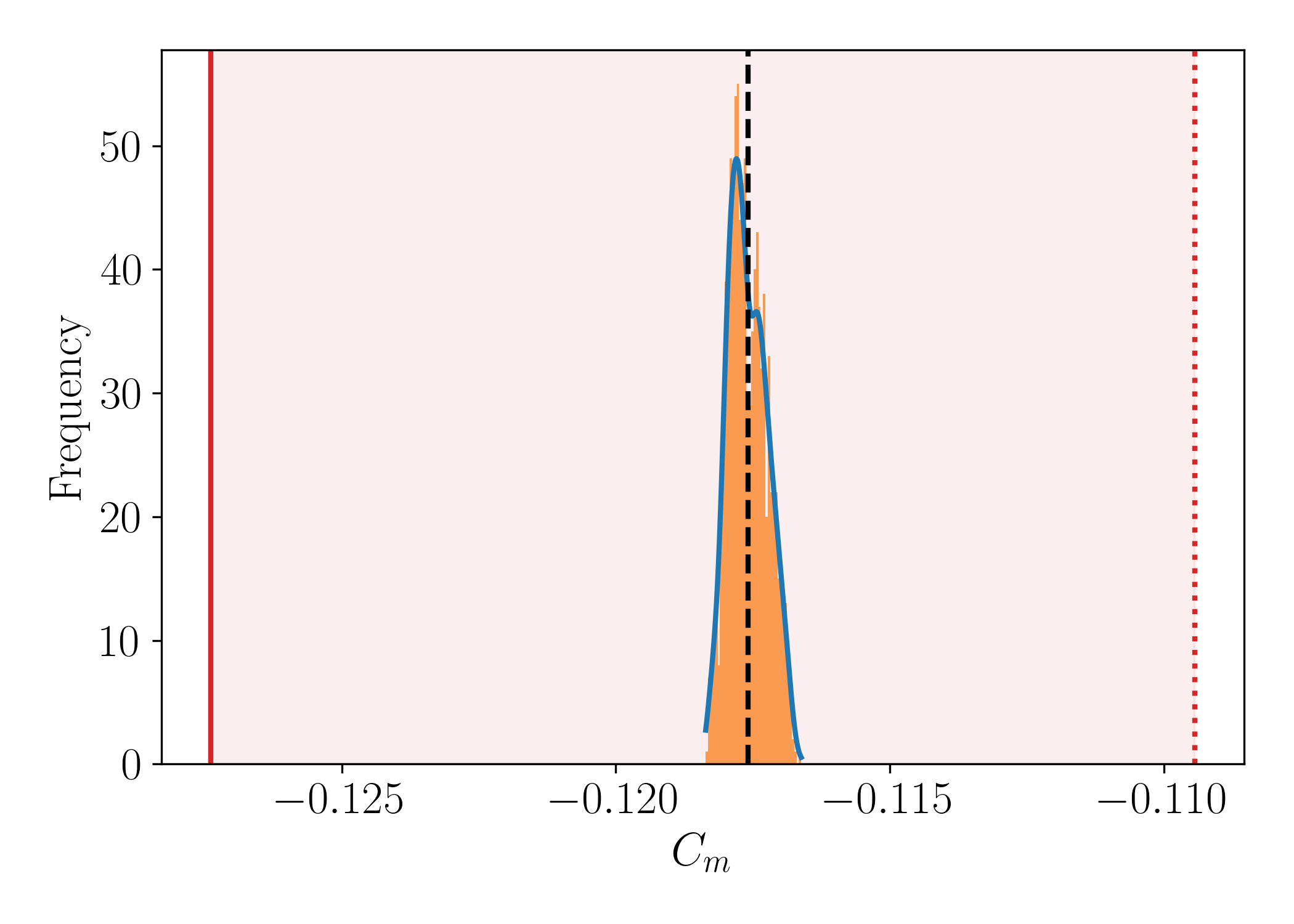}
    \end{subfigure}%
    \begin{subfigure}{.49\textwidth}
        \centering
        \includegraphics[width=\linewidth]{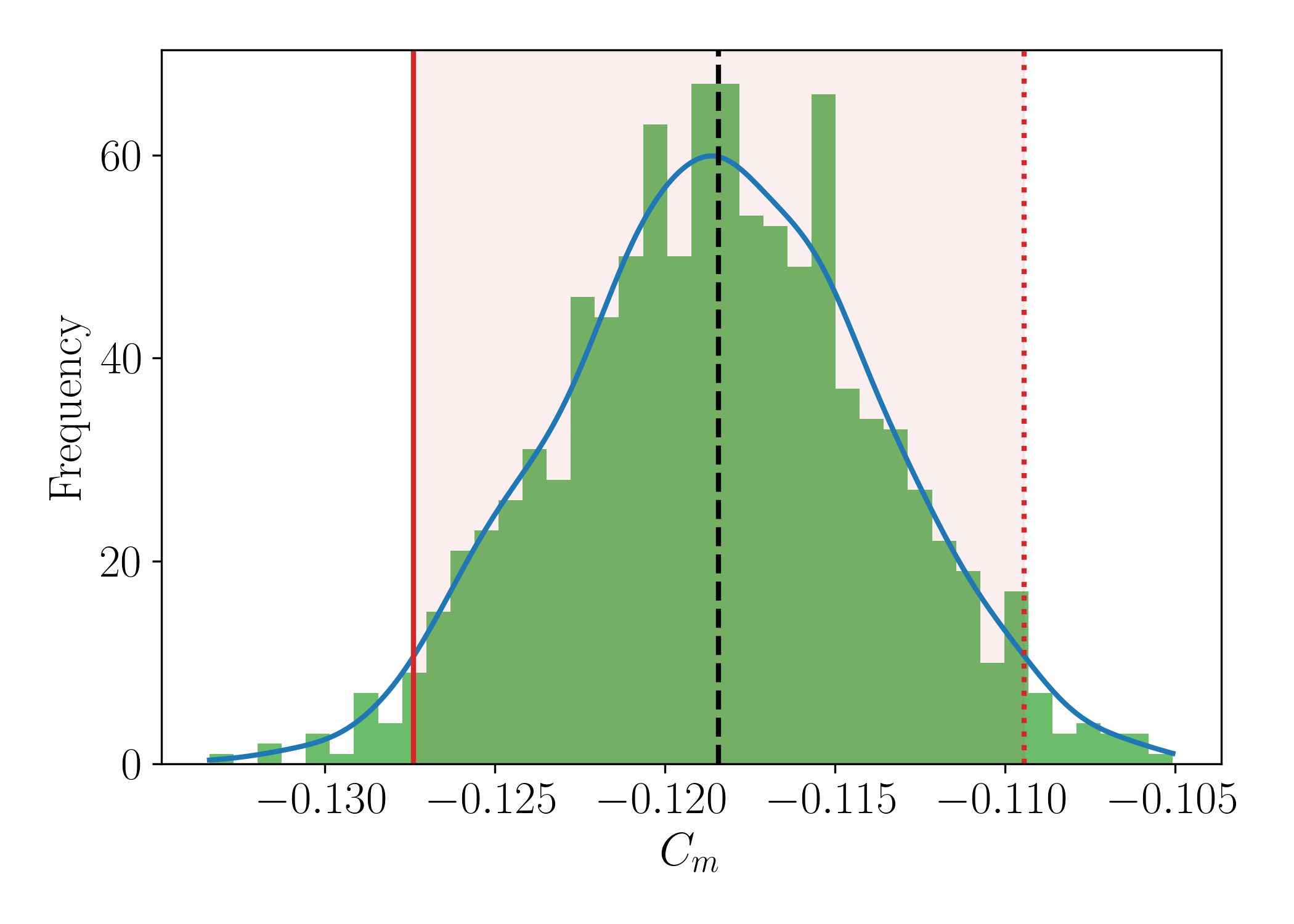}
    \end{subfigure}\\
               \caption*{$C_m$}
        \vspace{3mm}
    \begin{subfigure}{1\textwidth}
        \centering
        \includegraphics[width=\linewidth]{figures/legend.png}
    \end{subfigure}
    \caption{Prediction bounds at $\alpha=7^\circ$ and $\beta_{\mathrm{flap}}=15^\circ$. Each row shows one aerodynamic coefficient, with the first-moment-calibrated result on the left and the distribution-calibrated result on the right.}
    \label{fig:prediction_bounds_aoa_7_flap_15}
\end{figure}

\Cref{fig:prediction_bounds_aoa_m3_flap_m3,fig:prediction_bounds_aoa_0_flap_0,fig:prediction_bounds_aoa_2_flap_2,fig:prediction_bounds_aoa_7_flap_15} compare the Monte Carlo predictive distributions obtained from the first-moment-calibrated model and the distribution-calibrated model at the four prediction locations. In each panel, the red vertical lines and shaded region denote the released epistemic bounds for the corresponding quantity of interest, while the colored histograms show samples from the calibrated predictive distribution after propagation of input uncertainty. Thus, a well-calibrated result should do more than place its mean near the center of the red interval; it should also assign approximately the correct amount of probability mass to that interval.

The distribution-calibrated results satisfy this criterion consistently. Across all four prediction locations and all three aerodynamic coefficients, the green predictive distributions are centered near the provided epistemic interval centroids. This agreement is visible in the individual histogram comparisons: for $C_l$, $C_d$, and $C_m$, the green samples span nearly the same range as the truth intervals, including at the off-design point $\alpha=7^\circ$, $\beta_{\mathrm{flap}}=15^\circ$, where extrapolation from the calibration data is most demanding. Quantitatively, the empirical mass of the green samples lying inside the supplied truth interval remains close to the intended 95\% level for every QoI, ranging from about 94.2\% to 95.8\% over the twelve coefficient/location combinations.

The first-moment calibrated results, shown in orange, are less reliable because the calibration constrains only the centroid of the truth data and not the epistemic width of the truth intervals. As a result, the orange distributions are often shifted, too narrow, or both. For example, at $\alpha=7^\circ$, $\beta_{\mathrm{flap}}=15^\circ$, the first-moment calibrated $C_d$ distribution lies mostly to the right of the released truth interval, while the $C_l$ distribution places a substantial fraction of its mass below the interval. In several other cases, especially for $C_d$ and $C_m$ near $\alpha=0^\circ$ and $\alpha=2^\circ$, the orange distributions fall inside the red interval but are much narrower than the truth bounds; this gives apparent coverage of the interval edges without representing the released epistemic uncertainty. Over all twelve comparisons, the fraction of orange samples inside the truth interval varies widely, from roughly 11\% to 100\%, which indicates that first-moment calibration does not produce uniformly calibrated epistemic bounds.

\begin{figure}[htb!]
    \centering
    \includegraphics[width=\linewidth]{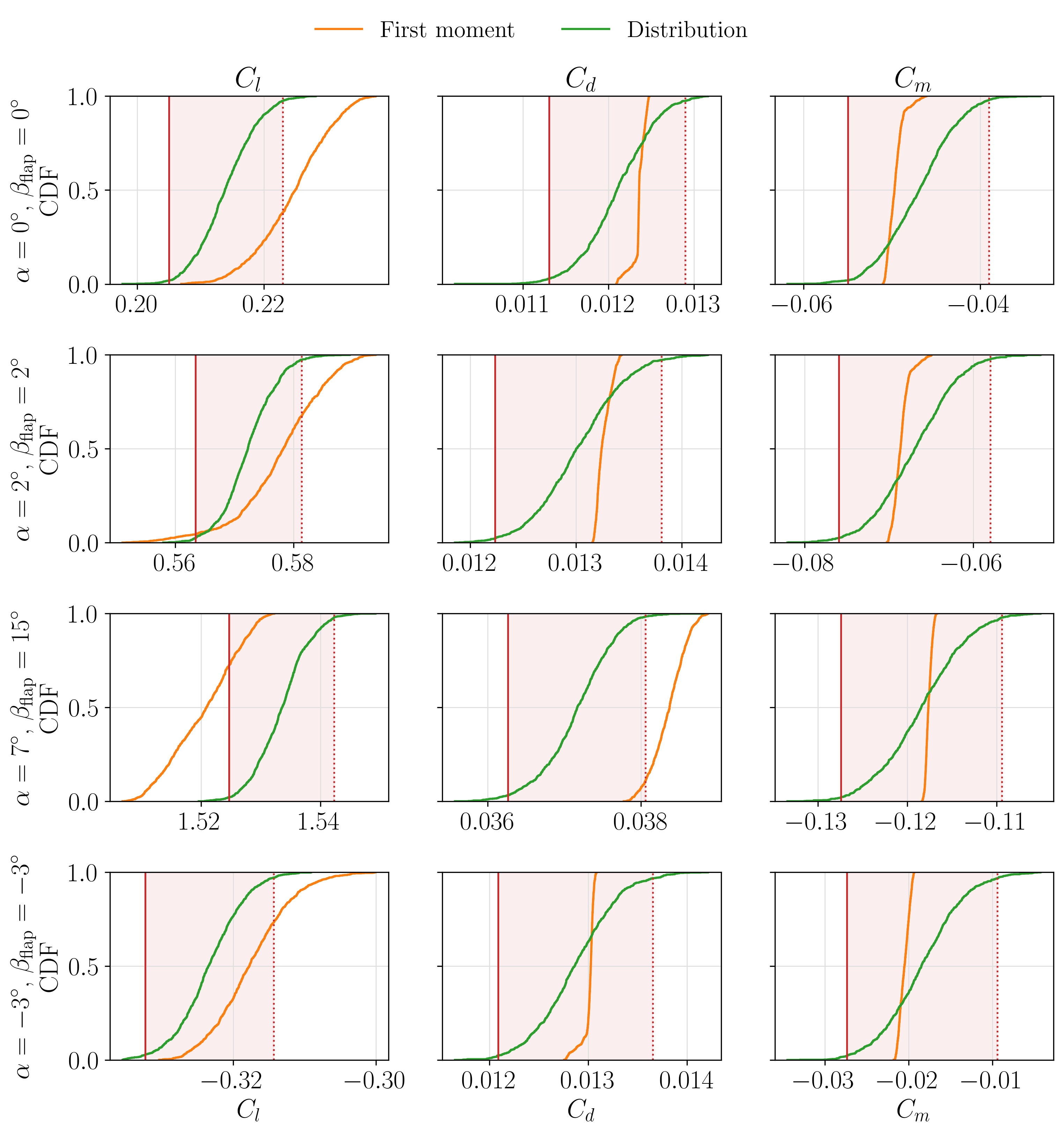}
    \caption{Overview of empirical CDFs for the four prediction settings and three aerodynamic coefficients. The orange curves show the first-moment-calibrated samples, the green curves show the distribution-calibrated samples, and the red vertical lines mark the provided epistemic interval edges.}
    \label{fig:calibration_cdf_overview}
\end{figure}

\Cref{fig:calibration_cdf_overview} gives the same comparison in cumulative form. If the released bounds are interpreted as 95\% intervals, the left and right red lines should intersect a calibrated CDF near probability levels 0.025 and 0.975, respectively. The green CDFs follow this behavior closely, whereas the orange CDFs are typically steeper and displaced relative to the red interval. This confirms that the improvement from distributional calibration is not only a correction of the posterior mean but also a correction of the full marginal predictive distribution of each QoI.

\Cref{tab:uq_empirical_calibration} summarizes this calibration check quantitatively. The three reported quantities are complementary: the interval mass measures how much of the predictive distribution lies within the released epistemic bounds, while $F(L)$ and $F(U)$ measure whether the lower and upper truth edges occur at the intended lower and upper tail probabilities. For a central 95\% epistemic interval, the target values are therefore 95\% mass, $F(L)=0.025$, and $F(U)=0.975$. The distribution-calibrated predictions are close to these targets across all twelve QoI/location pairs. In particular, the mass inside $[L,U]$ remains between 94.2\% and 95.8\%, and the endpoint CDF values stay close to the nominal 0.025 and 0.975 levels.

The same table also clarifies why matching only the first moment is insufficient. Several first-moment calibrated predictions have 100\% of their samples inside the truth interval, but their $F(L)$ and $F(U)$ values are 0 and 1, respectively; these cases are over-concentrated relative to the released epistemic uncertainty, not genuinely distribution calibrated. Other cases fail because the predictive distribution is displaced relative to the truth interval: for example, at $\alpha=7^\circ$, $\beta_{\mathrm{flap}}=15^\circ$, the first-moment calibrated $C_l$ result gives $F(L)=0.726$, meaning that most of the predicted mass lies below the lower truth edge, while the corresponding $C_d$ result places only 11\% of the samples inside the truth interval. By contrast, the distribution-calibrated results at the same off-design point retain near-nominal interval mass and endpoint probabilities for all three QoIs. Thus, \Cref{tab:uq_empirical_calibration} supports the visual conclusion from \Cref{fig:calibration_cdf_overview} that distributional calibration corrects the predictive uncertainty itself, not merely the location of the predictive mean.

\begin{table}[H]
\centering
\caption{Empirical calibration of the prediction distributions against the released epistemic intervals. The truth target corresponds to a 95\% interval, so $F(L)=0.025$, $F(U)=0.975$, and the interval mass is 95\%.}
\label{tab:uq_empirical_calibration}
\scriptsize
\setlength{\tabcolsep}{3pt}
\begin{tabular}{cccrrrrrrrrr}
\toprule
$\alpha$ & $\beta_{\mathrm{flap}}$ & QoI & \multicolumn{3}{c}{Mass in $[L,U]$ (\%)} & \multicolumn{3}{c}{$F(L)$} & \multicolumn{3}{c}{$F(U)$} \\
\cmidrule(lr){4-6}\cmidrule(lr){7-9}\cmidrule(lr){10-12}
 & & & Truth & First-moment & Distribution & Truth & First-moment & Dist. & Truth & First-moment & Distribution \\
\midrule
$-3^\circ$ & $-3^\circ$ & $C_l$ & 95.0 & 73.3 & {\bf 94.2} & 0.025 & 0.000 & {\bf 0.029} & 0.975 & 0.733 & {\bf 0.971} \\
$-3^\circ$ & $-3^\circ$ & $C_d$ & 95.0 & 100.0 & {\bf 94.3} & 0.025 & 0.000 & {\bf 0.024} & 0.975 & 1.000 & {\bf 0.967} \\
$-3^\circ$ & $-3^\circ$ & $C_m$ & 95.0 & 100.0 & {\bf 94.5} & 0.025 & 0.000 & {\bf 0.022} & 0.975 & 1.000 & {\bf 0.967} \\
\addlinespace
$0^\circ$ & $0^\circ$ & $C_l$ & 95.0 & 38.2 & {\bf 95.5} & 0.025 & 0.000 & {\bf 0.019} & 0.975 & 0.382 & {\bf 0.974} \\
$0^\circ$ & $0^\circ$ & $C_d$ & 95.0 & 100.0 & {\bf 94.4} & 0.025 & 0.000 & {\bf 0.030} & 0.975 & 1.000 & {\bf 0.974} \\
$0^\circ$ & $0^\circ$ & $C_m$ & 95.0 & 100.0 & {\bf 95.8} & 0.025 & 0.000 & {\bf 0.020} & 0.975 & 1.000 & {\bf 0.978} \\
\addlinespace
$2^\circ$ & $2^\circ$ & $C_l$ & 95.0 & 63.0 & {\bf 94.9} & 0.025 & 0.045 & {\bf 0.025} & 0.975 & 0.675 & {\bf 0.974} \\
$2^\circ$ & $2^\circ$ & $C_d$ & 95.0 & 100.0 & {\bf 94.9} & 0.025 & 0.000 & {\bf 0.023} & 0.975 & 1.000 & {\bf 0.972} \\
$2^\circ$ & $2^\circ$ & $C_m$ & 95.0 & 100.0 & {\bf 94.9} & 0.025 & 0.000 & {\bf 0.026} & 0.975 & 1.000 & {\bf 0.975} \\
\addlinespace
$7^\circ$ & $15^\circ$ & $C_l$ & 95.0 & 27.4 & {\bf 95.8} & 0.025 & 0.726 & {\bf 0.022} & 0.975 & 1.000 & {\bf 0.980} \\
$7^\circ$ & $15^\circ$ & $C_d$ & 95.0 & 11.0 & {\bf 95.1} & 0.025 & 0.000 & {\bf 0.033} & 0.975 & 0.110 & {\bf 0.984} \\
$7^\circ$ & $15^\circ$ & $C_m$ & 95.0 & 100.0 & {\bf 95.5} & 0.025 & 0.000 & {\bf 0.023} & 0.975 & 1.000 & {\bf 0.978} \\
\bottomrule
\end{tabular}
\end{table}

We tabulate the learned latent input locations for each of the $7$ validation points in \Cref{tab:cl,tab:cd,tab:cm}; most of the learned latent inputs $\tilde{\x}$ tend to lie closer to one of the bounds.

\section{Conclusions}
\label{sec:conclusions}

This work developed a calibrated Bayesian surrogate framework for the second AIAA UQ challenge problem, where aerodynamic coefficients must be predicted at unseen operating conditions while accounting for epistemic uncertainty in both the inputs and outputs. The method combines three ingredients: a GP surrogate for the XFOIL response, a Kennedy--O'Hagan-style GP discrepancy model for the residual between XFOIL and truth data, and a latent-input formulation that infers the unknown physical calibration locations within their prescribed epistemic intervals. The principal additional contribution is a distributional calibration step that uses the supplied output intervals as constraints on the corrected predictive distribution, thereby calibrating not only the posterior mean but also the marginal predictive variance of each aerodynamic coefficient.

The empirical results show that this distribution-level correction is essential for reliable uncertainty quantification. Across the four withheld prediction locations and three quantities of interest, the distribution-calibrated model places between 94.2\% and 95.8\% of its predictive samples inside the released 95\% truth intervals, with endpoint CDF values close to the nominal 0.025 and 0.975 levels. This behavior holds even at the more extrapolative operating point $\alpha=7^\circ$, $\beta_{\mathrm{flap}}=15^\circ$. In contrast, first-moment calibration often produces predictive distributions that are shifted, over-concentrated, or both, with interval masses ranging from 11\% to 100\%. These comparisons indicate that matching the center of the truth data is not sufficient; the released epistemic widths must also inform the calibrated uncertainty model.

Several extensions are natural. A fully Bayesian treatment of the latent inputs, GP hyperparameters, and discrepancy field could better quantify posterior uncertainty than the MAP-based implementation used here, particularly with sparse truth data. The distributional calibration strategy could also be extended to correlated multi-output models so that joint uncertainty in $C_l$, $C_d$, and $C_m$ is represented directly. Finally, the framework provides a natural basis for active learning: new XFOIL runs or truth-data acquisitions could be selected where the calibrated predictive distribution is most uncertain, most poorly constrained by the interval information, or most influential for downstream aerodynamic decisions.

\section*{Acknowledgments}
This work was funded by the Penn State Institute of Computational and Data Sciences (ICDS) Junior Researcher Collaboration grant. The authors also acknowledge Penn State ICDS (RRID:SCR\_025154) for providing access to computational research infrastructure within the Roar Core Facility (RRID:SCR\_026424).

\begin{appendix}
\section{Appendix}

\subsection{MAP estimates of latent inputs}
\begin{table}[h!]
\centering
\caption{Learned latent input locations $\tilde{\x}$ for $C_l$ via the latent GP}
\resizebox{\textwidth}{!}{%
\begin{tabular}{c|ccc|ccc|ccc}
\hline
Points & \multicolumn{3}{c|}{$\alpha$} & \multicolumn{3}{c|}{$\beta_{\mathrm{flap}}$} & \multicolumn{3}{c}{Re} \\
 & Lb & Ub & $\tilde{\x}$ & Lb & Ub & $\tilde{\x}$ & Lb & Ub & $\tilde{\x}$ \\
\hline
1 & \( -0.02 \) & \( 0.02 \) & \(\mathbf{0.019962}\)
  & \( -0.10 \) & \( 0.10 \) & \(\mathbf{0.099801}\)
  & \( 696500 \) & \( 703500 \) & \(\mathbf{700000}\) \\

2 & \( 4.98 \) & \( 5.02 \) & \(\mathbf{4.980000}\)
  & \( -0.10 \) & \( 0.10 \) & \(\mathbf{0.099807}\)
  & \( 696500 \) & \( 703500 \) & \(\mathbf{700000}\) \\

3 & \( 9.98 \) & \( 10.02 \) & \(\mathbf{9.980000}\)
  & \( -0.10 \) & \( 0.10 \) & \(\mathbf{0.099807}\)
  & \( 696500 \) & \( 703500 \) & \(\mathbf{700000}\) \\

4 & \( -0.02 \) & \( 0.02 \) & \(\mathbf{0.019961}\)
  & \( 4.90 \) & \( 5.10 \) & \(\mathbf{5.099900}\)
  & \( 696500 \) & \( 703500 \) & \(\mathbf{700000}\) \\

5 & \( -0.02 \) & \( 0.02 \) & \(\mathbf{0.019959}\)
  & \( 9.90 \) & \( 10.10 \) & \(\mathbf{9.900200}\)
  & \( 696500 \) & \( 703500 \) & \(\mathbf{700000}\) \\

6 & \( 4.98 \) & \( 5.02 \) & \(\mathbf{4.980000}\)
  & \( 4.90 \) & \( 5.10 \) & \(\mathbf{4.900200}\)
  & \( 696500 \) & \( 703500 \) & \(\mathbf{700000}\) \\

7 & \( -5.02 \) & \( -4.98 \) & \(\mathbf{-4.980000}\)
  & \( -5.10 \) & \( -4.90 \) & \(\mathbf{-4.900200}\)
  & \( 696500 \) & \( 703500 \) & \(\mathbf{700000}\) \\
\hline
\end{tabular}
}
\label{tab:cl}
\end{table}
\begin{table}[h!]
\centering
\caption{Learned latent input locations $\tilde{\x}$ for $C_d$ via the latent GP}
\resizebox{\textwidth}{!}{%
\begin{tabular}{c|ccc|ccc|ccc}
\hline
Points & \multicolumn{3}{c|}{$\alpha$} & \multicolumn{3}{c|}{$\beta_{\mathrm{flap}}$} & \multicolumn{3}{c}{Re} \\
 & Lb & Ub & $\tilde{\x}$ & Lb & Ub & $\tilde{\x}$ & Lb & Ub & $\tilde{\x}$ \\
\hline
1 & \( -0.02 \) & \( 0.02 \) & \(\mathbf{0.019961}\)
  & \( -0.10 \) & \( 0.10 \) & \(\mathbf{0.099806}\)
  & \( 696500 \) & \( 703500 \) & \(\mathbf{700000}\) \\

2 & \( 4.98 \) & \( 5.02 \) & \(\mathbf{4.980000}\)
  & \( -0.10 \) & \( 0.10 \) & \(\mathbf{0.099803}\)
  & \( 696500 \) & \( 703500 \) & \(\mathbf{700000}\) \\

3 & \( 9.98 \) & \( 10.02 \) & \(\mathbf{9.980000}\)
  & \( -0.10 \) & \( 0.10 \) & \(\mathbf{0.099802}\)
  & \( 696500 \) & \( 703500 \) & \(\mathbf{700000}\) \\

4 & \( -0.02 \) & \( 0.02 \) & \(\mathbf{0.019960}\)
  & \( 4.90 \) & \( 5.10 \) & \(\mathbf{4.900200}\)
  & \( 696500 \) & \( 703500 \) & \(\mathbf{700000}\) \\

5 & \( -0.02 \) & \( 0.02 \) & \(\mathbf{0.019960}\)
  & \( 9.90 \) & \( 10.10 \) & \(\mathbf{9.900200}\)
  & \( 696500 \) & \( 703500 \) & \(\mathbf{700000}\) \\

6 & \( 4.98 \) & \( 5.02 \) & \(\mathbf{4.980000}\)
  & \( 4.90 \) & \( 5.10 \) & \(\mathbf{4.900200}\)
  & \( 696500 \) & \( 703500 \) & \(\mathbf{700000}\) \\

7 & \( -5.02 \) & \( -4.98 \) & \(\mathbf{-4.980000}\)
  & \( -5.10 \) & \( -4.90 \) & \(\mathbf{-4.900200}\)
  & \( 696500 \) & \( 703500 \) & \(\mathbf{700000}\) \\
\hline
\end{tabular}
}
\label{tab:cd}
\end{table}

\begin{table}[h!]
\centering
\caption{Learned latent input locations $\tilde{\x}$ for $C_m$ via the latent GP}
\resizebox{\textwidth}{!}{%
\begin{tabular}{c|ccc|ccc|ccc}
\hline
Points & \multicolumn{3}{c|}{$\alpha$} & \multicolumn{3}{c|}{$\beta_{\mathrm{flap}}$} & \multicolumn{3}{c}{Re} \\
 & Lb & Ub & $\tilde{\x}$ & Lb & Ub & $\tilde{\x}$ & Lb & Ub & $\tilde{\x}$ \\
\hline
1 & \( -0.02 \) & \( 0.02 \) & \(\mathbf{0.019961}\)
  & \( -0.10 \) & \( 0.10 \) & \(\mathbf{0.099805}\)
  & \( 696500 \) & \( 703500 \) & \(\mathbf{700000}\) \\

2 & \( 4.98 \) & \( 5.02 \) & \(\mathbf{4.980000}\)
  & \( -0.10 \) & \( 0.10 \) & \(\mathbf{0.099803}\)
  & \( 696500 \) & \( 703500 \) & \(\mathbf{700000}\) \\

3 & \( 9.98 \) & \( 10.02 \) & \(\mathbf{9.980000}\)
  & \( -0.10 \) & \( 0.10 \) & \(\mathbf{0.099802}\)
  & \( 696500 \) & \( 703500 \) & \(\mathbf{700000}\) \\

4 & \( -0.02 \) & \( 0.02 \) & \(\mathbf{0.019960}\)
  & \( 4.90 \) & \( 5.10 \) & \(\mathbf{4.900200}\)
  & \( 696500 \) & \( 703500 \) & \(\mathbf{700000}\) \\

5 & \( -0.02 \) & \( 0.02 \) & \(\mathbf{0.019960}\)
  & \( 9.90 \) & \( 10.10 \) & \(\mathbf{9.900200}\)
  & \( 696500 \) & \( 703500 \) & \(\mathbf{700000}\) \\

6 & \( 4.98 \) & \( 5.02 \) & \(\mathbf{4.980000}\)
  & \( 4.90 \) & \( 5.10 \) & \(\mathbf{4.900200}\)
  & \( 696500 \) & \( 703500 \) & \(\mathbf{700000}\) \\

7 & \( -5.02 \) & \( -4.98 \) & \(\mathbf{-4.980000}\)
  & \( -5.10 \) & \( -4.90 \) & \(\mathbf{-4.900200}\)
  & \( 696500 \) & \( 703500 \) & \(\mathbf{700000}\) \\
\hline
\end{tabular}
}
\label{tab:cm}
\end{table}

\subsection{Derivation of the calibrated mean and covariance}
\label{sec:derivation}

Let the uncalibrated corrected Gaussian process be
\[
g_0(\mathbf{x}) \sim \mathcal{GP}\bigl(m_0(\mathbf{x}),k_0(\mathbf{x},\mathbf{x}')\bigr).
\]
At the truth-data locations
\[
X_T = [\mathbf{x}_1^T,\ldots,\mathbf{x}_{N_T}^T]^\top,
\]
define
\[
\mathbf{g}_T
=
g_0(X_T)
=
\bigl[g_0(\mathbf{x}_1^T),\ldots,g_0(\mathbf{x}_{N_T}^T)\bigr]^\top .
\]
Under the uncalibrated corrected GP,
\[
\mathbf{g}_T \sim \mathcal{N}(\mathbf{m}_{0T},\mathbf{K}_{TT}),
\qquad
\mathbf{m}_{0T}=m_0(X_T),
\qquad
\mathbf{K}_{TT}=k_0(X_T,X_T).
\]
The distributional calibration target replaces this finite-dimensional marginal by
\[
\mathbf{g}_T \sim q_T
=
\mathcal{N}(\mathbf{c},\mathbf{V}_T).
\]
Equivalently, the calibrated process is defined by
\[
p_{\mathrm{cal}}(g\mid D_X,D_T)
=
\int
p_0(g\mid \mathbf{g}_T=\mathbf{z},D_X)\,
q_T(\mathbf{z})\,d\mathbf{z},
\qquad
q_T(\mathbf{z})=\mathcal{N}(\mathbf{z};\mathbf{c},\mathbf{V}_T).
\]

For any prediction input \(\mathbf{x}\), define
\[
k_{\mathbf{x}T}
=
k_0(\mathbf{x},X_T)
=
\bigl[
k_0(\mathbf{x},\mathbf{x}_1^T),
\ldots,
k_0(\mathbf{x},\mathbf{x}_{N_T}^T)
\bigr],
\]
and similarly
\[
k_{T\mathbf{x}'}
=
k_0(X_T,\mathbf{x}').
\]
The joint Gaussian distribution of \(g_0(\mathbf{x})\) and \(\mathbf{g}_T\) implies the
standard conditional mean
\[
\mathbb{E}_0
\left[
g_0(\mathbf{x})
\mid
\mathbf{g}_T=\mathbf{z},D_X
\right]
=
m_0(\mathbf{x})
+
k_{\mathbf{x}T}\mathbf{K}_{TT}^{-1}
(\mathbf{z}-\mathbf{m}_{0T}),
\]
and conditional covariance
\[
\operatorname{Cov}_0
\left[
g_0(\mathbf{x}),g_0(\mathbf{x}')
\mid
\mathbf{g}_T=\mathbf{z},D_X
\right]
=
k_0(\mathbf{x},\mathbf{x}')
-
k_{\mathbf{x}T}\mathbf{K}_{TT}^{-1}k_{T\mathbf{x}'}.
\]

Taking expectation over
\(\mathbf{z}\sim q_T=\mathcal{N}(\mathbf{c},\mathbf{V}_T)\), the calibrated mean is
\[
\begin{aligned}
m_{\mathrm{cal}}(\mathbf{x})
&=
\mathbb{E}_{q_T}
\left[
\mathbb{E}_0
\left[
g_0(\mathbf{x})
\mid
\mathbf{g}_T=\mathbf{z},D_X
\right]
\right] \\
&=
\mathbb{E}_{q_T}
\left[
m_0(\mathbf{x})
+
k_{\mathbf{x}T}\mathbf{K}_{TT}^{-1}
(\mathbf{z}-\mathbf{m}_{0T})
\right] \\
&=
m_0(\mathbf{x})
+
k_{\mathbf{x}T}\mathbf{K}_{TT}^{-1}
(\mathbf{c}-\mathbf{m}_{0T}).
\end{aligned}
\]
Therefore,
\[
\boxed{
m_{\mathrm{cal}}(\mathbf{x})
=
m_0(\mathbf{x})
+
k_{\mathbf{x}T}\mathbf{K}_{TT}^{-1}
(\mathbf{c}-\mathbf{m}_{0T})
}.
\]

For the covariance, use the law of total covariance:
\[
\begin{aligned}
k_{\mathrm{cal}}(\mathbf{x},\mathbf{x}')
&=
\mathbb{E}_{q_T}
\left[
\operatorname{Cov}_0
\left[
g_0(\mathbf{x}),g_0(\mathbf{x}')
\mid
\mathbf{g}_T=\mathbf{z},D_X
\right]
\right] \\
&\quad+
\operatorname{Cov}_{q_T}
\left(
\mathbb{E}_0[g_0(\mathbf{x})\mid \mathbf{g}_T=\mathbf{z},D_X],
\mathbb{E}_0[g_0(\mathbf{x}')\mid \mathbf{g}_T=\mathbf{z},D_X]
\right).
\end{aligned}
\]
The first term is
\[
k_0(\mathbf{x},\mathbf{x}')
-
k_{\mathbf{x}T}\mathbf{K}_{TT}^{-1}k_{T\mathbf{x}'},
\]
since the conditional covariance does not depend on \(\mathbf{z}\). The second term is
\[
\begin{aligned}
&\operatorname{Cov}_{q_T}
\left(
k_{\mathbf{x}T}\mathbf{K}_{TT}^{-1}\mathbf{z},
k_{\mathbf{x}'T}\mathbf{K}_{TT}^{-1}\mathbf{z}
\right)  =
k_{\mathbf{x}T}
\mathbf{K}_{TT}^{-1}
\mathbf{V}_T
\mathbf{K}_{TT}^{-1}
k_{T\mathbf{x}'}.
\end{aligned}
\]
Hence
\[
\boxed{
k_{\mathrm{cal}}(\mathbf{x},\mathbf{x}')
=
k_0(\mathbf{x},\mathbf{x}')
-
k_{\mathbf{x}T}\mathbf{K}_{TT}^{-1}k_{T\mathbf{x}'}
+
k_{\mathbf{x}T}
\mathbf{K}_{TT}^{-1}
\mathbf{V}_T
\mathbf{K}_{TT}^{-1}
k_{T\mathbf{x}'}
}.
\]

\end{appendix}
\clearpage
\bibliographystyle{plainnat}
\bibliography{references}

@article{yuan2026remal,
  title={REMAL: Residual Equilibrium Manifold Active Learning for Surrogate-Based Multidisciplinary Design Analysis},
  author={Yuan, Kail and Renganathan, Ashwin},
  journal={arXiv preprint arXiv:2606.13245},
  year={2026}
}

@inproceedings{adhikary2026adaptive,
  title={Adaptive Multitask Gaussian Process Surrogate Models for Supersonic Fluid-Structure Interactions},
  author={Adhikary, Srishti and Narayanaswamy, Venkat and Renganathan, Ashwin},
  booktitle={AIAA AVIATION 2026 Forum},
  pages={4781},
  doi={https://doi.org/10.2514/6.2026-4781},
  year={2026}
}

@article{renganathan2021lookahead,
  title={Lookahead acquisition functions for finite-horizon time-dependent bayesian optimization and application to quantum optimal control},
  author={Renganathan, S Ashwin and Larson, Jeffrey and Wild, Stefan M},
  journal={arXiv preprint arXiv:2105.09824},
  year={2021}
}

@inproceedings{davis2026uncertainty,
  title={Uncertainty Quantification via Latent Gaussian Process Surrogates for the Second AIAA Fluid Dynamics UQ Challenge Problem},
  author={Davis, Geoffrey and Renganathan, Ashwin},
  booktitle={AIAA SCITECH 2026 Forum},
  pages={0095},
  year={2026}
}

@article{loh1996latin,
  title={On Latin hypercube sampling},
  author={Loh, Wei-Liem},
  journal={The annals of statistics},
  volume={24},
  number={5},
  pages={2058--2080},
  year={1996},
  doi={https://doi.org/10.1214/aos/1069362310},
  publisher={Institute of Mathematical Statistics}
}

@inproceedings{cary2026summary,
  title={Summary of the Second AIAA Uncertainty Quantification Challenge Problem for Aerodynamics},
  author={Cary, Andrew W and Rumpfkeil, Markus and Hristov, Peter O and Schaefer, John A},
  booktitle={AIAA SCITECH 2026 Forum},
  pages={0091},
  doi={https://doi.org/10.2514/6.2026-0091},
  year={2026}
}

@article{sacks1989design,
  title={Design and analysis of computer experiments},
  author={Sacks, Jerome and Welch, William J and Mitchell, Toby J and Wynn, Henry P},
  journal={Statistical science},
  volume={4},
  number={4},
  pages={409--423},
  year={1989},
  doi     = {https://doi.org/10.1214/ss/1177012413},
  publisher={Institute of Mathematical Statistics}
}

@book{santner2003design,
  title={The design and analysis of computer experiments},
  author={Santner, Thomas J and Williams, Brian J and Notz, William I and Williams, Brain J},
  volume={1},
  year={2003},
  doi= {https://doi.org/10.1007/978-1-4939-8847-1},
  publisher={Springer}
}

@book{gramacy2020surrogates,
  title     = {Surrogates: Gaussian Process Modeling, Design, and Optimization for the Applied Sciences},
  author    = {Gramacy, Robert B.},
  publisher = {Chapman and Hall/CRC},
  address   = {Boca Raton, FL},
  year      = {2020},
  doi       = {https://doi.org/10.1201/9780367815493}
}

@article{kennedy2001bayesian,
  title   = {Bayesian Calibration of Computer Models},
  author  = {Kennedy, Marc C. and O'Hagan, Anthony},
  journal = {Journal of the Royal Statistical Society: Series B (Statistical Methodology)},
  volume  = {63},
  number  = {3},
  pages   = {425--464},
  year    = {2001},
  doi     = {https://doi.org/10.1111/1467-9868.00294},
  publisher={Wiley Online Library}
}

@article{higdon2004combining,
  title   = {Combining Field Data and Computer Simulations for Calibration and Prediction},
  author  = {Higdon, Dave and Kennedy, Marc and Cavendish, James C. and Cafeo, John A. and Ryne, Robert D.},
  journal = {SIAM Journal on Scientific Computing},
  volume  = {26},
  number  = {2},
  pages   = {448--466},
  year    = {2004},
  doi     = {https://doi.org/10.1137/S1064827503426693}
}

@article{higdon2008computer,
  title   = {Computer Model Calibration Using High-Dimensional Output},
  author  = {Higdon, Dave and Gattiker, James and Williams, Brian and Rightley, Maria},
  journal = {Journal of the American Statistical Association},
  volume  = {103},
  number  = {482},
  pages   = {570--583},
  year    = {2008},
  doi     = {https://doi.org/10.1198/016214507000000888}
}

@article{bayarri2007framework,
  title   = {A Framework for Validation of Computer Models},
  author  = {Bayarri, M. J. and Berger, James O. and Paulo, Rui and Sacks, Jerry and Cafeo, John A. and Cavendish, James and Lin, Chin-Hsu and Tu, Jian},
  journal = {Technometrics},
  volume  = {49},
  number  = {2},
  pages   = {138--154},
  year    = {2007},
  doi     = {https://doi.org/10.1198/004017007000000092}
}

@article{liu2009modularization,
  title   = {Modularization in Bayesian Analysis, with Emphasis on Analysis of Computer Models},
  author  = {Liu, Fei and Bayarri, Maria J. and Berger, James O.},
  journal = {Bayesian Analysis},
  volume  = {4},
  number  = {1},
  pages   = {119--150},
  year    = {2009},
  doi     = {https://doi.org/10.1214/09-BA404}
}

@article{brynjarsdottir2014learning,
  title   = {Learning about Physical Parameters: The Importance of Model Discrepancy},
  author  = {Brynjarsd{\'o}ttir, Jenn{\'y} and O'Hagan, Anthony},
  journal = {Inverse Problems},
  volume  = {30},
  number  = {11},
  pages   = {114007},
  year    = {2014},
  doi     = {https://doi.org/10.1088/0266-5611/30/11/114007}
}

@article{arendt2012quantification,
  title   = {Quantification of Model Uncertainty: Calibration, Model Discrepancy, and Identifiability},
  author  = {Arendt, Paul D. and Apley, Daniel W. and Chen, Wei},
  journal = {Journal of Mechanical Design},
  volume  = {134},
  number  = {10},
  pages   = {100908},
  year    = {2012},
  doi     = {https://doi.org/10.1115/1.4007390}
}

@article{tuo2015efficient,
  title   = {Efficient Calibration for Imperfect Computer Models},
  author  = {Tuo, Rui and Wu, C. F. Jeff},
  journal = {The Annals of Statistics},
  volume  = {43},
  number  = {6},
  pages   = {2331--2352},
  year    = {2015},
  doi     = {https://doi.org/10.1214/15-AOS1314}
}

@article{tuo2016theoretical,
  title   = {A Theoretical Framework for Calibration in Computer Models: Parametrization, Estimation and Convergence Properties},
  author  = {Tuo, Rui and Wu, C. F. Jeff},
  journal = {SIAM/ASA Journal on Uncertainty Quantification},
  volume  = {4},
  number  = {1},
  pages   = {767--795},
  year    = {2016},
  doi     = {https://doi.org/10.1137/151005841}
}

@article{plumlee2017bayesian,
  title   = {Bayesian Calibration of Inexact Computer Models},
  author  = {Plumlee, Matthew},
  journal = {Journal of the American Statistical Association},
  volume  = {112},
  number  = {519},
  pages   = {1274--1285},
  year    = {2017},
  doi     = {https://doi.org/10.1080/01621459.2016.1211016}
}

@article{wang2017propagation,
  title   = {Propagation of Input Uncertainty in Presence of Model-Form Uncertainty: A Multifidelity Approach for Computational Fluid Dynamics Applications},
  author  = {Wang, Jian-Xun and Roy, Christopher J. and Xiao, Heng},
  journal = {ASCE-ASME Journal of Risk and Uncertainty in Engineering Systems, Part B: Mechanical Engineering},
  volume  = {4},
  number  = {1},
  pages   = {011002},
  year    = {2018},
  doi     = {https://doi.org/10.1115/1.4037452}
}

@article{girard2003gpuncertain,
  title={Gaussian process priors with uncertain inputs application to multiple-step ahead time series forecasting},
  author={Girard, Agathe and Rasmussen, Carl and Candela, Joaquin Q and Murray-Smith, Roderick},
  journal={Advances in neural information processing systems},
  volume={15},
  year={2002}
}

@inproceedings{quinonero2003propagation,
  title     = {Propagation of Uncertainty in Bayesian Kernel Models---Application to Multiple-Step Ahead Forecasting},
  author    = {Qui{\~n}onero-Candela, Joaquin and Girard, Agathe and Larsen, Jan and Rasmussen, Carl Edward},
  booktitle = {Proceedings of the IEEE International Conference on Acoustics, Speech, and Signal Processing},
  volume    = {2},
  pages     = {701--704},
  publisher = {IEEE},
  year      = {2003},
  doi       = {https://doi.org/10.1109/ICASSP.2003.1202463}
}

@inproceedings{dallaire2009learning,
  title     = {Learning Gaussian Process Models from Uncertain Data},
  author    = {Dallaire, Patrick and Besse, Camille and Chaib-draa, Brahim},
  booktitle = {Neural Information Processing},
  series    = {Lecture Notes in Computer Science},
  volume    = {5863},
  pages     = {433--440},
  publisher = {Springer},
  year      = {2009},
  doi       = {https://doi.org/10.1007/978-3-642-10677-4_49}
}

@inproceedings{mchutchon2011gaussian,
  title     = {Gaussian Process Training with Input Noise},
  author    = {McHutchon, Andrew and Rasmussen, Carl Edward},
  booktitle = {Advances in Neural Information Processing Systems},
  volume    = {24},
  pages     = {1341--1349},
  year      = {2011}
}

@article{lawrence2005probabilistic,
  title   = {Probabilistic Non-Linear Principal Component Analysis with Gaussian Process Latent Variable Models},
  author  = {Lawrence, Neil D.},
  journal = {Journal of Machine Learning Research},
  volume  = {6},
  pages   = {1783--1816},
  year    = {2005}
}

@inproceedings{titsias2010bayesian,
  title     = {Bayesian Gaussian Process Latent Variable Model},
  author    = {Titsias, Michalis K. and Lawrence, Neil D.},
  booktitle = {Proceedings of the Thirteenth International Conference on Artificial Intelligence and Statistics},
  series    = {Proceedings of Machine Learning Research},
  volume    = {9},
  pages     = {844--851},
  year      = {2010}
}

@article{damianou2016variational,
  title   = {Variational Inference for Latent Variables and Uncertain Inputs in Gaussian Processes},
  author  = {Damianou, Andreas C. and Titsias, Michalis K. and Lawrence, Neil D.},
  journal = {Journal of Machine Learning Research},
  volume  = {17},
  pages   = {1--62},
  year    = {2016}
}

@article{perrin2019taking,
  title   = {Taking into Account Input Uncertainties in the Bayesian Calibration of Time-Consuming Simulators},
  author  = {Perrin, Guillaume and Durantin, C{\'e}dric},
  journal = {Journal de la Soci{\'e}t{\'e} Fran{\c c}aise de Statistique},
  volume  = {160},
  number  = {2},
  pages   = {24--46},
  year    = {2019}
}

@inproceedings{drela1989xfoil,
  title={XFOIL: An Analysis and Design System for Low Reynolds Number Airfoils},
  author={Drela, Mark},
  booktitle={Low Reynolds number aerodynamics},
  pages={1--12},
  year={1989},
  organization={Springer}
}

@article{renganathan2021enhanced,
  title={Enhanced data efficiency using deep neural networks and Gaussian processes for aerodynamic design optimization},
  author={Renganathan, S Ashwin and Maulik, Romit and Ahuja, Jai},
  journal={Aerospace Science and Technology},
  volume={111},
  pages={106522},
  year={2021},
  doi={https://doi.org/10.1016/j.ast.2021.106522},
  publisher={Elsevier}
}

@article{renganathan2023camera,
  title={CAMERA: A method for cost-aware, adaptive, multifidelity, efficient reliability analysis},
  author={Renganathan, S Ashwin and Rao, Vishwas and Navon, Ionel M},
  journal={Journal of Computational Physics},
  volume={472},
  pages={111698},
  year={2023},
  doi={https://doi.org/10.1016/j.jcp.2022.111698},
  publisher={Elsevier}
}

@inproceedings{renganathan2022multifidelity,
  title={Multifidelity Gaussian processes for failure boundary and probability estimation},
  author={Renganathan, Ashwin and Rao, Vishwas and Navon, Ionel},
  booktitle={AIAA Scitech 2022 Forum},
  pages={0390},
  doi={https://doi.org/10.2514/6.2022-0390},
  year={2022}
}

@article{carlson2025multiobjective,
  title={Multiobjective aerodynamic design optimization of the NASA common research model},
  author={Carlson, Kade and Renganathan, Ashwin},
  journal={Aerospace Science and Technology},
  pages={111120},
  year={2025},
  doi={https://doi.org/10.1016/j.ast.2025.111120},
  publisher={Elsevier}
}

@inproceedings{renganathan2024efficient,
  title={Efficient reliability analysis with multifidelity Gaussian processes and normalizing flows},
  author={Renganathan, Ashwin},
  booktitle={AIAA SCITECH 2024 Forum},
  pages={0576},
  doi={https://doi.org/10.2514/6.2024-0576},
  year={2024}
}

@article{ashwin2022data,
  title={Data-driven wind turbine wake modeling via probabilistic machine learning},
  author={Ashwin Renganathan, S and Maulik, Romit and Letizia, Stefano and Iungo, Giacomo Valerio},
  journal={Neural Computing and Applications},
  volume={34},
  number={8},
  pages={6171--6186},
  year={2022},
  doi = {https://doi.org/10.1007/s00521-021-06799-6},
  publisher={Springer}
}

@article{booth2025contour,
  title={Contour location for reliability in airfoil simulation experiments using deep gaussian processes},
  author={Booth, Annie S and Renganathan, S Ashwin and Gramacy, Robert B},
  journal={The Annals of Applied Statistics},
  volume={19},
  number={1},
  pages={191--211},
  year={2025},
  doi={https://doi.org/10.1214/24-AOAS1951},
  publisher={Institute of Mathematical Statistics}
}

@article{betancourt2017conceptual,
  title={A conceptual introduction to Hamiltonian Monte Carlo},
  author={Betancourt, Michael},
  journal={arXiv preprint arXiv:1701.02434},
  year={2017}
}

@article{blei2017variational,
  title={Variational inference: A review for statisticians},
  author={Blei, David M and Kucukelbir, Alp and McAuliffe, Jon D},
  journal={Journal of the American statistical Association},
  volume={112},
  number={518},
  pages={859--877},
  year={2017},
  doi={https://doi.org/10.1080/01621459.2017.1285773},
  publisher={Taylor \& Francis}
}

@inproceedings{renganathan2025q,
  title={qPOTS: Efficient Batch Multiobjective Bayesian Optimization via Pareto Optimal Thompson Sampling},
  author={Renganathan, Ashwin and Carlson, Kade},
  booktitle={International Conference on Artificial Intelligence and Statistics},
  pages={4051--4059},
  year={2025},
  organization={PMLR}
}

@article{Oberkampf2004,
  author    = {William L. Oberkampf and Christopher J. Roy},
  title     = {Verification and Validation in Scientific Computing},
  journal   = {Cambridge University Press},
  year      = {2010}
}

@inproceedings{Cary2022,
  author    = {Andrew W. Cary and John A. Schaefer and Earl P. N. Duque and Manas S. Khurana and Erin C. DeCarlo},
  title     = {Overview of Challenges in Performing Uncertainty Quantification for Fluids Engineering Problems},
  booktitle = {AIAA SciTech Forum},
  year      = {2022},
  doi       = {https://doi.org/10.2514/6.2022-2357}
}

@inproceedings{cary2024overview,
  title={Overview of fluid dynamics uncertainty quantification challenge problem and results},
  author={Cary, Andrew W and Schaefer, John A and DeCarlo, Erin C and Duque, Earl P and Khurana, Manas},
  booktitle={AIAA SCITECH 2024 Forum},
  pages={0705},
  doi={https://doi.org/10.2514/6.2024-0705},
  year={2024}
}

@book{Rasmussen2006,
  author    = {Carl E. Rasmussen and Christopher K. I. Williams},
  title     = {Gaussian Processes for Machine Learning},
  publisher = {MIT Press},
  year      = {2006}
}

\end{document}